\documentclass[twoside]{article}
\usepackage{geometry}

\usepackage{ecj,palatino,epsfig,latexsym,natbib}

\usepackage{booktabs} 
\usepackage{multirow}

\usepackage{amssymb}
\usepackage{amsthm}

\usepackage{amsmath}

\usepackage{colortbl}

\usepackage[figuresright]{rotating}

\usepackage{lineno,hyperref}

\usepackage{xcolor}
\definecolor{mycyan}{gray}{.8}

\newcommand{\bb}[1]{\multicolumn{1}{>{\columncolor{mycyan}}c}{\textbf{{#1}}}}

\usepackage{setspace}

\begin{document}


\ecjHeader{x}{x}{xxx-xxx}{201X}{Difficulty Adjustable and Scalable Constrained Multi-objective Test Problem Toolkit}{Z. Fan et al.}
\title{\bf Difficulty Adjustable and Scalable Constrained Multi-objective Test Problem Toolkit}

\author{\name{\bf Zhun Fan} \hfill \addr{zfan@stu.edu.cn}\\
        \addr{Department of Electronic Engineering, Shantou University,
        Guangdong, 515063, China.}\\
        \addr{Key Lab of Digital Signal and Image Processing of Guangdong Province, Guangdong, China.}
\AND
       \name{\bf Wenji Li} \hfill \addr{liwj@stu.edu.cn}\\
        \addr{Department of Electronic Engineering, Shantou University,
        Guangdong, 515063, China.}
\AND
       \name{\bf Xinye Cai} \hfill \addr{xinye@nuaa.edu.cn}\\
        \addr{College of Computer Science and Technology, Nanjing University of Aeronautics and Astronautics, Jiangsu, 210016, China.}
\AND
       \name{\bf Hui Li} \hfill \addr{lihui10@xjtu.edu.cn}\\
        \addr{School of Mathematics and Statistics, Xi'an Jiaotong University, Shan'xi, 710049, China.}
\AND
       \name{\bf Caimin Wei} \hfill \addr{cmwei@stu.edu.cn}\\
        \addr{Department of Mathematics, Shantou University, Guangdong, 515063, China.}
\AND
       \name{\bf Qingfu Zhang} \hfill \addr{qingfu.zhang@cityu.edu.hk}\\
        \addr{Department of Computer Science, City University of Hong Kong, Hong Kong, China.}
\AND
       \name{\bf Kalyanmoy Deb} \hfill \addr{kdeb@egr.msu.edu}\\
        \addr{BEACON Center for the Study of Evolution in Action, Michigan State University. East Lansing, Michigan, USA.}
\AND
       \name{\bf Erik Goodman} \hfill \addr{goodman@egr.msu.edu}\\
        \addr{BEACON Center for the Study of Evolution in Action, Michigan State University. East Lansing, Michigan, USA.}
}

\maketitle

\begin{abstract}
Multi-objective evolutionary algorithms (MOEAs) have progressed significantly in recent decades, but most of them are designed to solve unconstrained multi-objective optimization problems. In fact, many real-world multi-objective problems contain a number of constraints. To promote research on constrained multi-objective optimization, we first propose a problem classification scheme with three primary types of difficulty, which reflect various types of challenges presented by real-world optimization problems, in order to characterize the constraint functions in constrained multi-objective optimization problems (CMOPs). These are feasibility-hardness, convergence-hardness and diversity-hardness. We then develop a general toolkit to construct difficulty-adjustable and scalable CMOPs (DAS-CMOPs, or DAS-CMaOPs when the number of objectives is greater than three) with three types of parameterized constraint functions developed to capture the three proposed types of difficulty. In fact, the combination of the three primary constraint functions with different parameters allows the construction of a large variety of CMOPs, with difficulty that can be defined by a triplet, with each of its parameters specifying the level of one of the types of primary difficulty. Furthermore, the number of objectives in this toolkit can be scaled beyond three. Based on this toolkit, we suggest nine difficulty-adjustable and scalable CMOPs and nine CMaOPs, to be called DAS-CMOP1-9 and DAS-CMaOP1-9, respectively. To evaluate the proposed test problems, two popular CMOEAs - MOEA/D-CDP (MOEA/D with constraint dominance principle) and NSGA-II-CDP (NSGA-II with constraint dominance principle) and two popular constrained many-objective evolutionary algorithms (CMaOEAs)---C-MOEA/DD and C-NSGA-III---are used to compare performance on DAS-CMOP1-9 and DAS-CMaOP1-9 with a variety of difficulty triplets, respectively. The experimental results reveal that mechanisms in MOEA/D-CDP may be more effective in solving convergence-hard DAS-CMOPs, while mechanisms of NSGA-II-CDP may be more effective in solving DAS-CMOPs with simultaneous diversity-, feasibility- and convergence-hardness. Mechanisms in C-NSGA-III may be more effective in solving feasibility-hard CMaOPs, while mechanisms of C-MOEA/DD may be more effective in solving CMaOPs with convergence-hardness. In addition, none of them can solve these problems efficiently, which stimulates us to continue to develop new CMOEAs and CMaOEAs to solve the suggested DAS-CMOPs and DAS-CMaOPs.
\end{abstract}


\section{Introduction}

Practical optimization problems usually involve simultaneous optimization of multiple conflicting objectives with many constraints. Without loss of generality, constrained multi-objective optimization problems (CMOPs) can be defined as follows:
\begin{eqnarray}
\mbox{minimize} &\mathbf{F}(\mathbf{x}) = {(f_{1}(\mathbf{x}),\ldots,f_{m}(\mathbf{x}))} ^ {T} \label{CMOP}\\
\nonumber \mbox{subject to} & g_i(\mathbf{x}) \ge 0, i = 1,\ldots,q \\
\nonumber & h_j(\mathbf{x}) = 0, j= 1,\ldots,p \\
\nonumber &\mathbf{x} \in{\mathbb{R}^n}
\end{eqnarray}
where $\mathbf{F}(\mathbf{x}) = ({f_1}(\mathbf{x}),{f_2}(\mathbf{x}), \ldots ,{f_m}(\mathbf{x})) ^ T \in \mathbb{R}^m$ is an $m$-dimensional objective vector, ${g_i}(\mathbf{x}) \ge 0$ defines the $i$-th of $q$ inequality constraints, and ${h_j}(\mathbf{x})=0$ defines the $j$-th of $p$ equality constraints. If $m$ is greater than three, we usually call it a constrained many-objective optimization problem (CMaOP).

A solution $\mathbf{x}$ is said to be feasible if it meets ${g_i}(\mathbf{x}) \ge 0, i = 1,\ldots,q$ and ${h_j}(\mathbf{x})=0 , j= 1,\ldots,p$ at the same time. For two feasible solutions $\mathbf{x}^{1}$ and $\mathbf{x}^{2}$, solution $\mathbf{x}^{1}$ is said to dominate $\mathbf{x}^{2}$ if $f_i(\mathbf{x} ^{1}) \le f_i(\mathbf{x}^{2})$ for each $i \in {\{1,\ldots,m\}}$ and $f_j(\mathbf{x} ^{1}) < f_j(\mathbf{x}^{2})$ for at least one $j \in {\{1,\ldots,m\}}$, denoted as $\mathbf{x} ^{1} \preceq \mathbf{x} ^ {2}$. For a feasible solution $\mathbf{x} ^{*} \in {\mathbb{R}^n}$, if there is no other feasible solution $\mathbf{x} \in {\mathbb{R}^n}$ dominating $\mathbf{x} ^{*}$, $\mathbf{x} ^{*}$ is said to be a feasible Pareto-optimal solution. The set of all feasible Pareto-optimal solutions is called the Pareto set ($PS$). Mapping the $PS$ into the objective space defines a set of objective vectors, denoted as the Pareto front ($PF$), where $PF = \{ \mathbf{F}( \mathbf{x}) \in \mathbb{R} ^{m} | \mathbf{x} \in PS\}$.

For CMOPs, more than one objective must be optimized simultaneously, subject to constraints. Generally speaking, CMOPs are much more difficult to solve than their unconstrained counterparts---unconstrained multi-objective optimization problems (MOPs). Constrained multi-objective evolutionary algorithms (CMOEAs) are particularly designed to solve CMOPs, with the capability of balancing the search between the feasible and infeasible regions in the search space \citep{runarsson2005search}. In fact, two basic issues must be considered carefully when designing a CMOEA. One is to manage the effort devoted to seeking feasible, rather than infeasible, solutions; the other is to balance the convergence and diversity of the CMOEA.

To address the first issue, constraint handling mechanisms must be carefully designed by researchers. The existing constraint-handling methods can be broadly classified into five different types:  penalty functions, special representations and operators, repair algorithms, separation of constraints and objectives, and ensemble-of-constraint-handling-methods approaches, as delineated in \citep{CoelloCoello:2002gp,mezura2011constraint,Coello:2018}. The penalty function-based method is one of the most popular of such approaches. The overall constraint violation is added to each objective with a predefined penalty factor, which indicates a preference between the satisfaction of the constraints and the extremization of the objectives. Penalty function-based methods include use of a death penalty \citep{hoffmeister1996problem}, static penalty \citep{hoffmeister1996problem}, dynamic penalty \citep{joines1994use}, adaptive penalty \citep{coit1996adaptive,ben1997genetic}, co-evolutionary penalty \citep{huang2007effective}, and self-adaptive penalty \citep{tessema2006self,woldesenbet2009constraint}, etc. The methods using special representations and operators try to map chromosomes from the infeasible region into the feasible region. A more interesting type of approach within this group is the so-called "Decoders" method, which is based on the idea of mapping the feasible region into an easier-to-sample space where an evolutionary algorithm can provide better performance \citep{koziel1998decoder}. Representative examples of this type include homomorphous maps (HM) \citep{koziel1999evolutionary} and Riemann mapping \citep{kim1998riemann}. The third type, repair algorithms, try to transform an infeasible solution into a feasible one. A typical example of this type is Genocop III (genetic algorithm for numerical optimization of constrained problems) \citep{487460}. However, this type of approach is problem-dependent and is usually used in combinatorial optimization problems \citep{salcedo2009survey}. For a particular problem, a specific repair algorithm must be designed. In the methods using separation of constraints and objectives, the constraint functions and the objective functions are treated separately. Some examples of this type include coevolution \citep{paredis1994co}, constraint dominance principle (CDP) \citep{deb2000efficient,deb2001constrained}, stochastic ranking \citep{runarsson2000stochastic}, $\alpha$-constrained method (also called the $\varepsilon$-constrained method) \citep{1514470,laumanns2006efficient,takahama2005constrained}, use of multi-objective optimization concepts \citep{cai2006multiobjective,ray2009infeasibility} and COMOGA \citep{surry1997comoga}. The ensemble-of-constraint-handling-methods approaches usually adopt several constraint-handling methods to deal with constraints. Representative methods include the adaptive trade-off model (ATM) \citep{wang2008adaptive} and the ensemble of constraint handling methods (ECHM) \citep{mallipeddi2010ensemble,qu2011constrained}.

To address the second issue, selection methods need to be designed to balance the performance in terms of convergence versus the diversity in an MOEA. At present, MOEAs can be generally classified into three categories based on selection strategies. They are Pareto-dominance-based (e.g., NSGA-II \citep{deb2002fast}, PAES-II \citep{corne2001pesa} and SPEA-II \citep{zitzler2001spea2}), decomposition-based (e.g., MOEA/D \citep{zhang2007moea}, MOEA/D-DE \citep{li2009multiobjective}, MOEA/D-M2M \citep{liu2014decomposition} and EAG-MOEA/D \citep{cai2015external}) and indicator-based (e.g., IBEA \citep{zitzler2004indicator}, R2-IBEA \citep{phan2013r2}, SMS-EMOA \citep{beume2007sms} and HypE \citep{bader2011hype}). In the group of Pareto-dominance-based methods, such as NSGA-II \citep{deb2002fast}, the first set of non-dominated solutions is selected to improve the convergence performance, and a crowding distance measure in the last-ranked non-dominated set is adopted to maintain the diversity performance. In decomposition-based methods, the convergence performance is maintained by minimizing the aggregation functions and the diversity performance is obtained by setting the weight vectors uniformly. In indicator-based methods, such as HypE \citep{bader2011hype}, the convergence and diversity performance are both achieved using a hypervolume metric.

A CMOP includes objectives and constraints. A number of features have already been identified to define the difficulty of objectives, which include:
\begin{enumerate}
\item Geometry of a $PF$ (linear, convex, concave, degenerate, disconnected or a mixture of them)
\item Search space (biased, or unbiased)
\item Unimodal or multi-modal objectives
\item Dimensionality of variable space and objective space
\end{enumerate}

The first of these features is the geometry of the $PF$. The $PF$ of an MOP can be linear, convex, concave, degenerate, disconnected or a mixture of them. Representative benchmarking MOPs reflecting this type of difficulty include ZDT \citep{deb1999multi}, F1-9 \citep{li2009multiobjective} and DTLZ \citep{deb2005scalable}. The second feature is the biased or unbiased nature of the search space, which means whether an evenly distributed sample of decision vectors in the search space maps to an evenly distributed set of objective vectors in the objective space \citep{huband2006review}. Representative benchmarking MOPs with biased search spaces include MOP1-7 \citep{liu2014decomposition} and IMB1-14 \citep{Liu:2016jf}. The third feature is the modality of objectives. The objectives of an MOP can be either unimodal (for example, DTLZ1 \citep{deb2005scalable}) or multi-modal (for example, F8 \citep{li2009multiobjective}). Multi-modal objectives have multiple locally optimal solutions, which increases the likelihood of an algorithm being trapped in a local optimum. High dimensionality of variable space and objective space are also critical features in defining the difficulty of objectives. LSMOP1-9 \citep{Cheng:2016ca} have high dimensionality in the variable space. DTLZ \citep{deb2005scalable} and WFG \citep{huband2006review} have high dimensionality in the objective space.

On the other hand, constraint functions in general greatly increase the difficulty of solving CMOPs. However, as far as we know, only relatively few test suites (CTP \citep{deb2001multi}, CF \citep{zhang2008multiobjective}) have been designed for CMOPs.

The CTP test problems \citep{deb2001multi} allow adjusting of the difficulty of the constraint functions. They offer two types of difficulties: the difficulty near the $PF$ and the difficulty in the entire search space. The test problem CTP1 gives difficulty near the $PF$, because the constraint functions of CTP1 make the search region near the Pareto front infeasible. Test problems CTP2-CTP8 provide difficulty for an optimizer throughout the search space.

The CF test problems \citep{zhang2008multiobjective} are also commonly used benchmarks, and provide two types of difficulties. For CF1-CF3 and CF8-CF10, the $PFs$ are parts of their unconstrained $PFs$. The rest of the CF test problems CF4-CF7 have difficulties near their $PFs$, and many constrained Pareto-optimal points lie on constraint boundaries.

Even though CTP \citep{deb2001multi} and CF \citep{zhang2008multiobjective} offer the above-mentioned advantages, they have some limitations:
\begin{itemize}
\item The difficulty level of each type is not adjustable.
\item No constraint functions with low ratios of feasible regions in the entire search space are suggested.
\item The number of objectives is not scalable.
\end{itemize}

Other sometimes used two-objective test problems include BNH \citep{binh1997mobes}, TNK \citep{tanaka1995ga}, SRN \citep{srinvas1994multi} and OSY \citep{osyczka1995new} problems, which are not scalable in the number of objectives, and in which it is harder to identify their types of difficulties.

In this paper, we propose a general framework to construct difficulty adjustable and objective scalable CMOPs which can overcome the limitations of existing CMOPs. CMOPs constructed by this toolkit can be classified into three major types, which are feasibility-hard, convergence-hard and diversity-hard CMOPs. A feasibility-hard CMOP is a type of problem that makes it difficult for CMOEAs to find feasible solutions in the search space. CMOPs with feasibility-hardness usually have small portions of the entire search space that are feasible regions. In contrast, CMOPs with convergence-hardness mainly make it difficult for CMOEAs to approach the $PFs$ efficiently, by setting many infeasibility obstacles before the $PFs$. CMOPs with diversity-hardness mainly make it difficult for CMOEAs to distribute their solutions along the complete $PFs$. In our work, any or all three types of difficulty may be embedded into CMOPs through proper construction of constraint functions.

Of course, one interested in solving a particular CMOP must be cautious not to put too much reliance in the performance of a given algorithm on \emph{any} set of benchmark problems. It would be extremely useful if there was a technique to characterize an arbitrary real-world problem, with its generally long evaluation time, according to a set of characteristics that then could be captured in benchmark problems that could be evaluated much more quickly to compare search algorithms. Unfortunately, no such function characterization method exists, so selecting a CMOEA must be based on less formal understanding of the problem characteristics. The No-Free-Lunch Theorem \citep{585893} establishes that different algorithms are better suited to optimizing different sorts of functions, so it is useful to benchmark algorithms on problems that are as similar as possible to the one which is actually being solved. Since there is no analytical way to do this, it appears that having a set of benchmark functions with tunable levels of difficulty of different sorts is at least a step in a productive direction.

In summary, the contributions of this paper are as follows:
\begin{enumerate}
\item This paper defines three primary types of difficulty for constraints in CMOPs. When designing new constraint handling mechanisms for a CMOEA, one has to investigate the nature of constraints in a CMOP that the CMOEA is aiming to address, including the types and levels of difficulties embedded in the constraints. Therefore, the ability to define arbitrary amounts of multiple types of difficulty for constraints in CMOPs is necessary and desirable.

\item This paper also defines a level of difficulty for each type of difficulty for constraints in the constructed CMOPs, which can be adjusted by users. A difficulty level is uniquely defined by a triplet with each of its parameters specifying the level of one primary difficulty type. Combination of the three primary constraint types with different difficulty triplets can lead to construction of a large variety of CMOPs.

\item Based on the proposed three primary types of difficulty for constraints, nine difficulty-adjustable and scalable CMOPs and CMaOPs, called DAS-CMOP1-9 and DAS-CMaOP1-9, have been constructed.
\end{enumerate}

The remainder of this paper is organized as follows. Section \ref{section:constraint_feature} discusses the effects of constraints on $PFs$. Section \ref{section:difficulty_types} introduces the types and levels of difficulty provided by constraints in CMOPs. Section \ref{section:toolkit} explains the proposed toolkit of construction methods for generating constraints in CMOPs with different types and levels of difficulty. Section \ref{section:scalable_objectives} specifies the scalability of the number of objectives in CMOPs using the proposed toolkit. Section \ref{section:scalable_cmops} generates a set of difficulty-adjustable CMOPs using the proposed toolkit. In Section \ref{section:experimental_study}, the performance of two CMOEAs on DAS-CMOP1-9 with different difficulty levels is compared by experimental studies, and Section \ref{section:conclusion} concludes the paper.

\section {Effects of Constraints on PFs} \label{section:constraint_feature}
\begin{figure*}
\begin{tabular}{ll}
\begin{minipage}[t]{0.3\linewidth}
\includegraphics[width= 5.0cm]{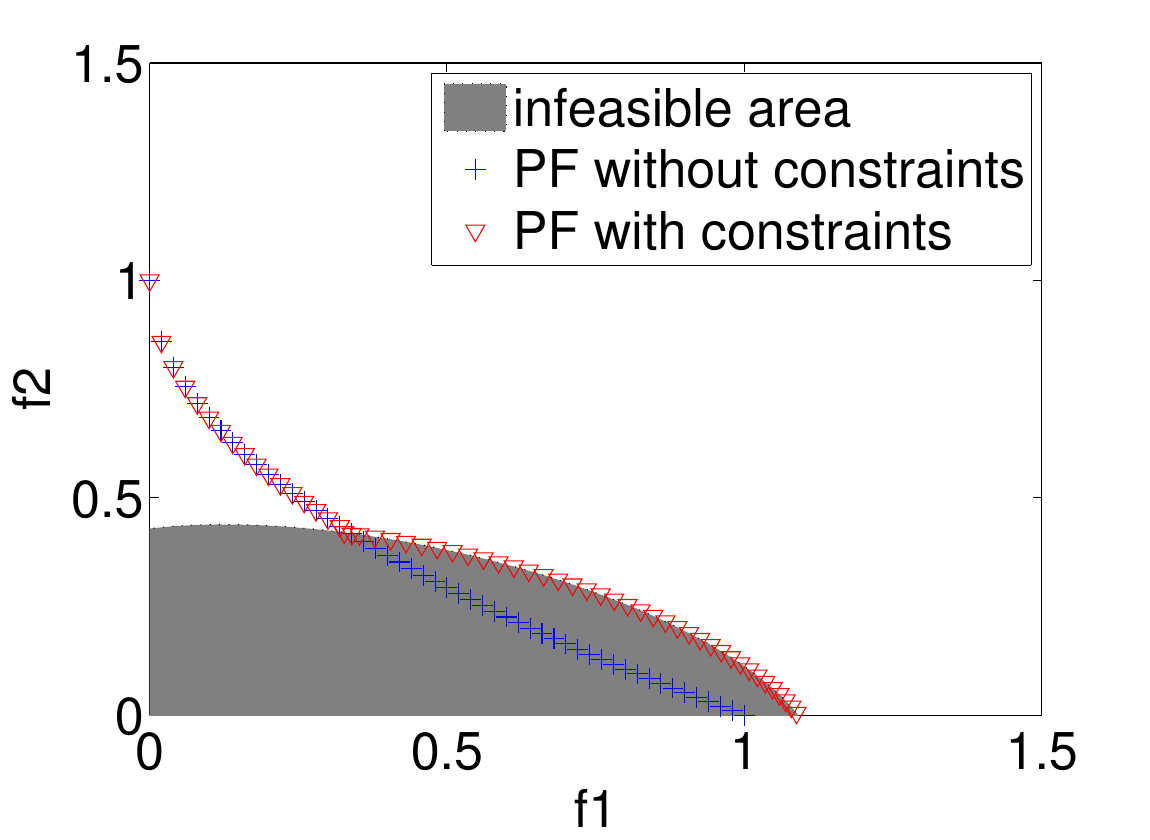}
\centerline{\scriptsize{(a)}}
\label{Fig:pf_a}
\end{minipage}
\hspace{0.25cm}
\begin{minipage}[t]{0.3\linewidth}
\includegraphics[width= 5.0cm]{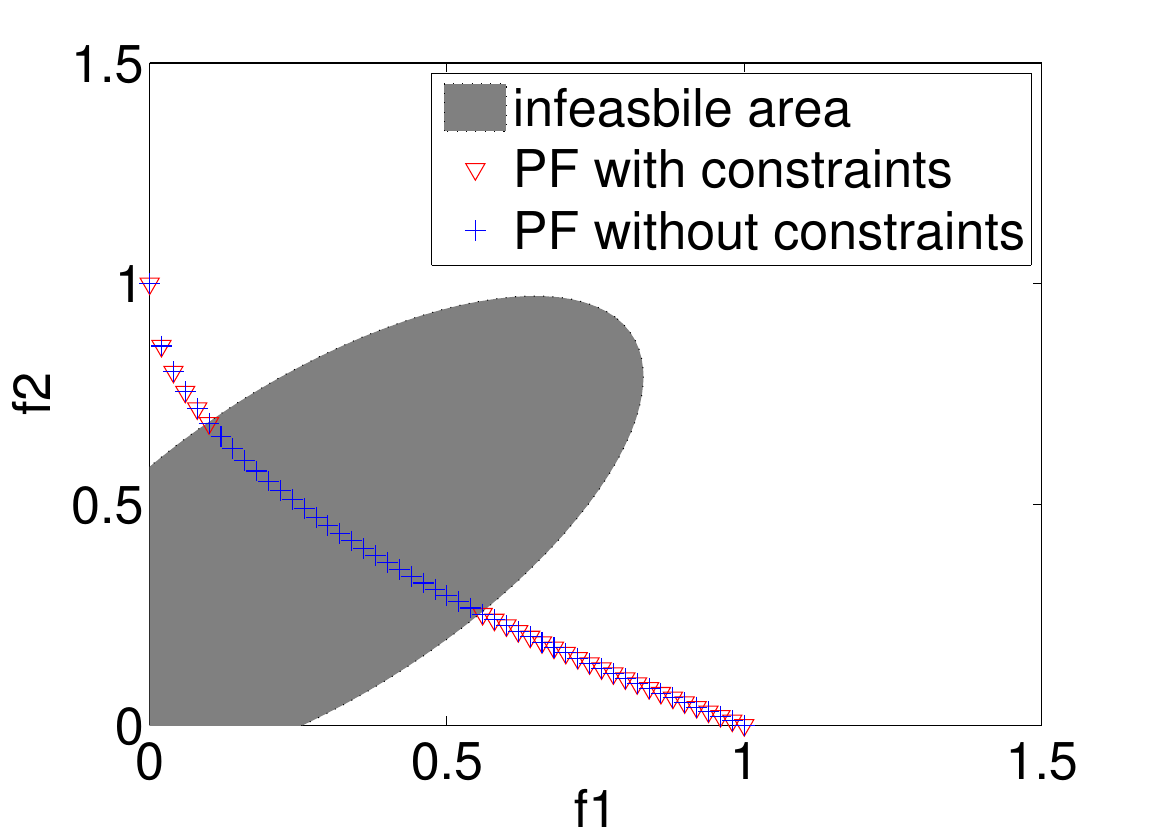}
\centerline{\scriptsize{(b)}}
\label{Fig:pf_b}
\end{minipage}
\hspace{0.25cm}
\begin{minipage}[t]{0.3\linewidth}
\includegraphics[width= 5.0cm]{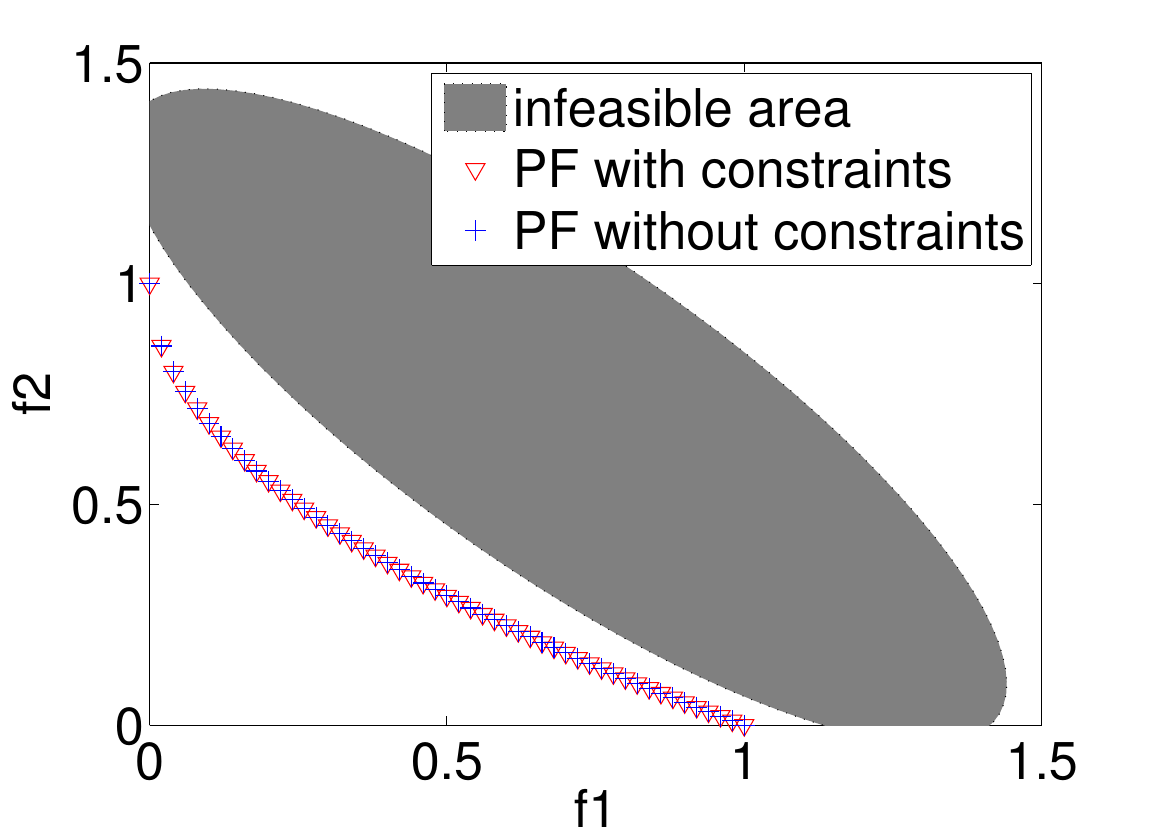}
\centerline{\scriptsize{(c)}}
\label{Fig:pf_c}
\end{minipage}
\end{tabular}

\vspace{-0.3cm}
\begin{tabular}{ll}
\begin{minipage}[t]{0.3\linewidth}
\includegraphics[width= 5.0cm]{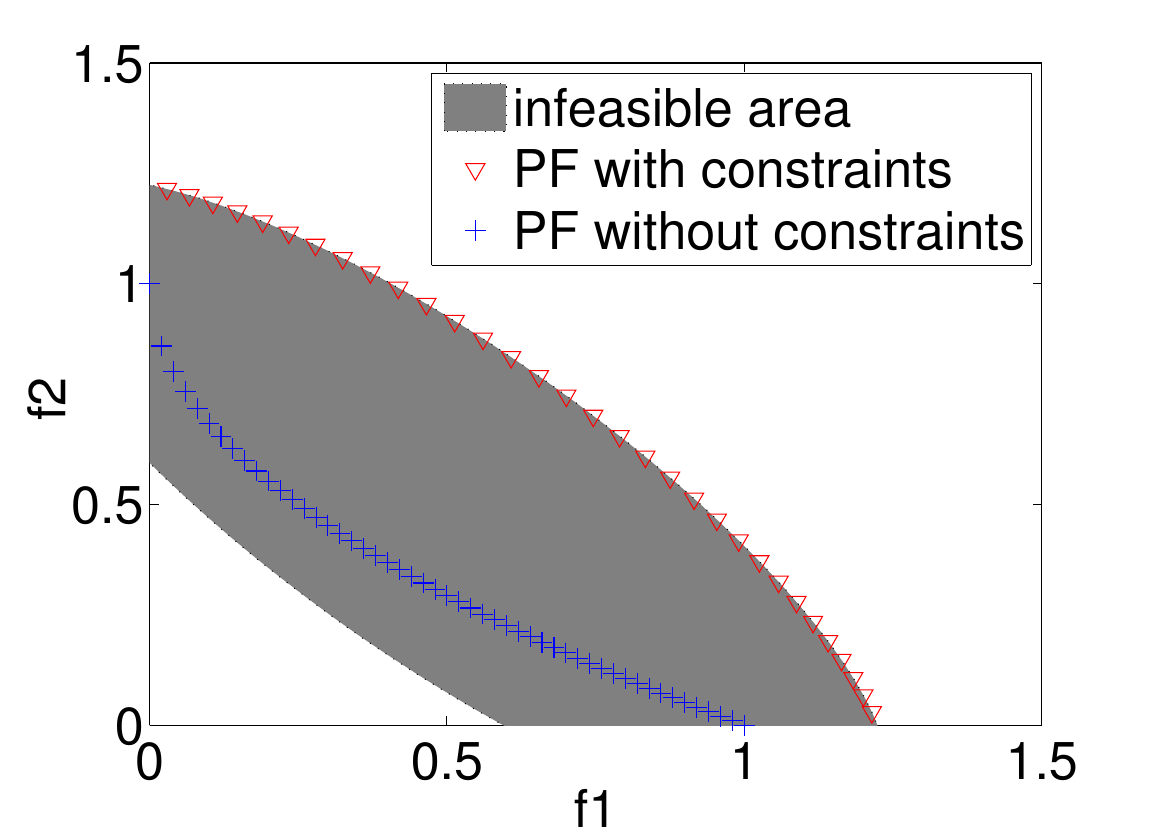}
\centerline{\scriptsize{(d)}}
\label{Fig:pf_d}
\end{minipage}
\hspace{0.25cm}
\begin{minipage}[t]{0.3\linewidth}
\includegraphics[width= 5.0cm]{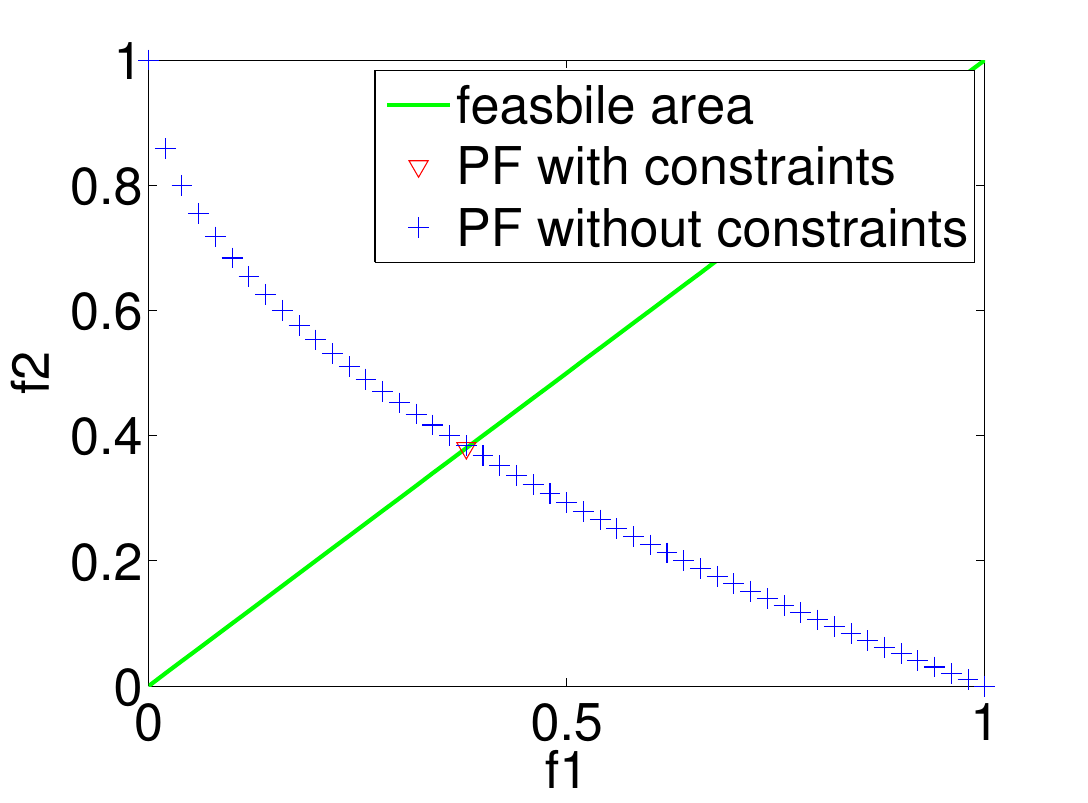}
\centerline{\scriptsize{(e)}}
\label{Fig:pf_e}
\end{minipage}
\end{tabular}
\vspace{-0.6cm}
\caption{Illustration of the effects of constraints on $PFs$. (a) Infeasible regions make the original unconstrained $PF$ partially feasible. Many constrained Pareto optimal solutions lie on constraint boundaries. (b) Infeasible regions make the original unconstrained $PF$ partially feasible. The constrained $PF$ is a portion of its unconstrained $PF$. (c) Infeasible regions block the way toward converging to the $PF$. The constrained $PF$ is same to its unconstrained $PF$. (d) The complete original $PF$ is no longer feasible. Every constrained Pareto optimal solution lies on a constraint boundary. (e) Constraints reduce the dimensionality of the $PF$. A two-objective optimization problem is transformed into a constrained single-objective optimization problem.}
\label{Fig:pf_type}
\end{figure*}

Constraints define the infeasible regions in the search space, leading to different types and levels of difficulty for the resulting CMOPs. Some major possible effects of the constraints on the $PFs$ in CMOPs include the following \citep{jain2014evolutionary}:
\begin{enumerate}
\item Infeasible regions make the original unconstrained $PF$ only partially feasible. This can be further divided into two situations. In the first situation, the $PF$ of the constrained problem consists of a part of its unconstrained $PF$ and a set of solutions on some boundaries of constraints, as illustrated by Fig. \ref{Fig:pf_type}(a). In the second situation, the $PF$ of the constrained problem is only a portion of its unconstrained $PF$, as illustrated by Fig. \ref{Fig:pf_type}(b).
\item Infeasible regions block the way towards the $PF$, as illustrated by Fig. \ref{Fig:pf_type}(c).
\item The complete original $PF$ (of the unconstrained problem) is covered by infeasible regions and is no longer feasible. Every constrained Pareto optimal point lies on a constraint boundary, as illustrated by Fig. \ref{Fig:pf_type}(d).
\item Constraints may reduce the dimensionality of the $PF$, with one example illustrated by Fig. \ref{Fig:pf_type}(e). In general, although the problem is $M-$dimensional, constraints can make the constrained $PF$ $K-$dimensional (where $K<M$). In the particular case of Fig. \ref{Fig:pf_type}(e), $M = 2, K = 1$.
\end{enumerate}

\section{Difficulty Types and Levels of CMOPs} \label{section:difficulty_types}
In this section, three primary difficulty types, including convergence-hardness, diversity-hardness, and feasibility-hardness, are illustrated. The difficulty level for each primary difficulty type is also described.

\subsection{Difficulty 1: Diversity-hardness}
Generally, the $PFs$ of CMOPs with diversity-hardness have many discrete segments, or some parts that are more difficult to achieve than other parts, because large infeasible regions are imposed in their vicinity. As a result, achieving the complete $PF$ is difficult for CMOEAs.

\subsection{Difficulty 2: Feasibility-hardness}
For feasibility-hard CMOPs, the proportion of feasible regions in the search space is usually very low. It is difficult for a CMOEA to find any feasible solutions on CMOPs with feasibility-hardness. Often in the initial stage of a CMOEA, most or all solutions in the population are infeasible.

\subsection{Difficulty 3: Convergence-hardness}
CMOPs with convergence-hardness hinder the convergence of CMOEAs towards the $PFs$. Usually, CMOEAs encounter more difficulty in achieving the $PFs$ because infeasible regions block the way as they converge toward the $PFs$. In other words, the generational distance (GD) metric \citep{van1998evolutionary}, which indicates the convergence performance, is difficult to minimize in the evolutionary process.

\subsection{Difficulty Level of Each Primary Difficulty Type}
\begin{figure*}
\centering
\includegraphics[height= 4.5cm]{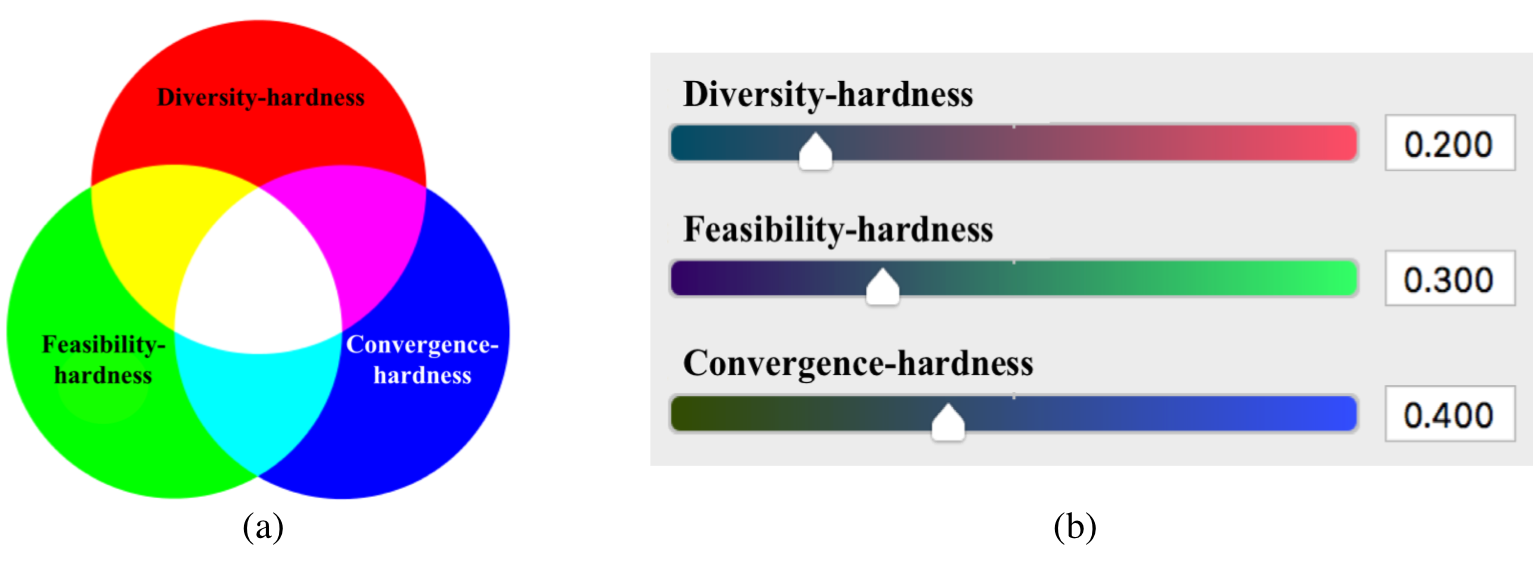}
\vspace{-0.5cm}
\caption{Illustration of difficulty types and levels. (a) The three primary difficulty types and their combinations resulting in seven basic difficulty types (as shown in Table \ref{Table:difficulty_types}), using the analogy with three primary colors and their combinations forming seven basic colors. (b) Combining three parameterized constraint functions using a triplet composed of three parameters. The three primary constraint functions correspond to the three primary difficulty types.}
\label{Fig:primaryType}
\end{figure*}

A difficulty level of each primary difficulty type can be defined by a parameter in the parameterized constraint function corresponding to the primary difficulty type. Each parameter is normalized from 0 to 1. Three parameters, corresponding to the difficulty levels of the three primary difficulty types, form a triplet $(\eta,\zeta,\gamma)$ that exactly defines the difficulty signature of a CMOP constructed by the three parameterized constraint functions. If we allow the three parameters to take any value between 0 and 1, then we can literally get countless varieties of difficulty (analogous to countless colors in the color space, as shown in Fig. \ref{Fig:primaryType}). A difficulty signature is then precisely depicted by a triplet $(\eta,\zeta,\gamma)$.

\begin{table*}
\centering
\caption{Basic difficulty types of CMOPs}
\tabcolsep 0.05in
\label{Table:difficulty_types}
\scalebox{0.64}[0.64]{
\begin{tabular}{|l||l|}
\hline
Basic Difficulty Types& Comment\\
\hline
T1: Diversity-hardness   & Distributing the feasible solutions in the complete $PF$ is difficult.\\
T2: Feasibility-hardness & Obtaining a feasible solution is difficult.\\
T3: Convergence-hardness & Approaching a Pareto optimal solution is difficult.\\
\hline
T4: Diversity-hardness and feasibility-hardness & Obtaining a feasible solution and the complete $PF$ is difficult.\\
T5: Diversity-hardness and convergence-hardness & Approaching a Pareto optimal solution and the complete $PF$ is difficult.\\
T6: Feasibility-hardness and convergence-hardness & Obtaining a feasible solution and approaching a Pareto optimal solution is difficult.\\
\hline
T7: Diversity-hardness, feasibility-hardness and convergence-hardness & Obtaining a Pareto optimal solution and the complete $PF$ is difficult.\\
\hline
\end{tabular}
}
\end{table*}

\section{Construction Toolkit}\label{section:toolkit}

As we know, constructing a CMOP involves constructing two major parts --- objective functions and constraint functions. \cite{li2014multiobjective} suggested a general framework for constructing objective functions for benchmark problems. It is stated as follows:
\begin{eqnarray}
\label{Form:obj}
&f_i(\mathbf{x}) = \alpha_{i}(x_{1 : m - 1}) + \beta_ {i}(x_{1 : m - 1},x_{m : n})
\end{eqnarray}
where $x_{1 : m-1} = (x_1,\ldots,x_{m-1}) ^{T}, x_{m : n} = (x_m,\ldots,x_n)^{T}$ are two sub-vectors of $x = (x_1,\ldots,x_n)^T$. The function $\alpha_{i}(x_{1 : m-1})$ is called the shape function, and  $\beta_{i}(x_{1 : m-1},x_{m : n})$ is called the nonnegative distance function. The objective function $f_i(x), i=1,\ldots,m$ is the sum of the shape function $\alpha_{i}(x_{1 : m-1})$ and the nonnegative distance function $\beta_{i}(x_{1 : m-1},x_{m : n})$. We adopt the method of \citep{li2014multiobjective} to construct objective functions in this work.

In terms of constructing the constraint functions, three different types of constraint functions are suggested in this paper, corresponding to the three proposed primary types of difficulties of CMOPs. More specifically, Type-I constraint functions provide difficulty of diversity-hardness, Type-II constraint functions introduce difficulty of feasibility-hardness, and Type-III constraint functions generate difficulty of convergence-hardness. The detailed definitions of the three types of constraint functions are given as follows:

\subsection{Type-I Constraint Functions: Diversity-hardness}

Type-I constraint functions are defined to limit the boundaries of sub-vector $x_{1 : m-1}$. More specifically, this type of constraint function divides the $PF$ of a CMOP into a number of disconnected segments, generating difficulty of diversity-hardness. Here, we use a parameter $\eta\in[0,1]$ to represent the level of difficulty. $\eta = 0$ means the constraint functions impose no effect on the CMOP, while $\eta = 1$ means the constraint functions provide their maximum effect.

\begin{figure*}
\begin{tabular}{cc}
\hspace{-0.5cm}
\begin{minipage}[t]{0.25\linewidth}
\includegraphics[width= 4.2cm]{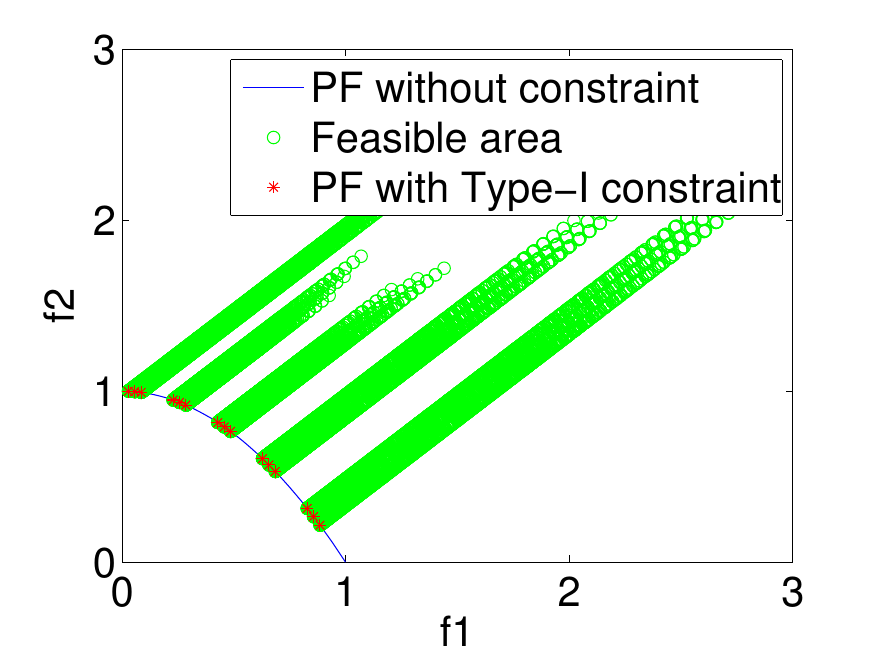}
\centerline{\scriptsize{(a) $\eta = 0.55$}}
\label{type-1-a}
\end{minipage}
\begin{minipage}[t]{0.25\linewidth}
\includegraphics[width= 4.2cm]{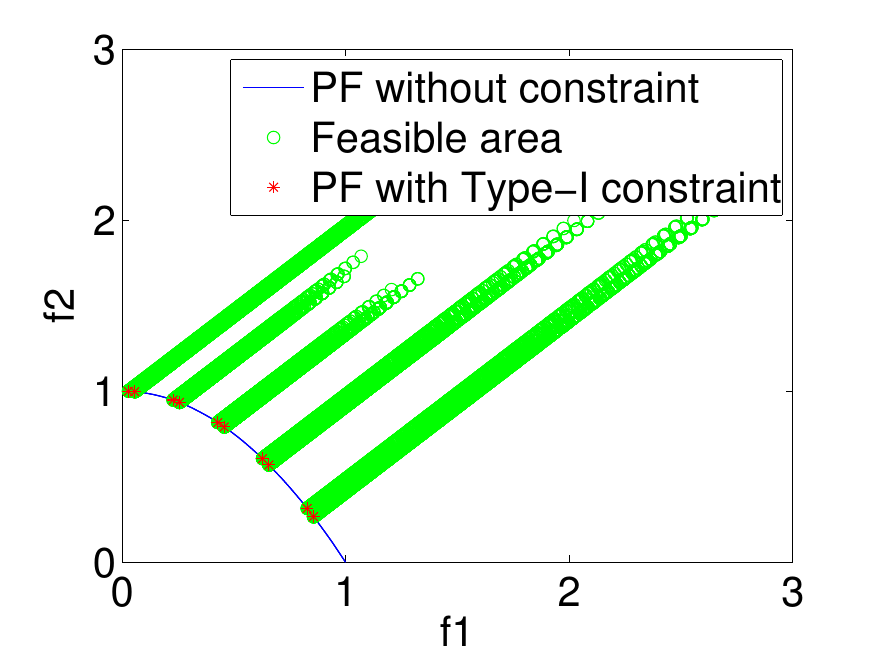}
\centerline{\scriptsize{(b) $\eta = 0.75$}}
\label{type-1-b}
\end{minipage}
\begin{minipage}[t]{0.25\linewidth}
\includegraphics[width= 4.2cm]{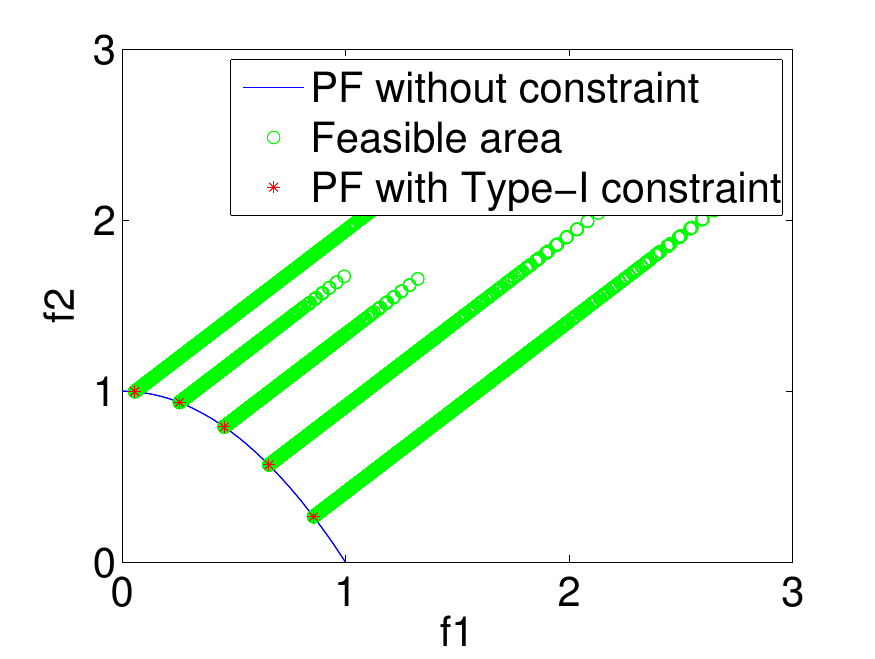}
\centerline{\scriptsize{(c) $\eta = 1.0$}}
\label{type-1-c}
\end{minipage}
\begin{minipage}[t]{0.25\linewidth}
\includegraphics[width= 4.2cm]{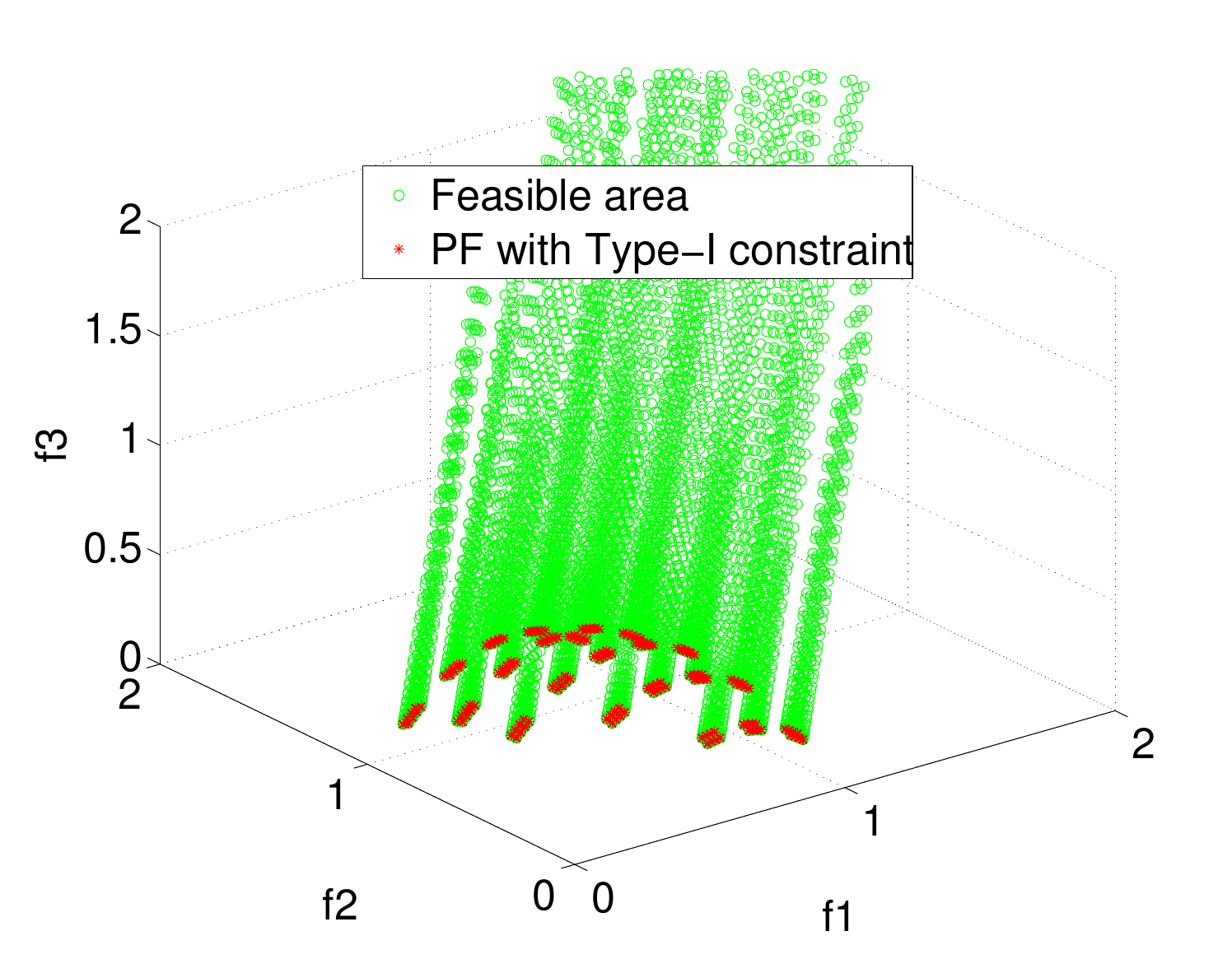}
\centerline{\scriptsize{(d) $m = 3, \eta = 0.75$}}
\label{type-1-d}
\end{minipage}
\end{tabular}
\vspace{-0.5cm}
\caption{Illustrations of the influence of Type-I constraint functions. When the parameter of difficulty level $\eta$ increases, the width of each segment in the $PF$ decreases, and the difficulty level of a CMOP increases. The $PF$ of a CMOP with Type-I constraint functions is disconnected and usually has many discrete segments, so obtaining the complete $PF$ is difficult. Thus a CMOP with Type-I constraint functions is diversity-hard. (a) A two-objective CMOP with $\eta = 0.55$. (b) A two-objective CMOP with $\eta = 0.75$. (c) A two-objective CMOP with $\eta = 1.0$. (d) A three-objective CMOP with $\eta = 0.75$.} \label{Fig:pf_type1}
\end{figure*}

An example of a CMOP with diversity-hardness is suggested as follows:
\begin{eqnarray}
\label{Form:example_constraint_1}
\begin{cases} \text{minimize} & f_1(x) = x_1 + g(x) \\
\text{minimize} & f_2(x) = 1 - x_1^2 + g(x) \\
& g(x) = \sum_{i = 2} ^ {n} {(x_i  - \sin(0.5 \pi x_1))} ^{2} \\
\text{subject to} & c(x) = \sin(a\pi x_1) - b \ge 0 \\
& x_i \in [0,1]
\end{cases}
\end{eqnarray}
where $a > 0$, $b \in [-1,1]$. As an example, $a=10$ and $n=2$ are set here. The parameter $\eta$ specifying the level of difficulty determines $b = 2\eta - 1$. The number of disconnected segments in the $PF$ is controlled by $a$. Moreover, the value of $b$ controls the width of each segment. The width of each segment is at its maximum when $b = -1$ (and $\eta = 0$). When $b$ increases, the width of each segment decreases, and the difficulty level increases, and so does the parameter of the difficulty level $\eta$. As a result, if $\eta$ is set to $0.55$, the $PF$ is shown in Fig. \ref{Fig:pf_type1}(a). If $\eta = 0.75$, the $PF$ appears as shown in Fig. \ref{Fig:pf_type1}(b). It can be observed that the width of each segment of the $PF$ is reduced as $\eta$ keeps increasing. If $\eta = 1.0$, the width of each segment shrinks to zero as shown in Fig. \ref{Fig:pf_type1}(c), which provides the maximum level of difficulty to the CMOP. The $PF$ of a three-objective CMOP with Type-I constraint functions is also shown in Fig. \ref{Fig:pf_type1}(d), with the difficult level $\eta = 0.75$. It can be seen that Type-I constraint functions can be applied in more than two-objective CMOPs, which means that a CMOP with a scalable number of objectives can be constructed using this type of constraint functions.

\subsection{Type-II Constraint Functions: Feasibility-hardness}
Type-II constraint functions are set to limit the reachable boundaries of the distance function of $\beta_{i}(x_{1 : m-1},x_{m : n})$, and thereby control the proportion of the search space that is feasible. That is, they generate the difficulty of feasibility-hardness. Here, we use a parameter $\zeta$ to represent the level of difficulty, which ranges from 0 to 1. $\zeta = 0$ means the constraints are the weakest, and $\zeta=1.0$ means the constraint functions are the strongest.

\begin{figure*}
\begin{tabular}{cc}
\begin{minipage}[t]{0.25\linewidth}
\includegraphics[width= 4.2cm]{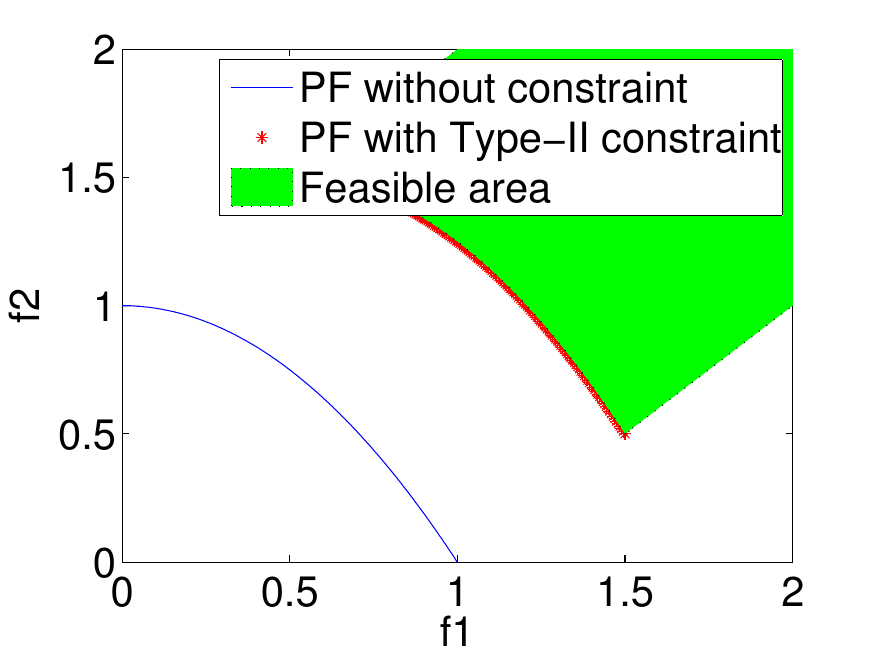}
\centerline{\footnotesize{(a) $\zeta = 0$}}
\label{type-2-a}
\end{minipage}
\begin{minipage}[t]{0.25\linewidth}
\includegraphics[width= 4.2cm]{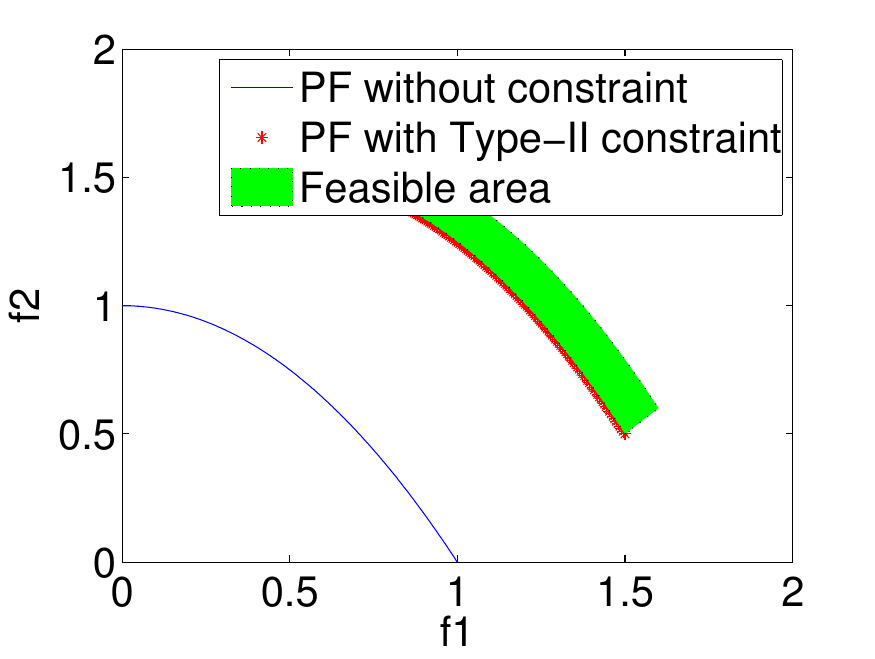}
\centerline{\footnotesize{(b) $\zeta = 0.905$}}
\label{type-2-b}
\end{minipage}
\begin{minipage}[t]{0.25\linewidth}
\includegraphics[width= 4.2cm]{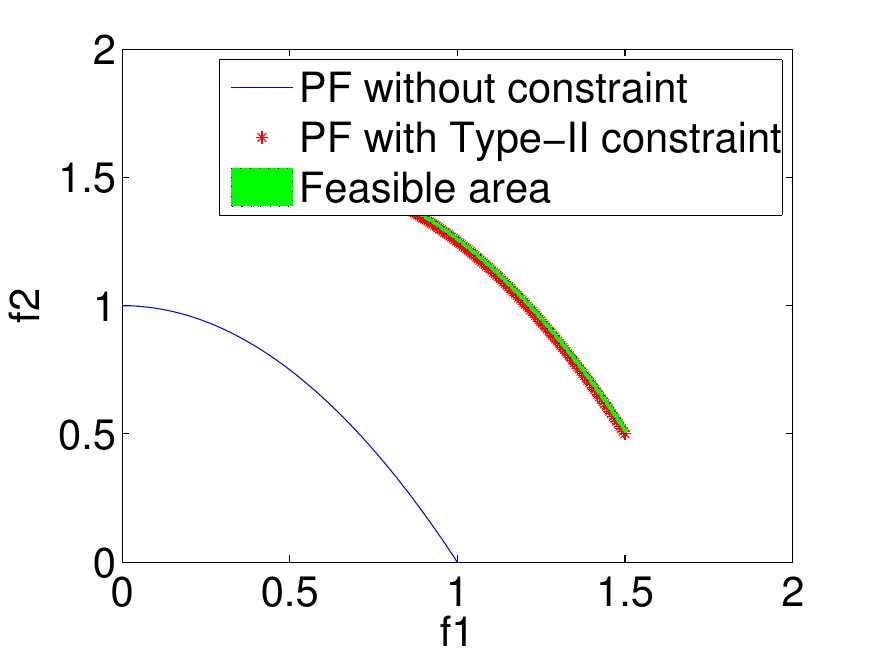}
\centerline{\footnotesize{(c) $\zeta = 1.0$}}
\label{type-2-c}
\end{minipage}
\begin{minipage}[t]{0.25\linewidth}
\includegraphics[width= 4.2cm]{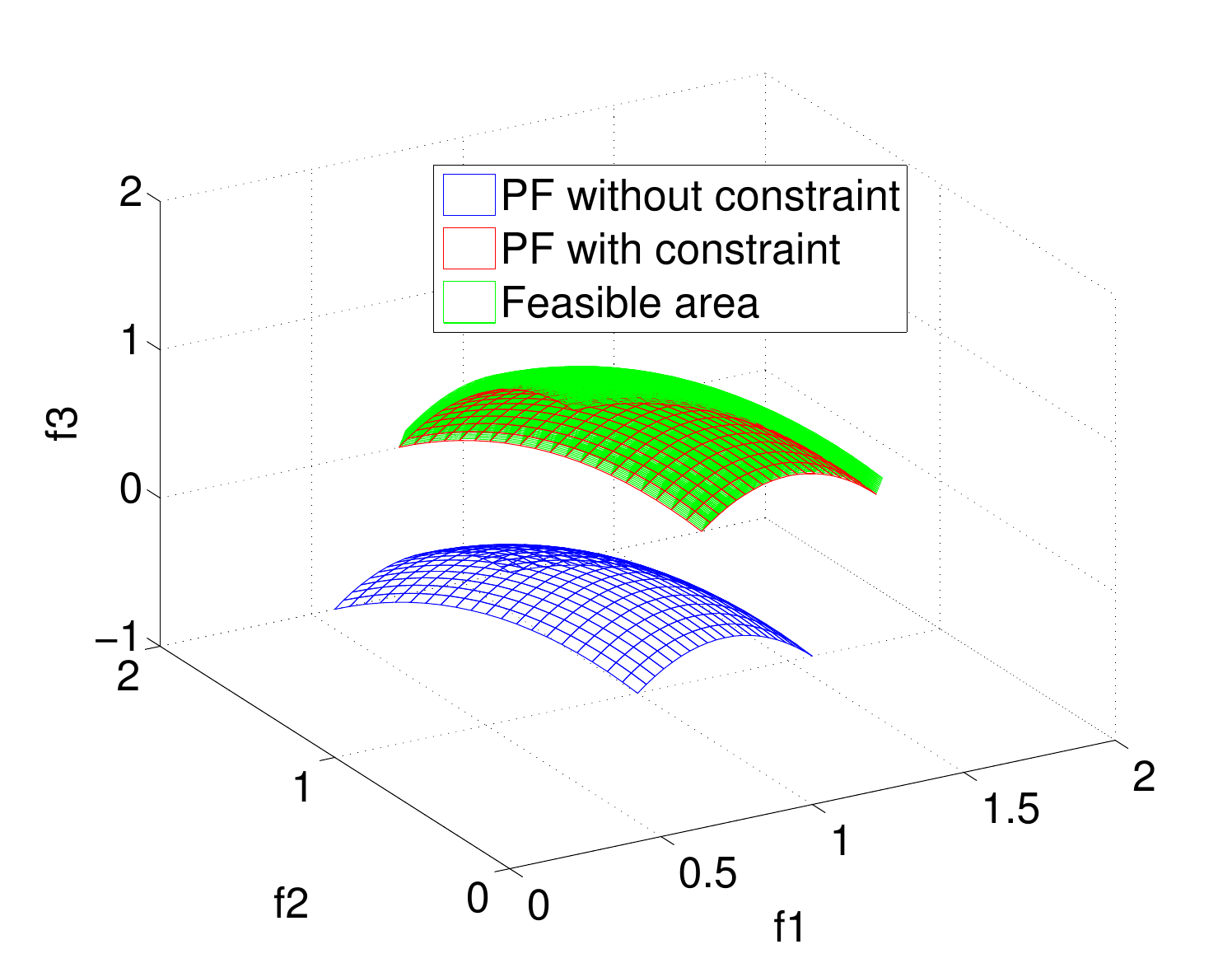}
\centerline{\footnotesize{(d) $m = 3, \zeta = 0.905$}}
\label{type-2-d}
\end{minipage}
\end{tabular}
\vspace{-0.5cm}
\caption{Illustrations on the influence of Type-II constraint functions. The parameter of difficulty degree $\zeta = \frac{1}{\exp(b-a)}$ determines the proportion of feasible regions. Here, the value of $a$, the offset of the constrained $PF$ from the unconstrained $PF$, is defaulted to 0.5. When parameter $\zeta$ increases, the proportion of feasible area decreases, and the difficulty level of feasibility increases. (a) $\zeta = 0.0$. (b) $\zeta = 0.905$. (c) $\zeta = 1.0$. (d) Type-II constraint functions can also be applied in three-objective optimization problems, as shown with $\zeta = 0.905$.} \label{Fig:pf_type2}
\end{figure*}

For example, a CMOP with Type-II constraint functions can be defined as follows:
\begin{eqnarray}
\label{Form:example_constraint_2}
\begin{cases} \text{minimize} & f_1(x) = x_1 + g(x) \\
\text{minimize} & f_2(x) = 1 - x_1^2 + g(x) \\
& g(x) = \sum_{i = 2} ^ {n} {(x_i  - \sin(0.5 \pi x_1))} ^{2} \\
\text{subject to} & c_1(x) = g(x) - a \ge 0 \\
& c_2(x) = b - g(x) \ge 0 \\
& n = 30, x_i \in [0,1]
\end{cases}
\end{eqnarray}
where $\zeta$ equals to $\frac{1}{\exp(b-a)}$, $a \ge 0$, $b \ge 0$ and $b \ge a$. The distance between the constrained $PF$ and unconstrained $PF$ is controlled by $a$, and $a = 0.5$ in this example. The proportion of feasible regions is controlled by $b - a$. For an arbitrary $a$, $b$ must be set to $a - \ln\zeta$ to achieve a hardness in (0,1]. As $b-a$ approaches  $+ \infty$, $\zeta$ approaches $0$, and the feasible area grows, as shown in Fig. \ref{Fig:pf_type2}(a). If $\zeta = 0.905$, $b-a = 0.1$, and the feasible area is decreased as shown in Fig. \ref{Fig:pf_type2}(b). For $\zeta = 1.0$, $b = a$, and the feasible area in the objective space is very small, which can be observed in Fig. \ref{Fig:pf_type2}(c). Type-II constraints can be also applied to CMOPs with three objectives, as shown in Fig. \ref{Fig:pf_type2}(d).

\subsection{Type-III Constraint Functions: Convergence-hardness}
Type-III constraint functions limit the reachable boundary of objectives. As a result, infeasible regions act like 'blocking' obstacles for the population of a CMOEA searching for the $PF$. As a result, Type-III constraint functions generate the difficulty of convergence-hardness. Here, we use a parameter $\gamma$ to represent the level of difficulty, which ranges from 0 to 1. $\gamma = 0$ means the constraints are the weakest, $\gamma = 1$ means the constraints are the strongest, and the difficulty level increases as $\gamma$ increases.

\begin{figure*}
\begin{tabular}{cc}
\hspace{-0.5cm}
\begin{minipage}[t]{0.25\linewidth}
\includegraphics[width= 4cm]{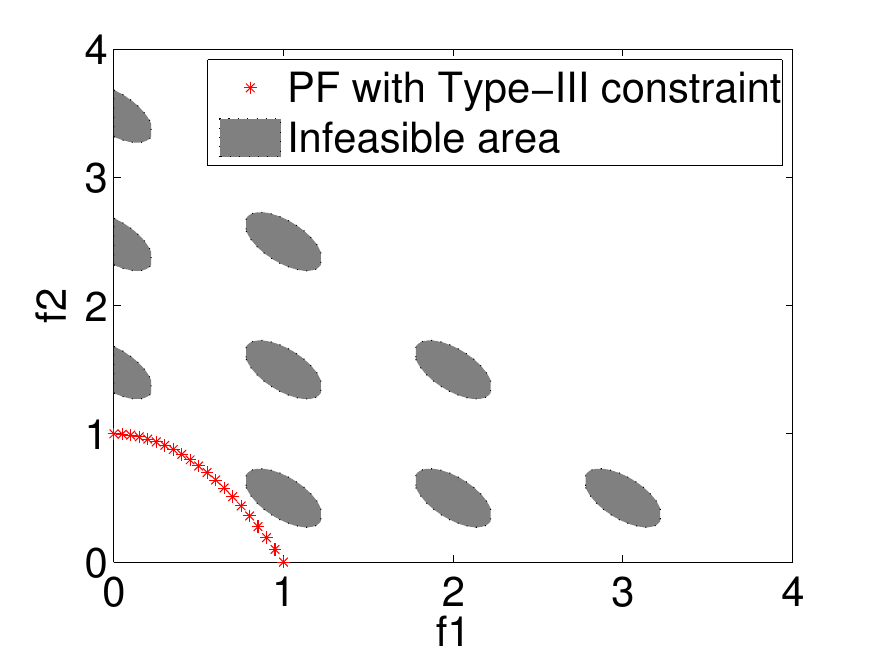}
\centerline{\footnotesize{(a) $\gamma = 0.1$ }}
\label{type3-a}
\end{minipage}
\begin{minipage}[t]{0.25\linewidth}
\includegraphics[width= 4cm]{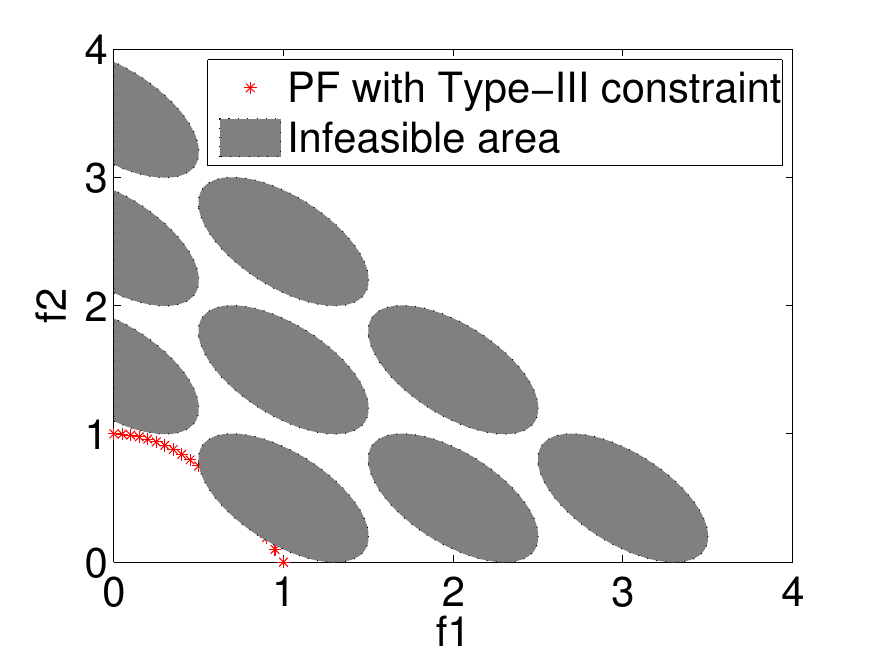}
\centerline{\footnotesize{(b) $\gamma = 0.5$}}
\label{type3-b}
\end{minipage}
\begin{minipage}[t]{0.25\linewidth}
\includegraphics[width= 4cm]{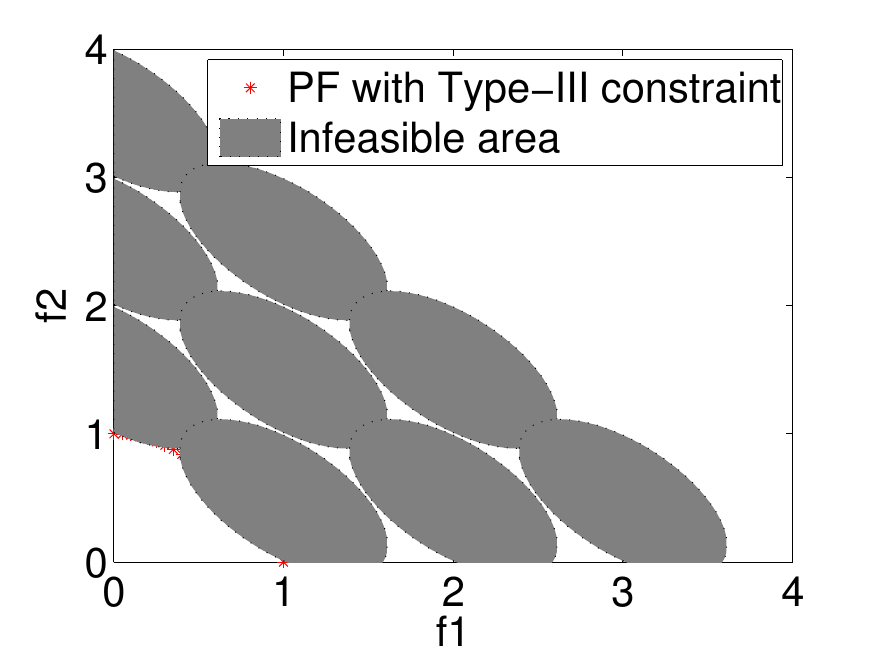}
\centerline{\footnotesize{(c) $\gamma = 0.75$}}
\label{type3-c}
\end{minipage}
\begin{minipage}[t]{0.25\linewidth}
\includegraphics[width= 4cm]{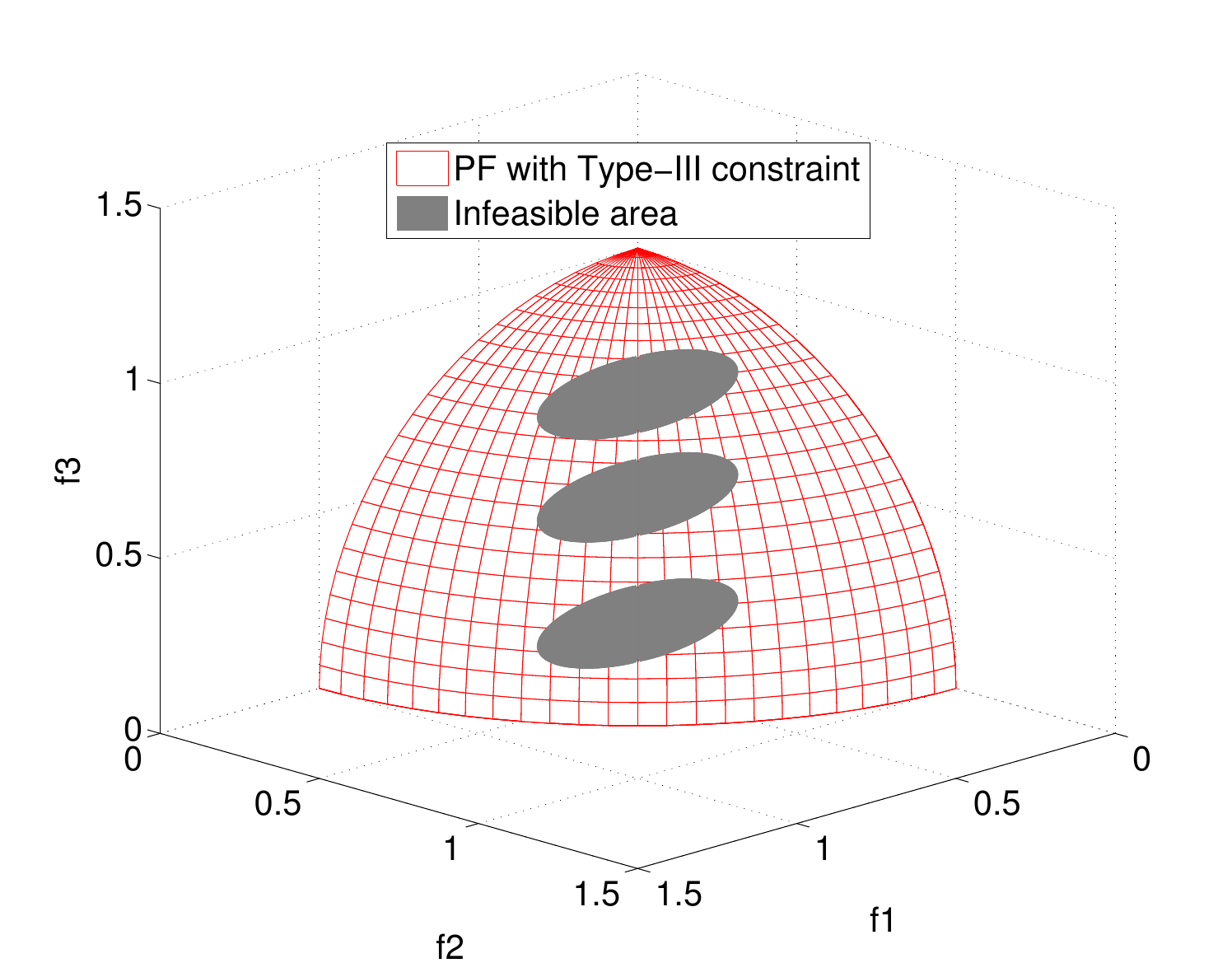}
\centerline{\footnotesize{(d) $\gamma = 0.5, m = 3$}}
\label{type3-d}
\end{minipage}
\end{tabular}
\vspace{-0.5cm}
\caption{Illustrations of the influence of Type-III constraint functions. Infeasible regions block the way of converging to the $PF$. The gray parts of each figure are infeasible regions. A parameter $\gamma$ is adopted to represent the level of difficulty, which ranges from 0 to 1. $\gamma = 0$ means the constraints are the weakest, and $\gamma = 1$ means the constraints are the strongest. When $\gamma$ increases, the difficulty level of convergence-hardness of a CMOP increases. (a) $\gamma = 0.1$. (b) $\gamma = 0.5$. (c) $\gamma = 0.75$. (d) Type-III constraint functions can also be applied in three-objective optimization problems, as shown here with $\gamma = 0.5$.} \label{Fig:pf_type3}
\end{figure*}

For example, a CMOP with Type-III constraint functions can be defined as follows:
\begin{eqnarray}
\label{Form:example_constraint_3}
\begin{cases} \text{min} & f_1(x) = x_1 + g(x) \\
\text{min} & f_2(x) = 1 - x_1^2 + g(x) \\
\text{where}& g(x) = \sum_{i = 2} ^ {n} {(x_i  - \sin(0.5 \pi x_1))} ^{2} \\
\text{s.t.} & c_k(x) = ((f_1 - p_k) \cos\theta_k - ( f_2 - q_k) \sin\theta_k)^2 / a_k^2 \\
& + ((f_1 - p_k) \sin\theta_k + (f_2 - q_k) \cos\theta_k)^2 / b_k ^{2} \ge r \\
& p_k=[0,1,0,1,2,0,1,2,3]\\
& q_k=[1.5,0.5,2.5,1.5,0.5,3.5,2.5,1.5,0.5]\\
& a_k^{2} = 0.4, b_k^{2} =1.6, \theta_k = -0.25\pi \\
& c = 20, n = 30, x_i \in [0,1], k= 1,\ldots,9\\
\end{cases}
\end{eqnarray}
where the level of difficulty parameter $\gamma$ determines parameter $r$ as $r = \gamma/2$. If $\gamma = 0.1$, the $PF$ is shown in Fig. \ref{Fig:pf_type3}(a). If $\gamma = 0.5$, the infeasible regions are increased and shown in Fig. \ref{Fig:pf_type3}(b). If $\gamma = 0.75$, the infeasible regions become bigger than those of $\gamma = 0.5$ as shown in Fig. \ref{Fig:pf_type3}(c). Type-III constraint functions can be also applied to CMOPs with three objectives, as shown in Fig. \ref{Fig:pf_type3}(d).

Type-III constraint functions can be expressed in a matrix form, which can be defined as follows:
\begin{eqnarray}
\label{Form:matrix_form}
&(F(\mathbf{x}) - H_k)^{T} S_{k} (F(\mathbf{x}) - H_k) \ge r
\end{eqnarray}
where $F(\mathbf{x}) = (f_1(\mathbf{x}),\ldots,f_m(\mathbf{x}))^{T}$. $H_k$ is a translation vector. $S_{k}$ is a transformation matrix that controls the degree of rotation and stretching of the vector $(F(\mathbf{x}) - H_k)$. According to the Type-III constraint functions in Eq. \eqref{Form:example_constraint_3}, $H_k = (p_k,q_k)^T$, and $S_k$ can be expressed as follows:

\begin{equation*}
S_k=\begin{bmatrix}
\frac{\cos^2\theta_k}{a_k^2} + \frac{\sin^2\theta_k}{b_k^2} & -\frac{\sin 2 \theta_k}{a_k^2}\\
\frac{\sin 2 \theta_k}{b_k^2} & \frac{\cos^2\theta_k}{b_k^2} + \frac{\sin^2\theta_k}{a_k^2} \\
\end{bmatrix}
\end{equation*}

To summarize, the three types of constraint functions discussed above correspond to the three primary difficulty types of CMOPs. The level of each primary difficulty type can be decided by a parameter. In this work, the three parameters are defined in a triplet $(\eta,\zeta,\gamma)$, which specifies the difficulty levels of the various primary types of difficulty. This approach of constructing a toolkit for CMOPs can also be scaled to generate CMOPs with more than three objective functions. The scalability to the number of objectives is discussed in more detail next.

\section{Scalability of the Number of Objectives} \label{section:scalable_objectives}
Many-objective optimization has attracted much research interest in recent years, which makes the scalability of the number of objectives of CMOPs a desirable feature. A general framework to construct CMOPs with a scalable number of objectives is given as follows:
\begin{eqnarray}
\label{eqn:das-cmop}
\begin{cases}
\text{min} & f_{1}(\mathbf{x}) = \alpha_{1}(x_{1:m-1}) + \beta_{1}(x_{1:m-1},x_{m:n})\\
\text{min}& f_{2}(\mathbf{x}) = \alpha_{2}(x_{1:m-1}) + \beta_{2}(x_{1:m-1},x_{m:n})\\
& \ \ \ \vdots\\
\text{min}& f_{m}(\mathbf{x}) = \alpha_{m}(x_{1:m-1}) + \beta_{m}(x_{1:m-1},x_{m:n})\\
\text{s.t} & c_k(\mathbf{x}) = \sin(a\pi x_k) - b \ge 0, \text{if } k \text{ is odd} \\
&c_k(\mathbf{x}) = \cos(a\pi x_k) - b \ge 0, \text{if } k \text{ is even} \\
&c_{K+p}(\mathbf{x}) = (e - \beta_{p})(\beta_{p} - d) \ge 0\\
&c_{K+P+q}(\mathbf{x}) = (F(\mathbf{x}) - H_q)^{T} S_{q} (F(\mathbf{x}) - H_q) \ge r\\
&k = 1,\ldots,K, \text{and } K \le m - 1\\
&p = 1,\ldots,P, \text{and } P \le m\\
&q = 1,\ldots,Q\\
&\eta = (b+1)/2\\
&\zeta = \exp(d - e), d \le e, \text{if }\zeta == 0, d = 0\\
&\gamma = 2r\\
\end{cases}
\end{eqnarray}

In Eq. \eqref{eqn:das-cmop}, we borrow an idea from the WFG toolkit \citep{huband2006review} to construct objectives, which can be scaled to any number of objectives. More specifically, the number of objectives is controlled by a user-defined parameter $m$.

The three different types of constraint functions proposed in Section \ref{section:scalable_objectives} can be combined with the scalable objective functions to construct difficulty-adjustable and scalable CMOPs (DAS-CMOPs). More specifically, the first $K$ constraint functions of Type-I are defined to limit the reachable boundary of each decision variable in the shape functions ($\alpha_{1}(x_{1:m-1})$ to $\alpha_{m}(x_{1:m-1})$), which have the ability to control the difficulty level of diversity-hardness using $\eta$. The $(K+1)$ to $(K+P)$ constraint functions belong to Type-II, which limit the reachable boundary of the distance functions ($\beta_{1}(x_{1:m-1},x_{m:n})$ to $\beta_{1}(x_{1:m-1},x_{m:n})$). They have the ability to control the difficulty level of feasibility-hardness using $\zeta$. The last $Q$ constraint functions are set directly on each objective, and belong to Type-III. They generate a number of infeasible regions, which hinder the working population of a CMOEA as it approaches the $PF$. The difficulty level of convergence-hardness generated by Type-III constraint functions is controlled by $\gamma$. The other parameters in Eq. \eqref{eqn:das-cmop} are illustrated as follows.

Three parameters---$K$, $P$ and $Q$---are used to control the number of each type of constraint function. $K \le m - 1, P \le m$ and $Q \ge 1$. The total number of constraint functions is controlled by $(K + P + Q)$. $n$ determines the dimension of decision variables, and $n \ge m$. $a$ decides the number of disconnected segments in a $PF$. $d$ indicates the distance between the constrained $PF$ and the unconstrained $PF$.

It is worth noting that the number of objectives of DAS-CMOPs can be easily scaled by tuning parameter $m$. The difficulty level of DAS-CMOPs can be also easily adjusted by assigning a difficulty triplet $(\eta, \zeta, \gamma)$ with three parameters ranging from 0 to 1.

\section{A Set of Difficulty-adjustable and Scalable CMOPs} \label{section:scalable_cmops}
In this section, as an example, a set of nine difficulty-adjustable and scalable CMOPs (DAS-CMOP1-9) and a set of nine difficulty-adjustable and scalable CMaOPs (DAS-CMaOP1-9) are suggested using the proposed toolkit.

As mentioned in Section \ref{section:toolkit}, constructing a CMOP includes constructing both objective functions and constraint functions. According to Eq. \eqref{eqn:das-cmop}, we suggest nine multi-objective functions, including convex, concave and discrete $PF$ shapes, to construct CMOPs. A set of difficulty-adjustable constraint functions is generated by Eq. \eqref {eqn:das-cmop}. Nine difficulty-adjustable and scalable CMOPs, called DAS-CMOP1-9, are generated by combining the suggested objective functions and the generated constraint functions. The detailed definitions of DAS-CMOP1-9 are given in Table \ref{tab:das-cmops}.

\begin{figure*}
\begin{tabular}{cc}
\begin{minipage}[t]{0.28\linewidth}
\includegraphics[width = 5cm]{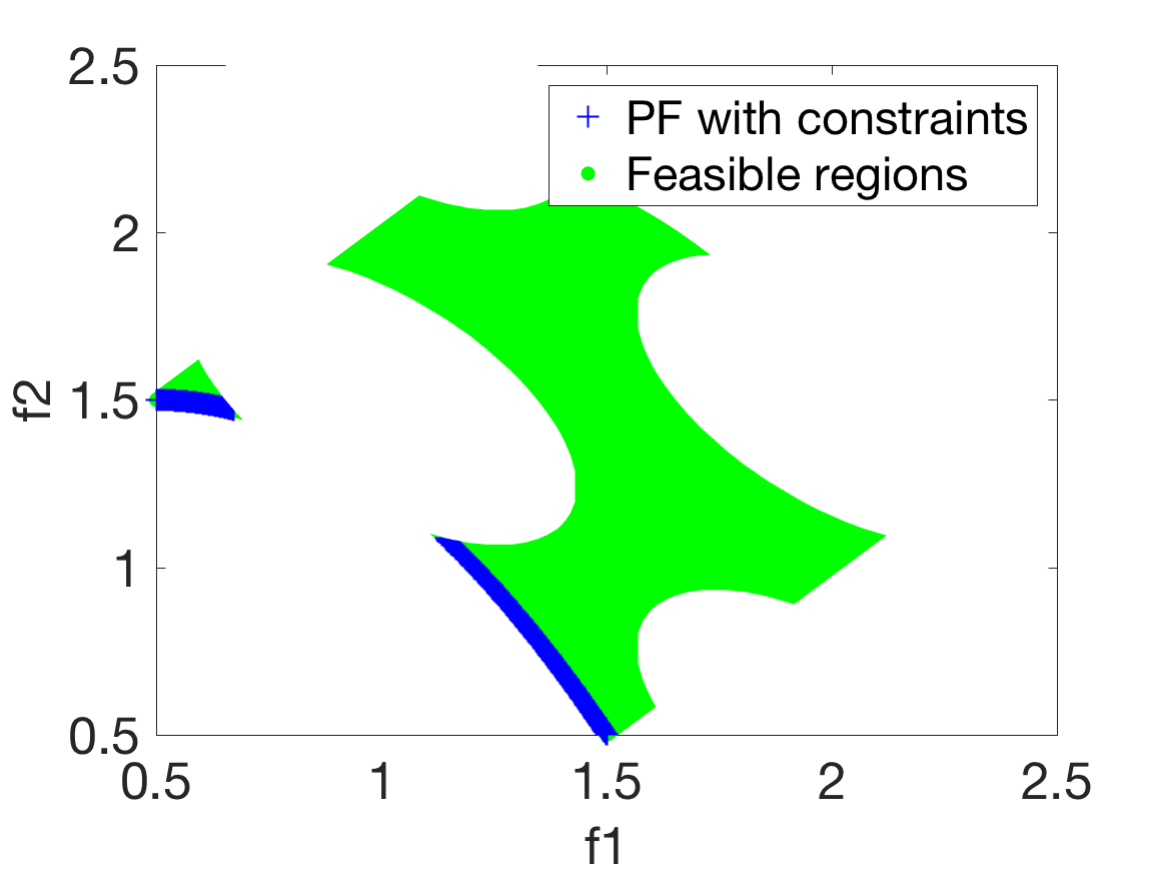}\\
\centering{\scriptsize{(a) DAS-CMOP1 (0,0.5,0.5)}}
\end{minipage}
\hspace{0.5cm}
\begin{minipage}[t]{0.28\linewidth}
\includegraphics[width = 5cm]{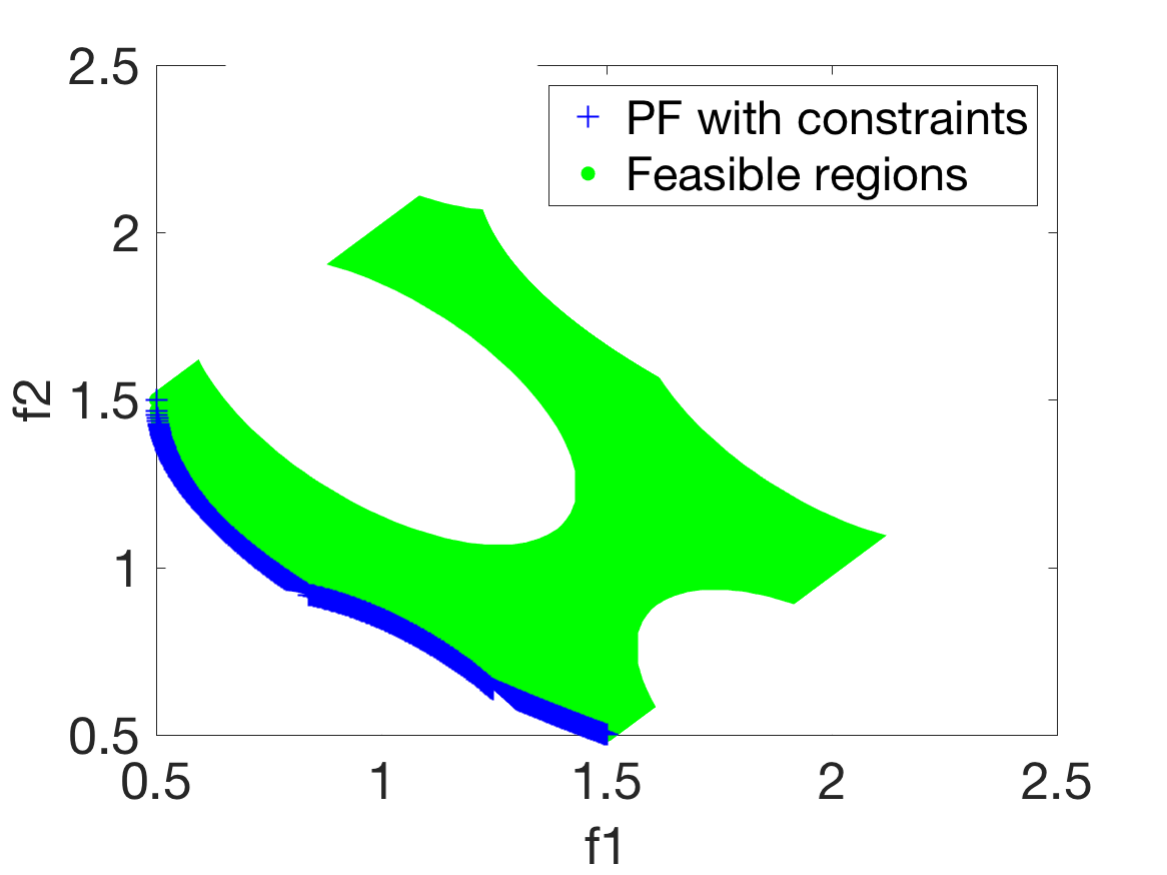}\\
\centering{\scriptsize{(b) DAS-CMOP2 (0,0.5,0.5)}}
\end{minipage}
\hspace{0.5cm}
\begin{minipage}[t]{0.28\linewidth}
\includegraphics[width = 5cm]{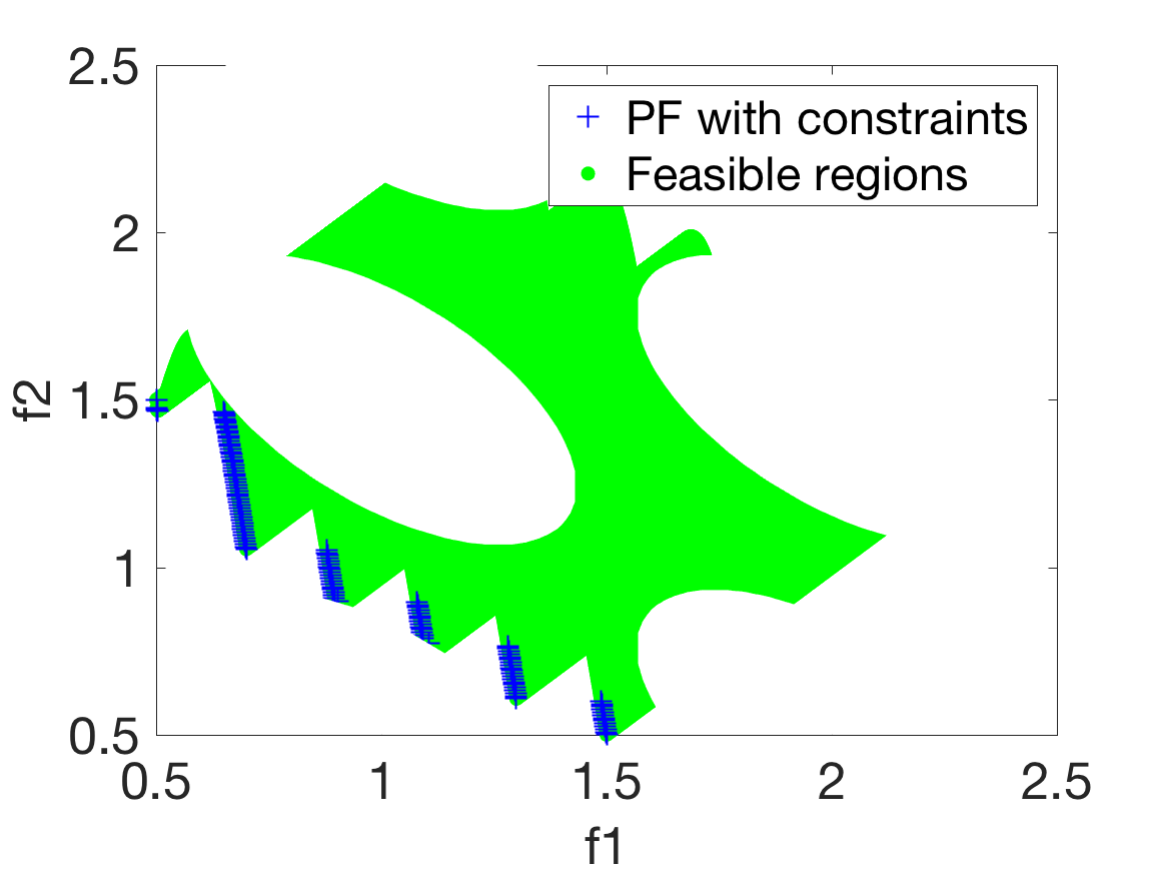}\\
\centering{\scriptsize{(c) DAS-CMOP3 (0,0.5,0.5)}}
\end{minipage}
\end{tabular}

\vspace{0.5cm}
\begin{tabular}{cc}
\begin{minipage}[t]{0.28\linewidth}
\includegraphics[width = 5cm]{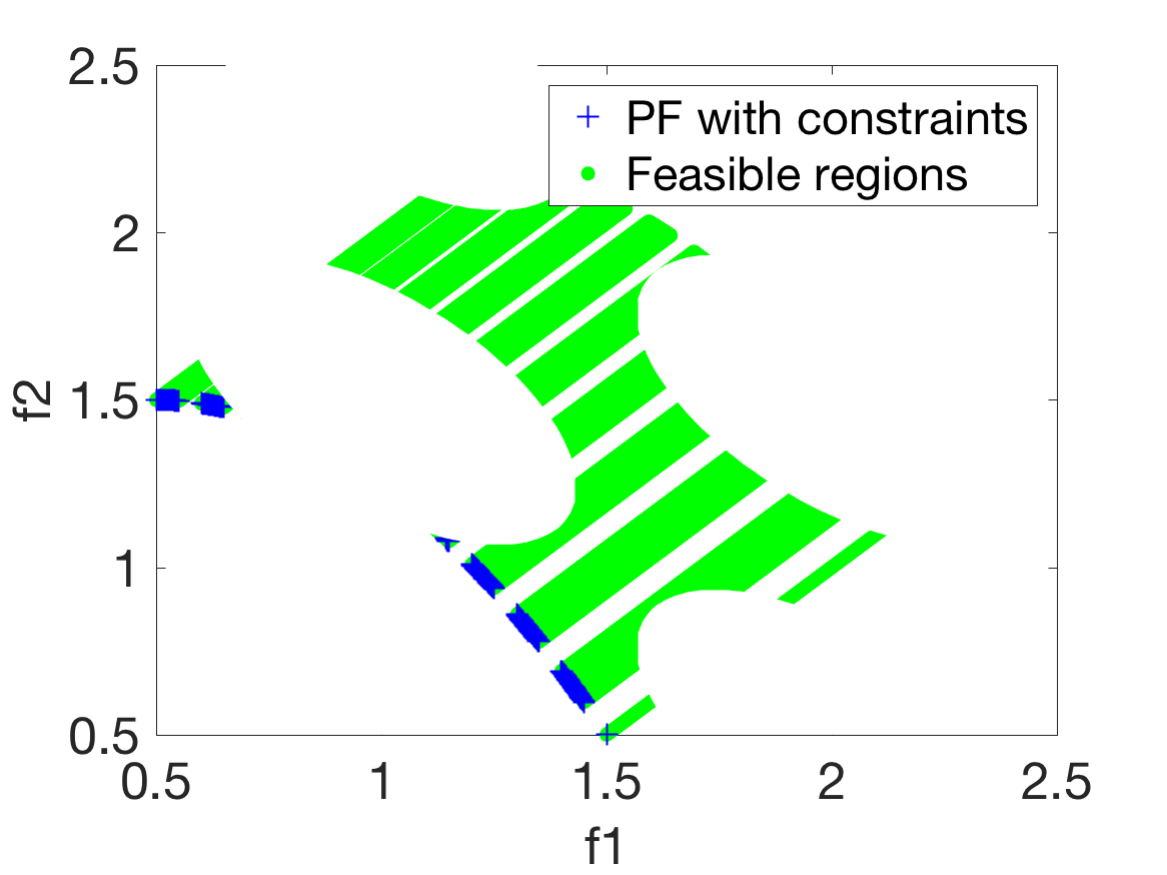}\\
\centering{\scriptsize{(d) DAS-CMOP4 (0.5,0.5,0.5)}}
\end{minipage}
\hspace{0.5cm}
\begin{minipage}[t]{0.28\linewidth}
\includegraphics[width = 5cm]{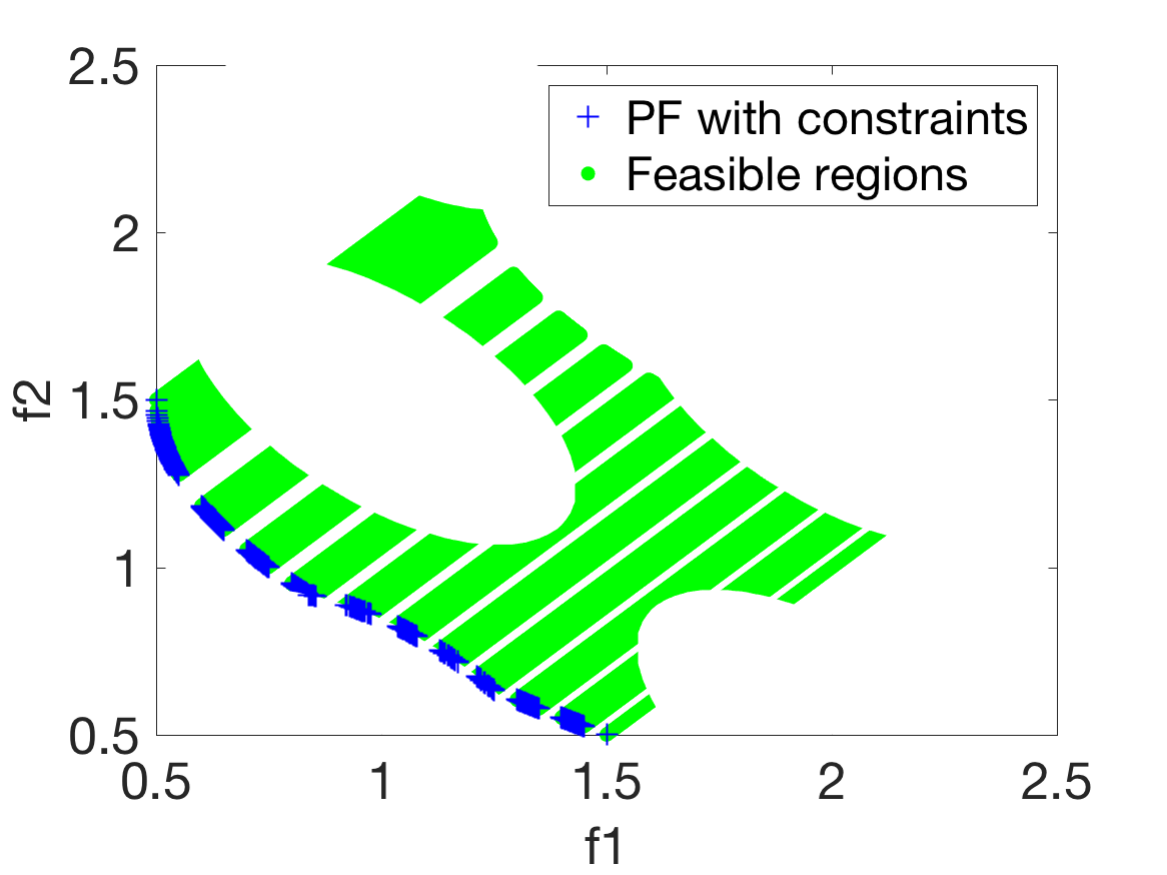}\\
\centering{\scriptsize{(e) DAS-CMOP5 (0.5,0.5,0.5)}}
\end{minipage}
\hspace{0.5cm}
\begin{minipage}[t]{0.28\linewidth}
\includegraphics[width = 5cm]{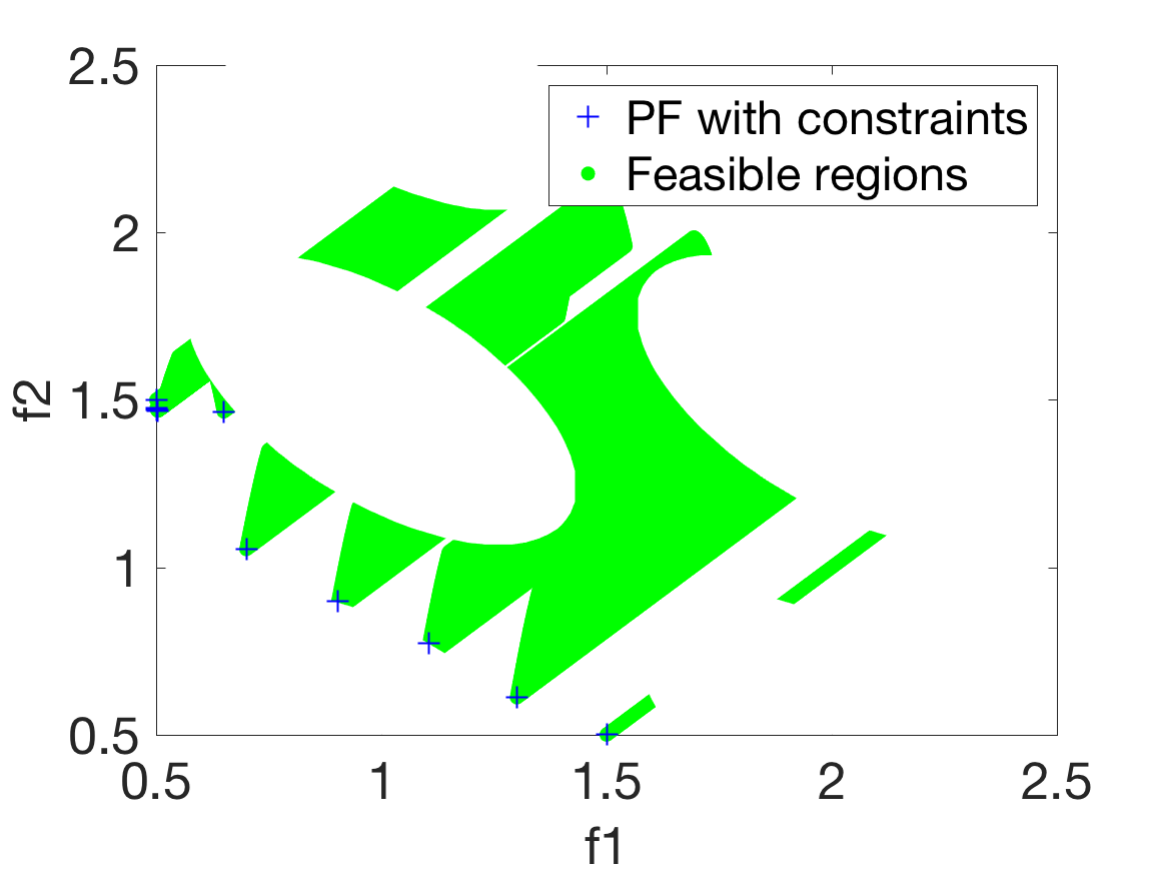}\\
\centering{\scriptsize{(f) DAS-CMOP6 (0.5,0.5,0.5)}}
\end{minipage}
\end{tabular}

\vspace{0.5cm}
\begin{tabular}{cc}
\begin{minipage}[t]{0.28\linewidth}
\includegraphics[width = 5cm]{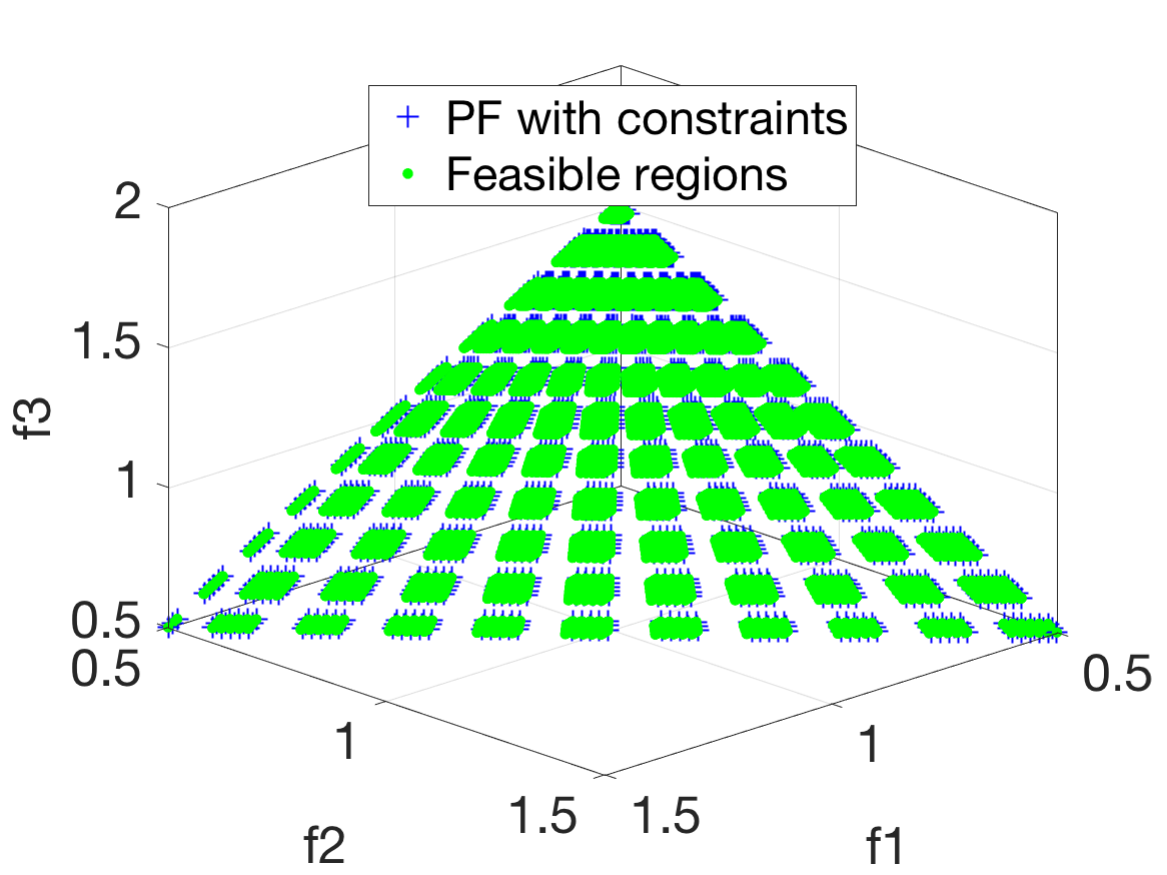}\\
\centering{\scriptsize{(g) DAS-CMOP7 (0.5,0.5,0.5)}}
\end{minipage}
\hspace{0.5cm}
\begin{minipage}[t]{0.28\linewidth}
\includegraphics[width = 5cm]{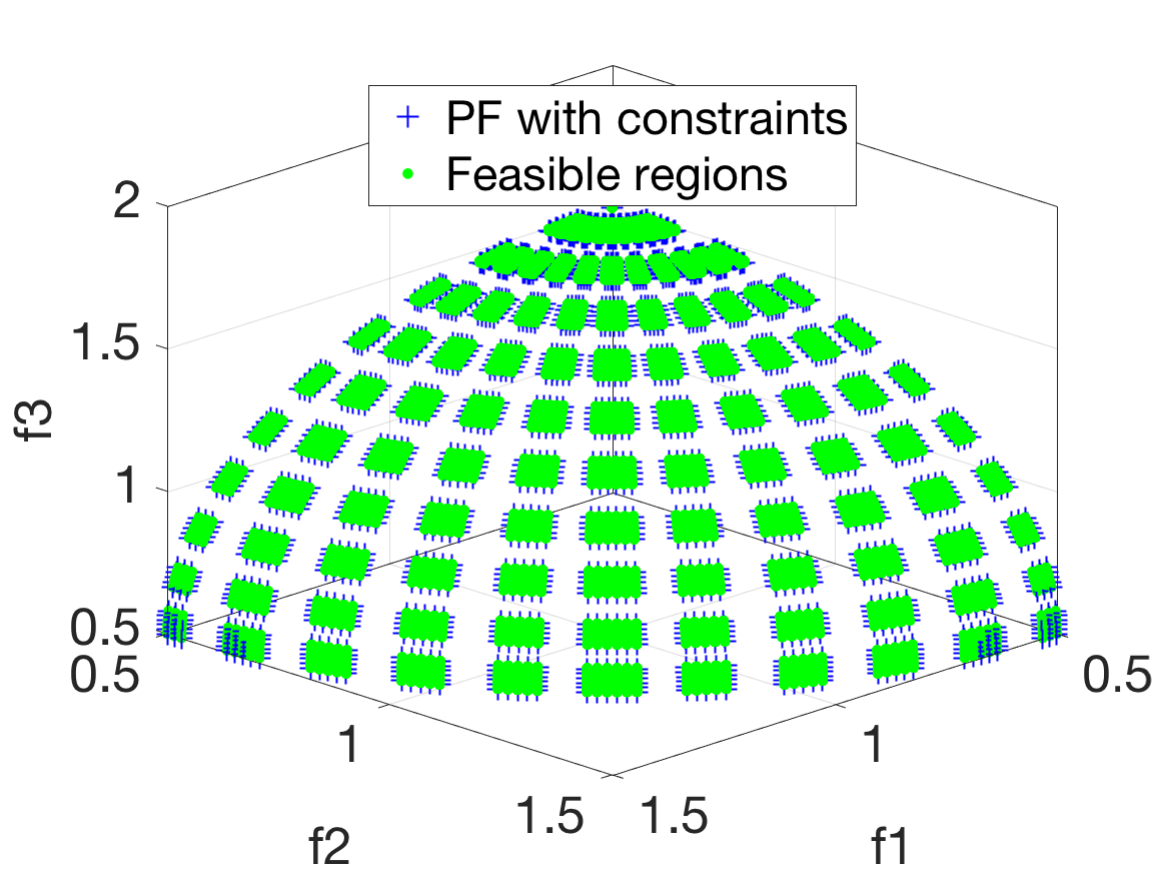}\\
\centering{\scriptsize{(h) DAS-CMOP8 (0.5,0.5,0.5)}}
\end{minipage}
\hspace{0.5cm}
\begin{minipage}[t]{0.28\linewidth}
\includegraphics[width = 5cm]{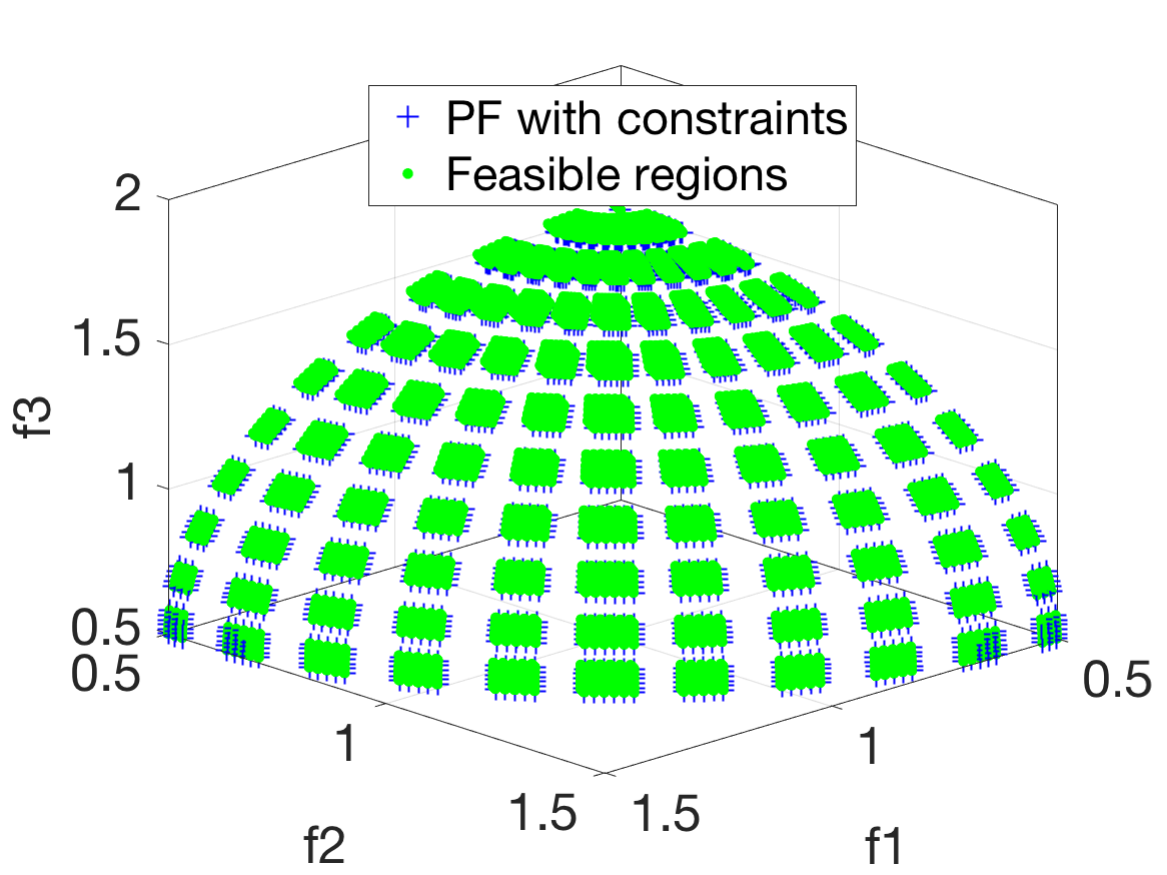}\\
\centering{\scriptsize{(i) DAS-CMOP9 (0.5,0.5,0.5)}}
\end{minipage}
\end{tabular}
\caption{\label{fig:dascmop-pfs} The feasible regions and $PFs$ with constraints (true $PFs$) of DAS-CMOP1-9 with different difficulty triplets are plotted. Since DAS-CMOP1-3 and DAS-CMOP4-6 have similar feasible regions and $PFs$ when they have the same difficulty triplets, the difficulty triplets for DAS-CMOP1-3 are set to $(0,0.5,0.5)$, and the difficulty triplets for DAS-CMOP4-6 are set to $(0.5,0.5,0.5)$. For DAS-CMOP7-9, the difficulty triplets are also set to $(0.5,0.5,0.5)$.}
\end{figure*}

In Table \ref{tab:das-cmops}, DAS-CMOP1-3 have the same constraint functions. DAS-CMOP4-6 also use that same set of constraint functions. The difference between DAS-CMOP1-3 and DAS-CMOP4-6 is that they have different distance functions. The number of objectives in Eq. \eqref{eqn:das-cmop} can be scaled to more than two. For example, DAS-CMOP7-9 have three objectives. The constraint functions of DAS-CMOP8 and DAS-CMOP9 are the same as those of DAS-CMOP7. The feasible regions and true $PFs$ ($PFs$ with constraints) of DAS-CMOP1-9 with different difficulty triplets are plotted in Fig. \ref{fig:dascmop-pfs}.

To construct CMOPs with more than three objectives, we borrow the idea from the WFG toolkit \citep{huband2006review} to construct objectives, which can be scaled to any number of objectives. A general formulation of DAS-CMaOPs is listed as follows.
\begin{eqnarray}
\label{eqn:das-CMaOPs}
\begin{cases}
\text{Given} &\mathbf{z} = \{z_1,\ldots,z_k,z_{k+1},\ldots,z_n\}\\
 \text{Minimize} & f_{m=1:M}(\mathbf{x}) = Dx_{M} + S_mh_m(x_{1:M-1})\\
\text{where} & \mathbf{x} = \{ x_1,\ldots,x_M \} \\
\text{subject to} & c_k(\mathbf{x}) = \sin(a\pi x_k) - b \ge 0, \text{if } k \text{ is odd} \\
&c_k(\mathbf{x}) = \cos(a\pi x_k) - b \ge 0, \text{if } k \text{ is even} \\
&c_M(\mathbf{x}) = (e - x_M)(x_M - d) \ge 0\\
&c_{M+P}(\mathbf{x}) = \sum _{j = 1,j\neq P}^M {(f_j(x)/S_j)^2} \\
&+ (f_P(x)/S_P - 1)^2 - r^2 \ge 0 \\
&c_{2M+1}(\mathbf{x}) = \sum _{j = 1}^M {(f_j(\mathbf{x})/S_j - \frac{1}{\sqrt{M}})^2}\\
& - r^2 \ge 0\\
&a = 20,d = 0.5, n = 30\\
& k = 1,\ldots,M-1, P = 1,\ldots,M\\

& b = \eta, \text{if $\eta = 0$, $c_{1:M-1}(\mathbf{x}) = 0$ }\\
& e - d = 10^{-2\zeta}, \text{if $\zeta = 0$, $c_{M}(\mathbf{x}) = 0$ }\\
& r = 0.5 * \gamma, \text{if $\gamma = 0$, $c_{M+1:2M+1}(\mathbf{x}) = 0$ }\\
\end{cases}
\end{eqnarray}
where $\mathbf{z}$ is a decision vector, and $\mathbf{x}$ is an intermediate vector. There is a mapping between $\mathbf{x}$ and $\mathbf{z}$, and the details can be found in the literature \citep{huband2006review}. In this paper, nine DAS-CMaOPs named DAS-CMaOP1-9 are suggested. The objective functions of DAS-CMaOP1-9 are the same as those of WFG1-9 \citep{huband2006review}. The number of position-related parameters in the decision vector $\mathbf{z}$ is set to 10, and the number of distance-related parameters in the decision vector $\mathbf{z}$ is set to 20.

\begin {table*} [tbp]
\centering
\caption{DAS-CMOPs Test suite: the objective functions and constraint functions of DAS-CMOP1-9.}
\label{tab:das-cmops}

\scalebox{0.65}[0.65]{
\begin{tabular}{|l|l|l|}
\hline
Problem & Objectives & Constraints \\
\hline
DAS-CMOP1
& $\begin{cases}
\text{min} & f_1(\mathbf{x}) = x_1 + g(\mathbf{x}) \\
\text{min} & f_2(\mathbf{x}) = 1 - x_1^2 + g(\mathbf{x}) \\
\text{where} &g(\mathbf{x}) = \sum_{j = 1} ^ {n}  {(x_j  - \sin(0.5 \pi x_1))} ^{2} \\
& n = 30, \mathbf{x}\in [0,1]^{n}
\end{cases}$
& $\begin{cases}
&c_1(\mathbf{x}) = \sin(a\pi x_1) - b \ge 0\\
&c_{2}(\mathbf{x}) = (e - g(\mathbf{x}))(g(\mathbf{x}) - d) \ge 0\\
&c_{k+2}(\mathbf{x}) = ((f_1 - p_k) \cos\theta_k - ( f_2 - q_k) \sin\theta_k)^2 / a_k^2 \\
& + ((f_1 - p_k) \sin\theta_k + (f_2 - q_k) \cos\theta_k)^2 / b_k ^{2} \ge r \\
& p_k=[0,1,0,1,2,0,1,2,3], a_k^{2} = 0.3, b_k^{2} =1.2, \theta_k = -0.25\pi\\
& q_k=[1.5,0.5,2.5,1.5,0.5,3.5,2.5,1.5,0.5]\\
& a = 20, d = 0.5,\eta = (b+1)/2,\zeta = \exp(d - e),\gamma = 2r\\
\end{cases}$\\
 \hline
DAS-CMOP2
& $\begin{cases}
\text{min} & f_1(\mathbf{x}) = x_1 + g(\mathbf{x}) \\
\text{min} & f_2(\mathbf{x}) = 1 - \sqrt{x_1} + g(\mathbf{x}) \\
\text{where} &g(\mathbf{x}) = \sum_{j = 1} ^ {n}  {(x_j  - \sin(0.5 \pi x_1))} ^{2} \\
& n = 30, \mathbf{x}\in [0,1]^{n}
\end{cases}$
& They are the same as those of DAS-CMOP1\\
 \hline
DAS-CMOP3
& $\begin{cases}
\text{min} & f_1(\mathbf{x}) = x_1 + g(\mathbf{x}) \\
\text{min} & f_2(\mathbf{x}) = 1 - \sqrt{x_1} + 0.5 * |\sin(5\pi x_1)| + g(\mathbf{x}) \\
\text{where} &g(\mathbf{x}) = \sum_{j = 1} ^ {n}  {(x_j  - \sin(0.5 \pi x_1))} ^{2} \\
& n = 30, \mathbf{x}\in [0,1]^{n}
\end{cases}$
& They are the same as those of DAS-CMOP1\\
 \hline
DAS-CMOP4
& $\begin{cases}
\text{min} & f_1(\mathbf{x}) = x_1 + g(\mathbf{x}) \\
\text{min} & f_2(\mathbf{x}) = 1 - x_1^2 + g(\mathbf{x}) \\
\text{where} &g(\mathbf{x}) = (n - 1) + \sum_{j = 2}^{n} {(x_j - 0.5)^2 - \cos(20\pi(x_j - 0.5))}\\
& n = 30, \mathbf{x} \in [0,1]^{n}
\end{cases}$
 & $\begin{cases}
&c_1(\mathbf{x}) = \sin(a\pi x_1) - b \ge 0\\
&c_{2}(\mathbf{x}) = (e - g(\mathbf{x}))(g(\mathbf{x}) - d) \ge 0\\
&c_{k+2}(\mathbf{x}) = ((f_1 - p_k) \cos\theta_k - ( f_2 - q_k) \sin\theta_k)^2 / a_k^2 \\
& + ((f_1 - p_k) \sin\theta_k + (f_2 - q_k) \cos\theta_k)^2 / b_k ^{2} \ge r \\
& p_k=[0,1,0,1,2,0,1,2,3], a_k^{2} = 0.3, b_k^{2} =1.2, \theta_k = -0.25\pi\\
& q_k=[1.5,0.5,2.5,1.5,0.5,3.5,2.5,1.5,0.5]\\
& a = 20, d = 0.5,\eta = (b+1)/2,\zeta = \exp(d - e),\gamma = 2r\\
\end{cases}$\\
 \hline

DAS-CMOP5
&$\begin{cases}
\text{min} & f_1(\mathbf{x}) = x_1 + g(\mathbf{x}) \\
\text{min} & f_2(\mathbf{x}) = 1 - \sqrt{x_1} + g(\mathbf{x}) \\
\text{where} &g(\mathbf{x}) = (n - 1) + \sum_{j = 2}^{n} {(x_j - 0.5)^2 - \cos(20\pi(x_j - 0.5))}\\
& n = 30, \mathbf{x} \in [0,1]^{n}
\end{cases}$
& They are the same as those of DAS-CMOP4\\
 \hline

DAS-CMOP6
&$\begin{cases}
\text{min} & f_1(\mathbf{x}) = x_1 + g(\mathbf{x}) \\
\text{min} & f_2(\mathbf{x}) = 1 - \sqrt{x_1} + 0.5 * |\sin(5\pi x_1)| + g(\mathbf{x}) \\
\text{where} &g(\mathbf{x}) = (n - 1) + \sum_{j = 2}^{n} {(x_j - 0.5)^2 - \cos(20\pi(x_j - 0.5))}\\
& n = 30, \mathbf{x} \in [0,1]^{n}
\end{cases}$
& They are the same as those of DAS-CMOP4\\
 \hline

DAS-CMOP7
& $\begin{cases}
\text{min} & f_1(\mathbf{x}) = x_1 * x_2 + g(\mathbf{x}) \\
\text{min} & f_2(\mathbf{x}) = x_2 * (1 - x_1) + g(\mathbf{x}) \\
\text{min} & f_3(\mathbf{x}) = 1 - x_2 + g(\mathbf{x}) \\
\text{where} &g(\mathbf{x}) = (n - 2) + \sum_{j = 3}^{n} {(x_j - 0.5)^2 - \cos(20\pi(x_j - 0.5))}\\
& n = 30, \mathbf{x} \in [0,1]^{n}
\end{cases}$
 & $\begin{cases}
&c_1(\mathbf{x}) = \sin(a\pi x_1) - b \ge 0\\
&c_2(\mathbf{x}) = \cos(a\pi x_2) - b \ge 0\\
&c_{3}(\mathbf{x}) = (e - g(\mathbf{x}))(g(\mathbf{x}) - d) \ge 0\\
&c_{k+3}(\mathbf{x}) = \sum _{j = 1,j\neq k}^3 {f_j^2} + (f_k - 1)^2 - r^2 \ge 0 \\
&c_{7}(\mathbf{x}) = \sum _{j = 1}^3 {(f_j - \frac{1}{\sqrt{3}})^2} - r^2 \ge 0\\
& a = 20, d = 0.5, k = 1,2,3\\
&\eta = (b+1)/2,\zeta = \exp(d - e),\gamma = 2r\\
\end{cases}$\\
\hline
DAS-CMOP8
& $\begin{cases}
\text{min} & f_1(\mathbf{x}) = \cos(0.5 \pi x_1) * \cos(0.5 \pi x_2) + g(\mathbf{x}) \\
\text{min} & f_2(\mathbf{x}) = \cos(0.5 \pi x_1) * \sin(0.5 \pi x_2) + g(\mathbf{x}) \\
\text{min} & f_3(\mathbf{x}) = \sin(0.5 \pi x_1) + g(\mathbf{x}) \\
\text{where} &g(\mathbf{x}) = (n - 2) + \sum_{j = 3}^{n} {(x_j - 0.5)^2 - \cos(20\pi(x_j - 0.5))}\\
& n = 30, \mathbf{x} \in [0,1]^{n}
\end{cases}$
& They are the same as those of DAS-CMOP7\\
\hline
DAS-CMOP9
& $\begin{cases}
\text{min} & f_1(\mathbf{x}) = \cos(0.5 \pi x_1) * \cos(0.5 \pi x_2) + g(\mathbf{x}) \\
\text{min} & f_2(\mathbf{x}) = \cos(0.5 \pi x_1) * \sin(0.5 \pi x_2) + g(\mathbf{x}) \\
\text{min} & f_3(\mathbf{x}) = \sin(0.5 \pi x_1) + g(\mathbf{x}) \\
\text{where} &g(\mathbf{x}) = \sum_{j = 3} ^ {n} {(x_j  - \cos(\frac{0.25j}{n} \pi (x_1 + x_2) ))} ^{2} \\
& n = 30, \mathbf{x} \in [0,1]^{n}
\end{cases}$
& They are the same as those of DAS-CMOP7\\
\hline
\end{tabular}

}
\end{table*}

\section{Experimental Study} \label{section:experimental_study}

\subsection{Experimental Settings}
To test the performance of CMOEAs on the DAS-CMOPs, two commonly used CMOEAs (i.e., MOEA/D-CDP and NSGA-II-CDP) are tested on DAS-CMOP1-9 with sixteen different difficulty triplets. The difficulty triplets for each DAS-CMOP are listed in Table \ref{tab:dif_triplets}. To test the performance of CMaOEAs on the DAS-CMaOPs, two popular CMaOEAs (i.e., C-MOEA/DD and C-NSGA-III) are tested on DAS-CMaOP1-9 with the first twelve different difficulty triplets in Table \ref{tab:dif_triplets}.

\begin{table}[htbp]
  \centering
  \caption{Sixteen difficulty triplets are set for each DAS-CMOP. The difficulty levels of the feasibility-hardness of the last four difficulty triplets are set to 1.0. In this circumstance, some inequality constraints of DAS-CMOPs are transformed into equality constraints, which can be used to test the performance of MOEA/D-CDP and NSGA-II-CDP in solving DAS-CMOPs with equality constraints. For each DAS-CMaOP, the first twelve difficulty triplets are used. Numbers 1, 5 and 9 represent problems that are diversity-hard. Numbers 2, 6 and 10 represent problems that are feasibility-hard. Numbers 3, 7 and 11 represent problems that are convergence-hard. Numbers 4, 8 and 12 represent problems that simultaneously exhibit all three types of difficulty. Numbers 13-15 represent problems with equality constraints.}
  \scalebox{0.9}[0.9]{
    \begin{tabular}{|l|c|l|c|l|c|l|c|}
    \hline
    No.   & \multicolumn{1}{l|}{Difficulty Triplet} & No.   & \multicolumn{1}{l|}{Difficulty Triplet} & No.   & \multicolumn{1}{l|}{Difficulty Triplet} & No.   & \multicolumn{1}{l|}{Difficulty Triplet} \\
    \hline
    1     & (0.25,0.0,0.0) & 2     & (0.0,0.25,0.0) & 3     & (0.0,0.0,0.25) & 4     & (0.25,0.25,0.25) \\

    5     & (0.5,0.0,0.0) & 6     & (0.0,0.5,0.0) & 7     & (0.0,0.0,0.5) & 8     & (0.5,0.5,0.5) \\
    9     & (0.75,0.0,0.0) & 10    & (0.0,0.75,0.0) & 11    & (0.0,0.0,0.75) & 12    & (0.75,0.75,0.75) \\
    13    & (0.0,1.0,0.0) & 14    & (0.5,1.0,0.0) & 15    & (0.0,1.0,0.5) & 16    & (0.5,1.0,0.5) \\
    \hline
    \end{tabular}%
    }
  \label{tab:dif_triplets}%
\end{table}%

\subsection{Test Algorithms}
To evaluate the proposed test problems with 2 and 3 objectives, two popular algorithms, NSGA-II \citep{deb2002fast} and MOEA/D \citep{zhang2007moea}, are tested. To deal with constraints, a widely used constraint-handling mechanism called CDP \citep{deb2001constrained} is integrated with NSGA-II and MOEA/D. The constrained versions of NSGA-II and MOEA/D are called NSGA-II-CDP and MOEA/D-CDP, respectively. Two popular algorithms for solving problems with more than 3 objectives, including NSGA-III \citep{deb2014evolutionary,jain2014evolutionary} and MOEA/DD \citep{RN507}, are tested on the proposed test problems with more than 3 objectives. Their constrained versions are called C-NSGA-III and C-MOEA/DD respectively. A brief introduction of each algorithm is given below.
\begin{enumerate}
\item NSGA-II \citep{deb2002fast}: NSGA-II is a widely used dominance-based MOEA. In NSGA-II, a fast non-dominated sorting operation is applied on the population. Each individual in the population is assigned to a non-dominated rank. Solutions are first selected into the next generation based their non-dominated rank until the number of solutions is equal to or greater than the population size. Then, crowding distance is applied to select the rest of the solutions to be included.

\item MOEA/D \citep{zhang2007moea}: MOEA/D is a representative of decomposition-based MOEAs. In MOEA/D, a multi-objective optimization problem is decomposed into a number of scalar optimization subproblems, and these subproblems are optimized simultaneously in a cooperative way.

\item NSGA-III \citep{deb2014evolutionary,jain2014evolutionary}: NSGA-III is an improved version of NSGA-II. It overcomes the drawbacks of NSGA-II for solving problems with more than 3 objectives. In NSGA-III, a set of uniformly distributed weight vectors is used to generate niches, associate and select elite individuals.

\item MOEA/DD \citep{RN507}: MOEA/DD is a unified paradigm that combines dominance- and decomposition-based approaches for many-objective optimization. It exploits the merits of both dominance- and decomposition-based approaches to balance the convergence and diversity of populations during the evolutionary process.

\end{enumerate}

The source codes of the proposed test problems - DAS-CMOP1-9 and DAS-CMaOP1-9, and the tested algorithms can be found in the following URL: \url{http://imagelab.stu.edu.cn/Content.aspx?type=content&Content_ID=1310}.

The detailed parameters of the algorithms are summarized as follows.
\begin{enumerate}
\item  Setting for reproduction operators: The mutation probability $Pm = 1/n$ ($n$ is the number of decision variables). For the polynomial mutation operator, the distribution index is set to 20. For the simulated binary crossover (SBX) operator, the distribution index is set to 20 for DAS-CMOP1-9, and 30 for DAS-CMaOP1-9. The rate of crossover $CR = 0.9$.

\item Population size: For DAS-CMOP1-9, $N = 300$. For 5-objective DAS-CMaOP1-9, $N = 210$, for 8-objective DAS-CMaOP1-9, $N = 156$, and for 10-objective DAS-CMaOP1-9, $N = 275$ \citep{deb2014evolutionary}.

\item Number of runs and stopping condition: For DAS-CMOP1-9, each algorithm runs 30 times independently on each test problem, for each of sixteen different difficulty triplets. The maximum number of function evaluations is 300,000 for DAS-CMOP1-9. For DAS-CMaOP1-9, each algorithm runs 20 times independently on each test problem, for each of twelve different difficulty triplets. For 5-, 8-, and 10-objective DAS-CMaOP1-9, the maximum numbers of function evaluations are 400,000, 500,000, and 600,000, respectively.

\item Neighborhood size: $T = \lfloor 0.1 N \rfloor$.

\item Probability of selecting individuals from its neighborhood: $\delta = 0.9$.

\item The maximal number of solutions replaced by a child: $nr = 2$.
\end{enumerate}

\subsection{Performance Metrics}
To measure the performance of MOEA/D-CDP and NSGA-II-CDP on DAS-CMOP1-9 with different difficulty triplets, four different performance metrics are adopted, which are defined as follows.

\begin{itemize}
\item \textbf{Inverted Generational Distance} ($IGD$) \citep{coello2005solving}:
\end{itemize}
The $IGD$ metric simultaneously reflects the performance of convergence and diversity, and is defined as follows:

\begin{equation} \label{IGD_metric}
\begin{cases}
IGD(P^*,A) = \frac{\sum \limits_{y^* \in P^*}d(y^*,A)}{| P^* |}\\
\\
d(y^*,A) = \min \limits_{y \in A} \{ \sqrt {\sum_{i = 1} ^m (y^{*}_{i} - y_i}) ^ 2 \}
\end{cases}
\end{equation}
where $P^*$ is the ideal $PF$ set, $A$ is an approximate $PF$ set achieved by an algorithm. $m$ represents the number of objectives. It is worth noting that smaller values of $IGD$ represent better performance with respect to both diversity and convergence. To construct $P^*$, uniform sampling is performed on the unconstrained $PF$ and constraint boundaries. For 2- and 3-objective problems, 1000 and 10000 points are sampled. Then, non-dominated sorting is used to select feasible and non-dominated solutions to construct $P^*$. For problems with more than 3 objectives, a huge number of uniformly distributed points along the $PF$ would be required to calculate the $IGD$ in a reliable manner \citep{ishibuchi2015behavior}. In this case, only the hypervolume metric is employed to evaluate the performance of the compared algorithms \citep{yuan2016balancing}.

\begin{itemize}
\item \textbf{Hypervolume} ($HV$) \citep{797969}:
\end{itemize}
$HV$ reflects the closeness of the set of non-dominated solutions achieved by a CMOEA to the true $PF$. A larger $HV$ means that the corresponding non-dominated set is closer to the true $PF$.

\begin{equation}
HV(S)=VOL(\bigcup \limits_{x\in S} [f_1(x),z_1^r]\times ...[f_m(x),z_m^r] ) \\
\end{equation}
where $VOL(\cdot)$ denotes the Lebesgue measure, $m$ is the number of objectives, and $\mathbf{z}^r=(z_1^r,...,z_m^r)^T$ is a user-defined reference point in the objective space. For each DAS-CMaOP, the reference point $\mathbf{z}^r$ is set to $(3.0,\ldots, 2.0 \times m + 1.0)^T$. For DAS-CMaOPs with 5 and 8 objectives, we adopt the WFG algorithm of \cite{while2012fast} to calculate the exact $HV$. The presented $HV$ values are all normalized to $[0,1]$ by dividing them by $z = \prod_{i=1}^{m} z_i^r$. For DAS-CMaOPs with 10 objectives, a Monte Carlo sampling is applied to approximate the $HV$, and the sampling size is set to 100000.

\begin{itemize}
\item \textbf{Coverage} ($C$-metric):
\end{itemize}
Let $A$ and $B$ be two approximation sets obtained by two different algorithms. $C(A,B)$ defines the fraction of solutions in $B$ that are dominated by at least one solution in $A$ \citep{1197687}.
\begin{equation}
C(A,B)= \frac{|\{u \in B \mid \exists \nu \in A: \nu \prec u \}|}{|B|}  \\
\end{equation}
$C(A,B) = 1$ means that each solution in $B$ is dominated by at least one solution in $A$, while $C(A,B) = 0$ indicates that no solutions in $B$ are dominated by solutions in $A$.

\begin{itemize}
\item \textbf{Spacing} ($Sp$):
\end{itemize}
$Sp$ is designed to measure how evenly the solutions of an approximation set are distributed. The definition of $Sp$ is as follows:
\begin{equation} \label{spacing_metric}
\begin{cases}
Sp(A) = \sqrt{\frac{1}{|A| - 1} \sum_{x \in A} (\bar{d} - d_x)^2}\\
\\
d_x = \min\limits_{x^{*} \neq x}^{x^{*} \in A}\{ \sum_{i=1}^{m} |f_i(x) - f_i(x ^ {*})| \} \\
\\
\bar{d} = \frac{1}{|A|} \sum_{x \in A} d_x \\
\end{cases}
\end{equation}
$Sp = 0$ means that all solutions of the approximation set are equidistantly spaced.

It is worth noting that only feasible solutions are employed to calculate $IGD$, $HV$, $C$-metric, and $Sp$ values. When dealing with equality constraints, we usually convert the equality constraints into inequality constraints by introducing an extremely small positive number $\epsilon$ as follows \citep{ullah2012handling}.
\begin{equation}
\label{equ:equ2inequ}
h_j(x)^{'} \equiv \epsilon - |h_j(x)| \ge 0 \\
\end{equation}
where $h_j(x)$ is an equality constraint, and $h_j(x)^{'}$ is an inequality constraint transformed from $h_j(x)$. Here $\epsilon$ is set to $1e-6$. If a solution satisfies Eq. \eqref{equ:equ2inequ}, it is considered to be a feasible solution. Otherwise, it is infeasible.

\subsection{Performance Comparisons on DAS-CMOPs}

Table \ref{tab:dascmop1-5} presents the statistics on $IGD$, $C$-metric and Spacing values for MOEA/D-CDP and NSGA-II-CDP on DAS-CMOP1-5 with sixteen different difficulty triplets as shown in Table \ref{tab:dif_triplets}. We can see that for DAS-CMOP1 and DAS-CMOP2 with simultaneous diversity-, feasibility- and convergence-hardness---i.e., the difficulty triplets are $(0.25,0.25,0.25)$, $(0.5,0.5,0.5)$ and $(0.75,0.75,0.75)$---NSGA-II-CDP is significantly better than MOEA/D-CDP in $IGD$ values. For DAS-CMOP1 and DAS-CMOP2 with a single difficulty level, MOEA/D-CDP is significantly better than NSGA-II-CDP in most cases.  For DAS-CMOP1 and DAS-CMOP2 with equality constraints---i.e., the difficulty triplets are $(0.0,1.0,0.0)$, $(0.5,1.0,0.0)$, $(0.0,1.0,0.5)$ and $(0.5,1.0,0.5)$---NSGA-II-CDP is better or significantly better than MOEA/D-CDP in $IGD$ values.

For DAS-CMOP3 with equality constraints, NSGA-II-CDP is better or significantly better than MOEA/D-CDP in $IGD$ values, and for DAS-CMOP3 with the remainder of the difficulty triplets, MOEA/D is significantly better than NSGA-II-CDP except for the case with the difficulty triplet $(0.5,0.5,0.5)$. For DAS-CMOP4 with equality constraints, MOEA/D-CDP is significantly better than NSGA-II-CDP in $IGD$ values. For DAS-CMOP4 with diversity-hardness---i.e., the difficulty triplets are $(0.25,0.0,0.0)$, $(0.5,0.0,0.0)$ and $(0.75,0.0,0.0)$---NSGA-II-CDP is significantly better than MOEA/D-CDP in $IGD$ values.

For DAS-CMOP5 with convergence-hardness---i.e., the difficulty triplets are $(0.0,0.0,0.25)$, $(0.0,0.0,0.5)$ and $(0.0,0.0,0.75)$---MOEA/D-CDP is significantly better than NSGA-II-CDP in $IGD$ values. For DAS-CMOP5 with the rest of the difficulty triplets, NSGA-II-CDP is significantly better than MOEA/D-CDP in most cases in $IGD$ values.

In terms of $C$-metric values, NSGA-II-CDP and MOEA/D-CDP perform almost the same on DAS-CMOP1-4. For DAS-CMOP5, MOEA/D-CDP is significantly better than NSGA-II-CDP in $C$-metric values. In terms of Spacing values, NSGA-II-CDP is better or significantly better than MOEA/D-CDP on DAS-CMOP1-5 in most cases.

Table \ref{tab:dascmop6-9} shows the statistics on $IGD$, $C$-metric and Spacing values for MOEA/D-CDP and NSGA-II-CDP on DAS-CMOP6-9 with sixteen different difficulty triplets as shown in Table \ref{tab:dif_triplets}. We can see that for DAS-CMOP6 with convergence-hardness, MOEA/D-CDP is significantly better than NSGA-II-CDP in $IGD$ values. For DAS-CMOP6 with diversity-hardness, NSGA-II-CDP is better or significantly better than MOEA/D-CDP in $IGD$ values.

For DAS-CMOP7 and DAS-CMOP8 with equality constraints---i.e., the difficulty triplets are $(0.0,1.0,0.0)$, $(0.5,1.0,0.0)$, $(0.0,1.0,0.5)$ and $(0.5,1.0,0.5)$---MOEA/D-CDP is better or significantly better than NSGA-II-CDP in $IGD$ values. For DAS-CMOP7 and DAS-CMOP8 with the rest of difficulty triplets, NSGA-II-CDP is significantly better than MOEA/D-CDP in most cases. For DAS-CMOP9 with equality constraints, NSGA-II-CDP is better or significantly better than MOEA/D-CDP except for the case with the difficult triplet $(0.0,1.0,0.0)$ in $IGD$ values. For DAS-CMOP9 with the other difficulty triplets, MOEA/D-CDP is better or significantly better than MOEA/D-CDP in $IGD$ values.

In terms of $C$-metric values, MOEA/D-CDP is better or significantly better than NSGA-II-CDP on DAS-CMOP6-9 in most cases. In terms of Spacing values, NSGA-II-CDP is better or significantly better than MOEA/D-CDP on DAS-CMOP7-9 in most cases.

\begin{figure*}
\begin{tabular}{cc}
\hspace{-0.5cm}
\begin{minipage}[t]{0.25\linewidth}
\includegraphics[width= 4.2cm]{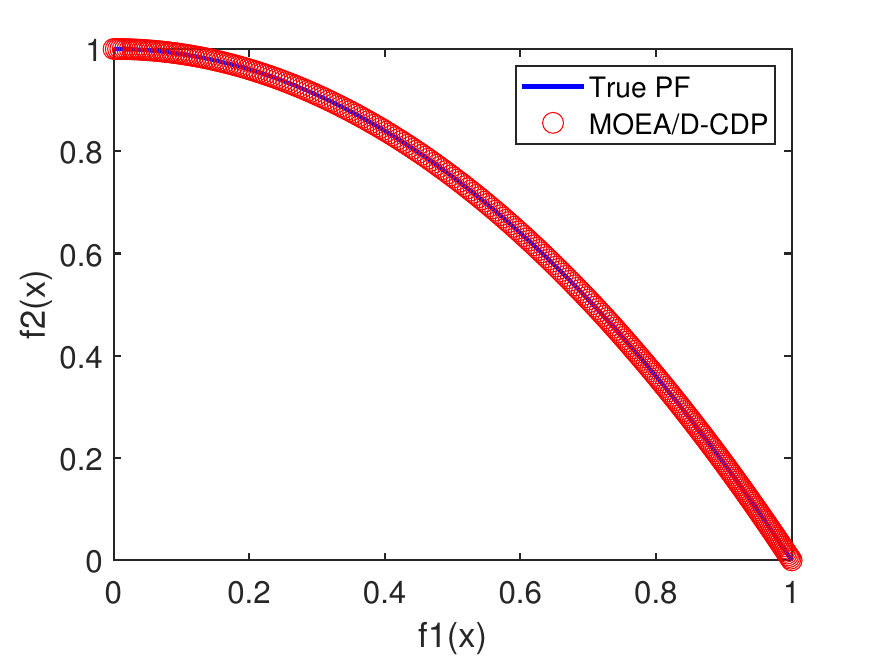}
\centerline{\scriptsize{(a) DAS-CMOP1(3)}}
\end{minipage}
\begin{minipage}[t]{0.25\linewidth}
\includegraphics[width= 4.2cm]{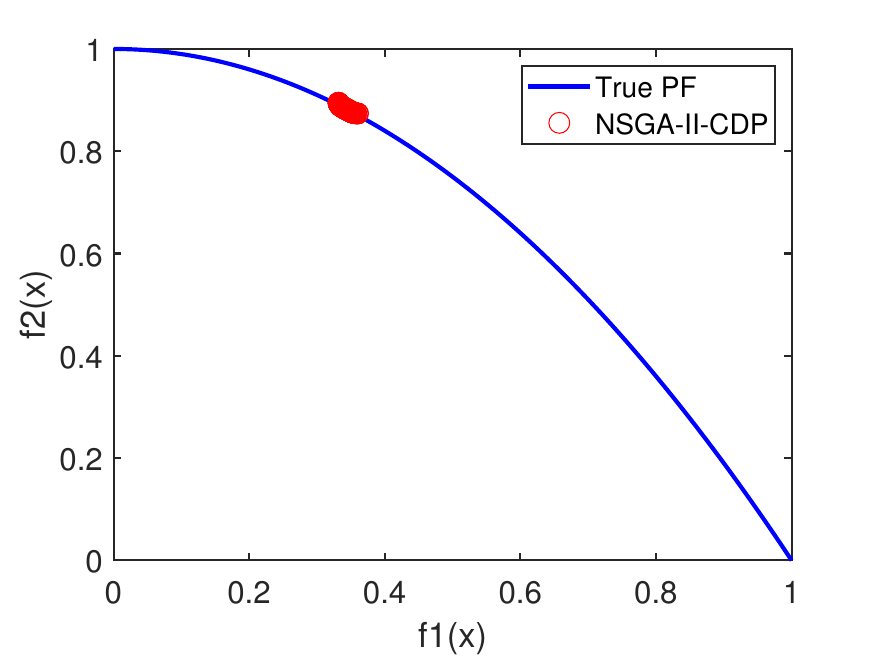}
\centerline{\scriptsize{(b) DAS-CMOP1(3)}}
\end{minipage}
\begin{minipage}[t]{0.25\linewidth}
\includegraphics[width= 4.2cm]{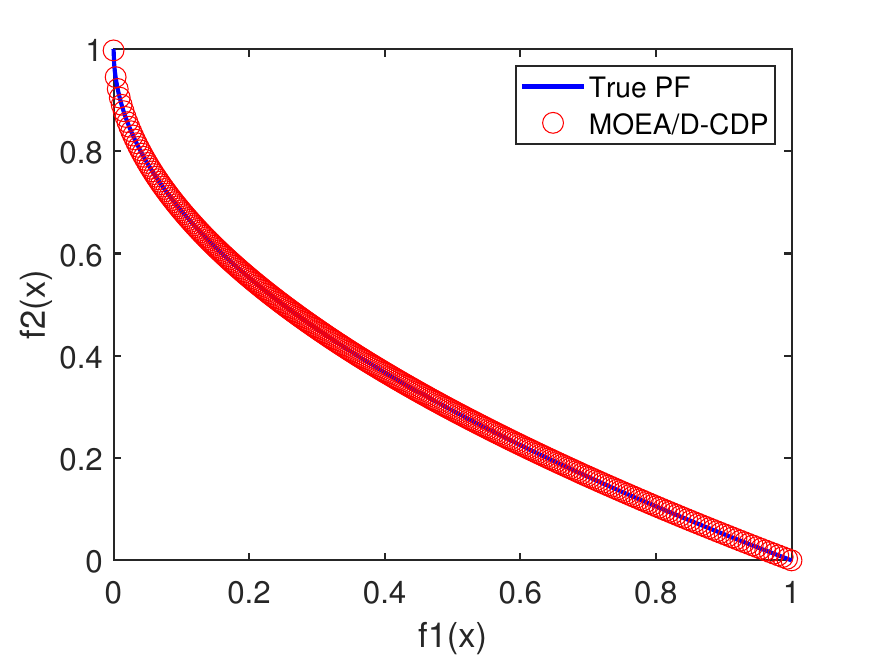}
\centerline{\scriptsize{(c) DAS-CMOP2(3)}}
\end{minipage}
\begin{minipage}[t]{0.25\linewidth}
\includegraphics[width= 4.2cm]{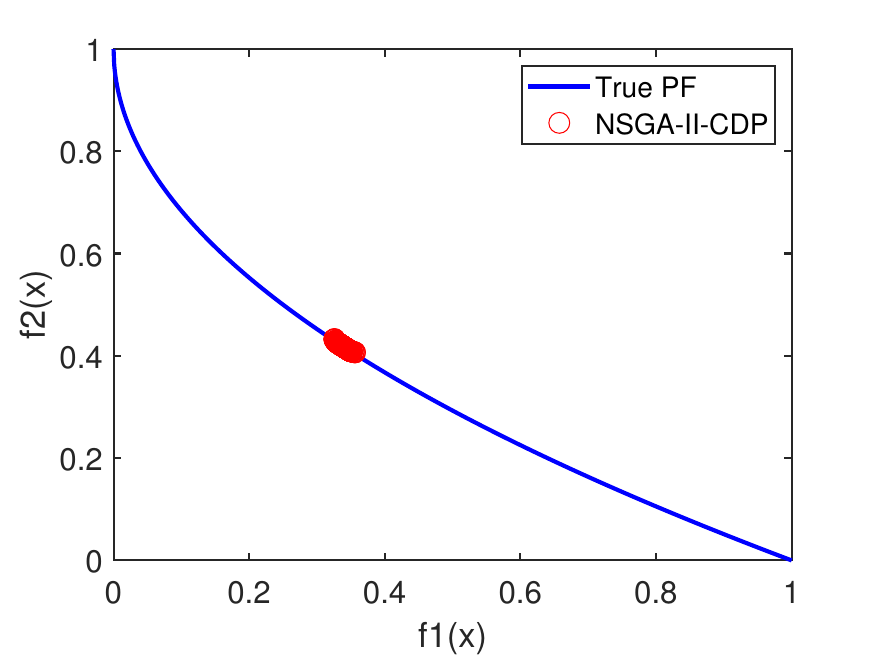}
\centerline{\scriptsize{(d) DAS-CMOP2(3)}}
\end{minipage}
\end{tabular}

\begin{tabular}{cc}
\hspace{-0.5cm}
\begin{minipage}[t]{0.25\linewidth}
\includegraphics[width= 4.2cm]{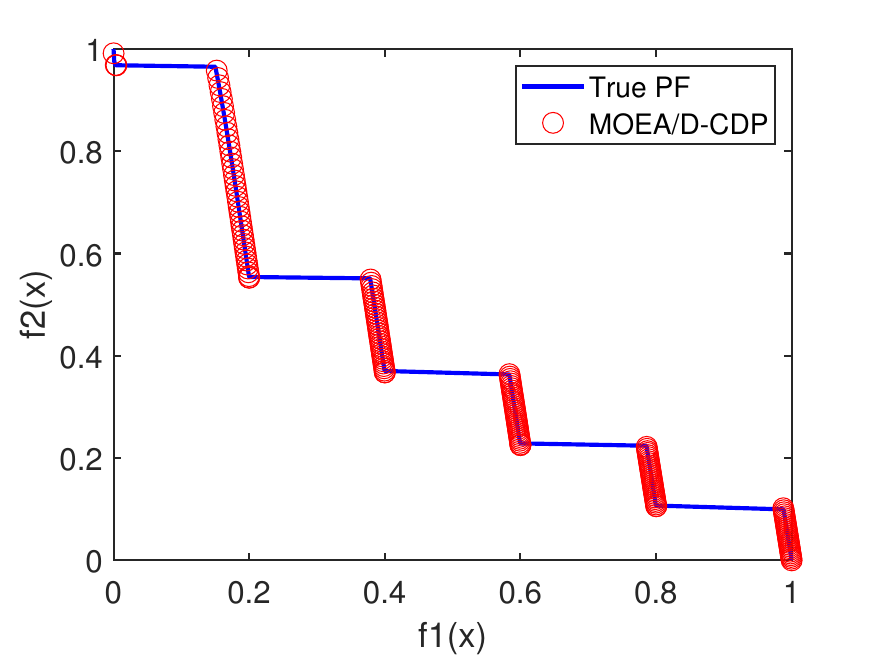}
\centerline{\scriptsize{(e) DAS-CMOP3(3)}}
\end{minipage}
\begin{minipage}[t]{0.25\linewidth}
\includegraphics[width= 4.2cm]{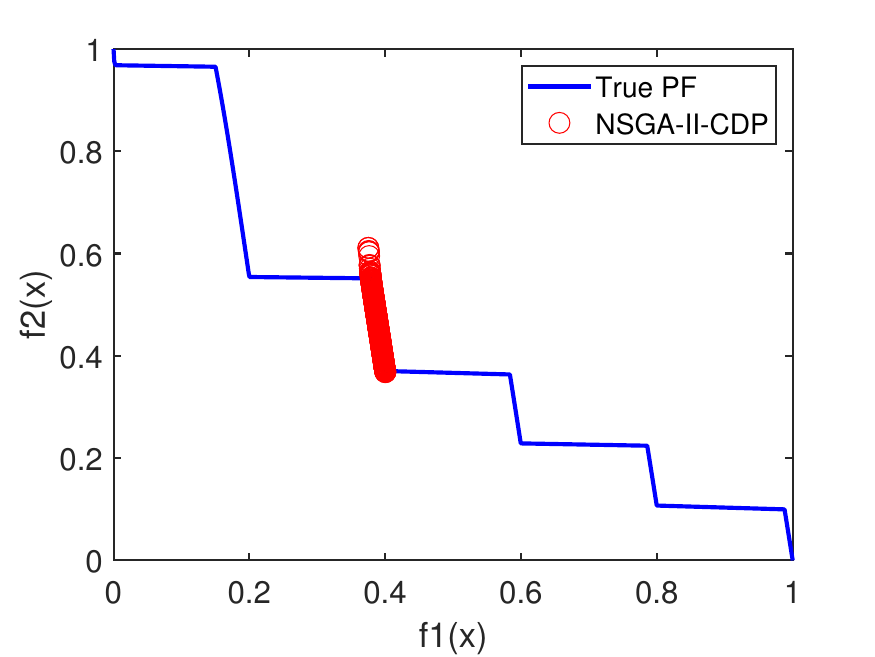}
\centerline{\scriptsize{(f) DAS-CMOP3(3)}}
\end{minipage}
\begin{minipage}[t]{0.25\linewidth}
\includegraphics[width= 4.2cm]{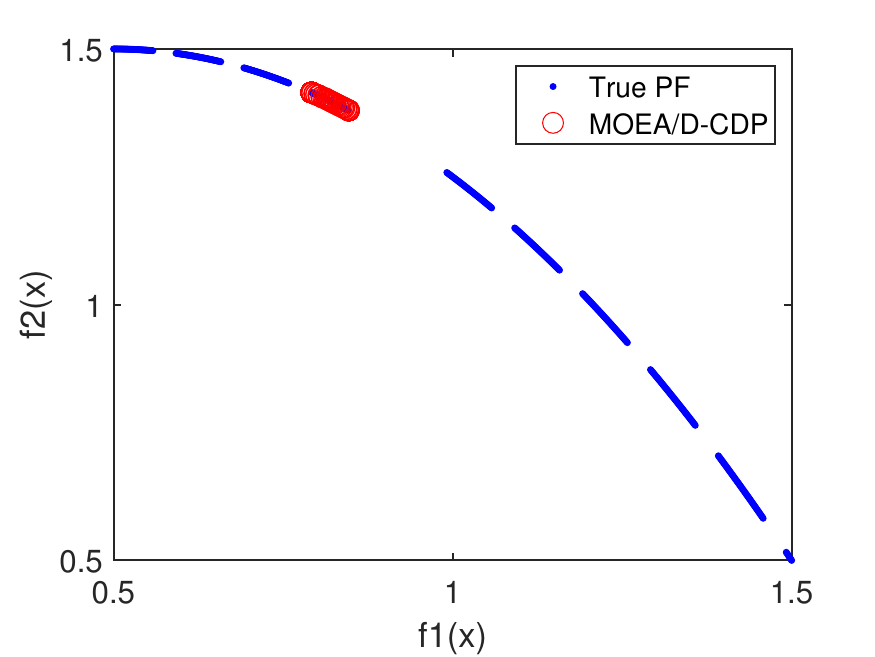}
\centerline{\scriptsize{(g) DAS-CMOP1(4)}}
\end{minipage}
\begin{minipage}[t]{0.25\linewidth}
\includegraphics[width= 4.2cm]{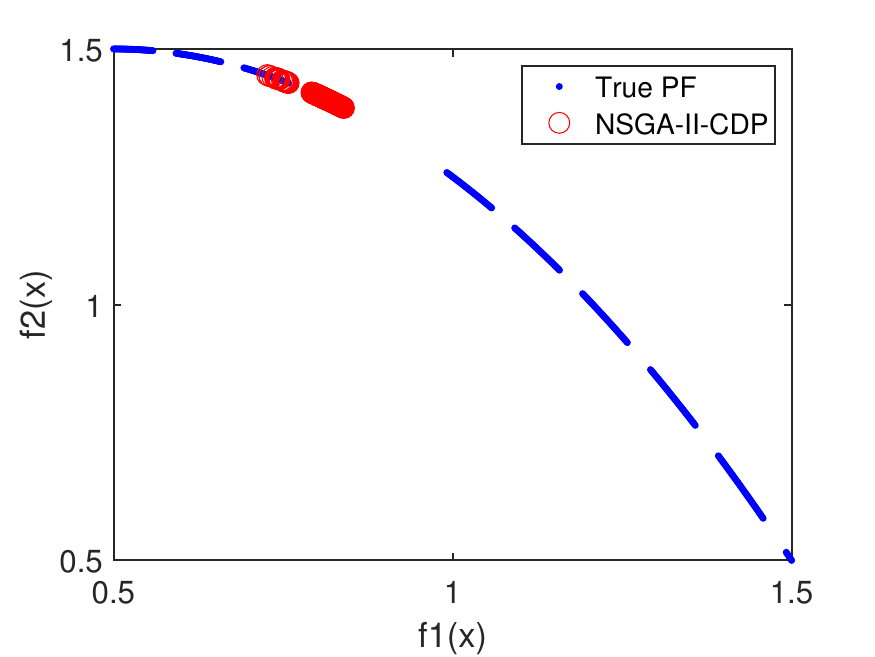}
\centerline{\scriptsize{(h) DAS-CMOP1(4)}}
\end{minipage}
\end{tabular}

\begin{tabular}{cc}
\hspace{-0.5cm}
\begin{minipage}[t]{0.25\linewidth}
\includegraphics[width= 4.2cm]{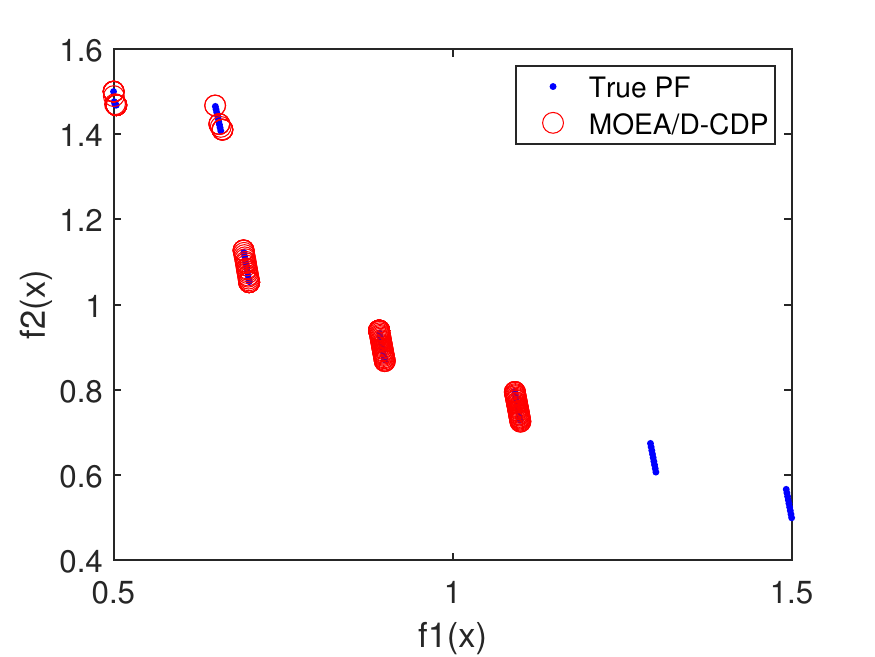}
\centerline{\scriptsize{(i) DAS-CMOP6(4)}}
\end{minipage}
\begin{minipage}[t]{0.25\linewidth}
\includegraphics[width= 4.2cm]{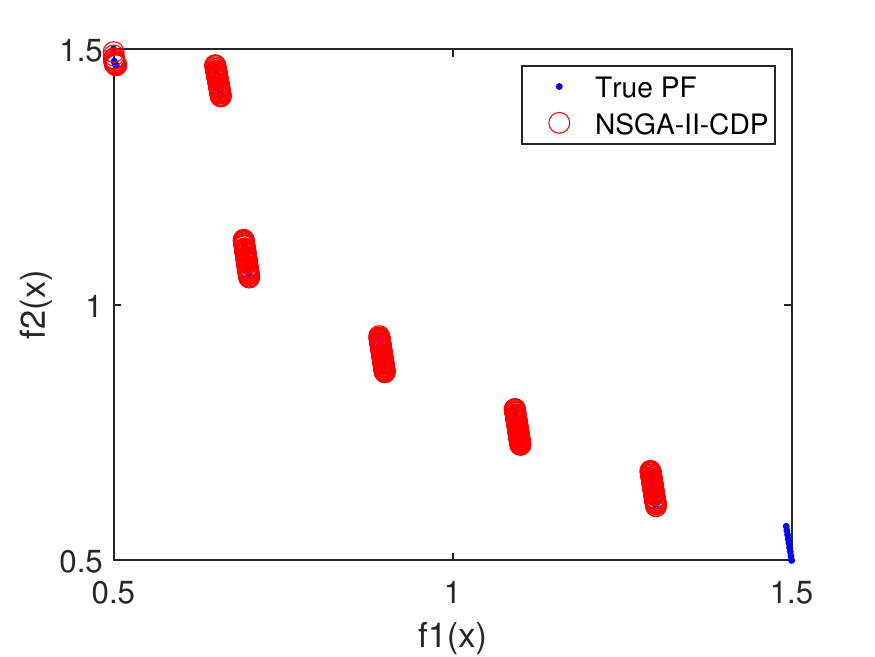}
\centerline{\scriptsize{(j) DAS-CMOP6(4)}}
\end{minipage}
\begin{minipage}[t]{0.25\linewidth}
\includegraphics[width= 4.2cm]{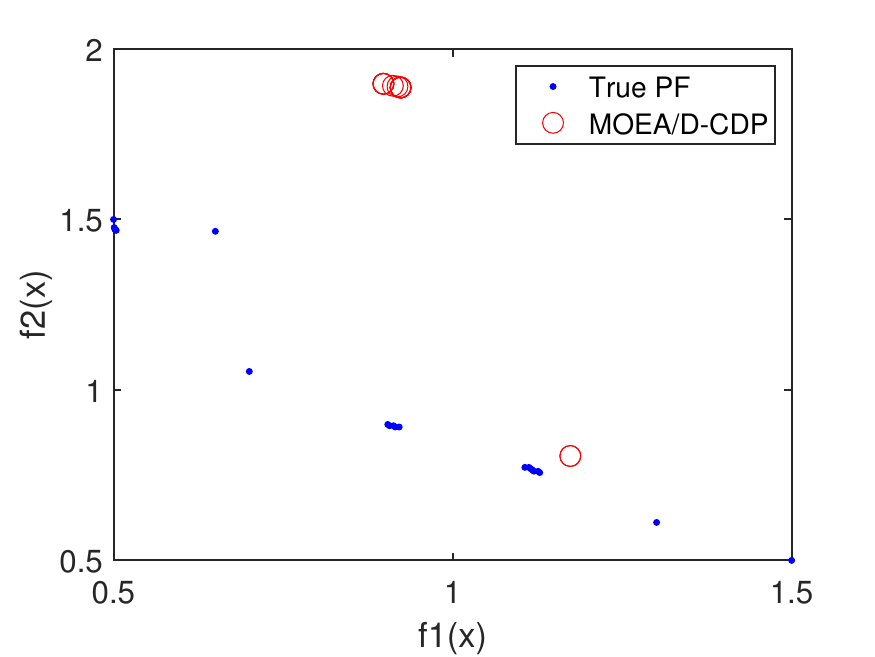}
\centerline{\scriptsize{(k) DAS-CMOP6(8)}}
\end{minipage}
\begin{minipage}[t]{0.25\linewidth}
\includegraphics[width= 4.2cm]{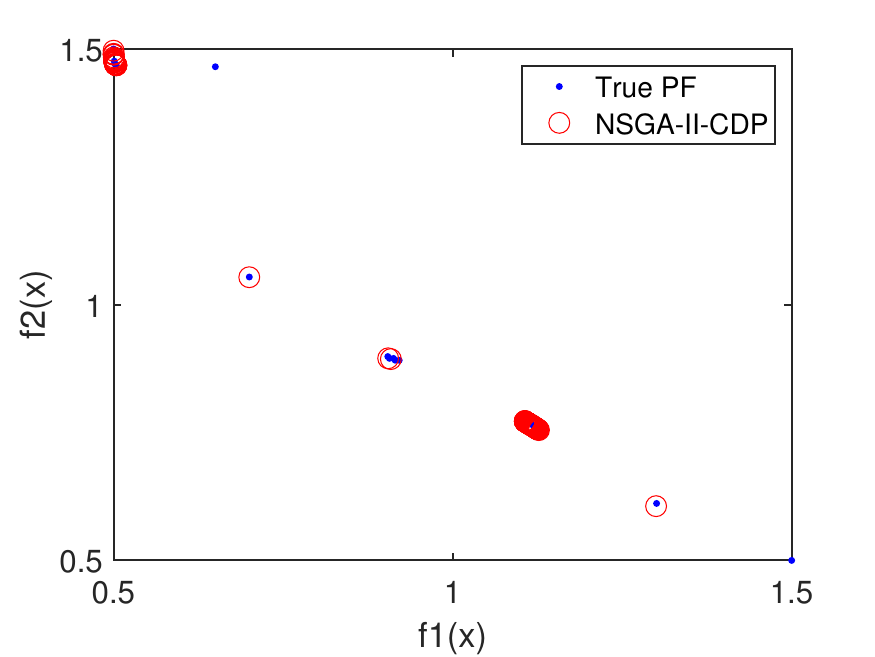}
\centerline{\scriptsize{(l) DAS-CMOP6(8)}}
\end{minipage}
\end{tabular}

\caption{The feasible and non-dominated solutions with the median $IGD$ values in 30 independent runs by using MOEA/D-CDP and NSGA-II-CDP on DAS-CMOP1-3 and DAS-CMOP6 with different difficulty triplets. (a)-(f) show the populations achieved by MOEA/D-CDP and NSGA-II-CDP on DAS-CMOPs with convergence-hardness. (g)-(l) show the populations achieved by MOEA/D-CDP and NSGA-II-CDP on DAS-CMOPs with simultaneous diversity-, feasibility- and convergence-hardness.} \label{Fig:DAS-CMOPs}
\end{figure*}

Figure \ref{Fig:DAS-CMOPs} shows the feasible and non-dominated solutions with the median $IGD$ values in 30 independent runs made using MOEA/D-CDP and NSGA-II-CDP on DAS-CMOP1-3 and DAS-CMOP6 with different difficulty triplets. We can see that MOEA/D-CDP is more suitable for solving convergence-hard DAS-CMOPs, whereas NSGA-II-CDP is more suitable for solving DAS-CMOPs with simultaneous diversity-, feasibility- and convergence-hardness.

\subsection{Performance Comparisons on DAS-CMaOPs}

Table \ref{tab:dascmaop1-6} presents the statistics on $HV$ values for C-MOEA/DD and C-NSGA-III on 5-, 8- and 10-objective DAS-CMaOP1-6 with the first twelve difficulty triplets in Table \ref{tab:dif_triplets}. We can see that for 5- and 8-objective DAS-CMaOP1 with feasibility-hardness, C-NSGA-III is significantly better than C-MOEA/DD. For 5-objective DAS-CMaOP1 with diversity-hardness, C-MOEA/DD is significantly better than C-NSGA-III except for the case with the difficulty triplet $(0.25, 0.0, 0.0)$. For 8-objective DAS-CMaOP1 with diversity-hardness, C-MOEA/DD is significantly better than C-NSGA-III. For 10-objective DAS-CMaOP1, C-NSGA-III is significantly better than C-MOEA/DD in most cases.

For 5-objective DAS-CMaOP2 with each difficulty level less than 0.5, C-NSGA-III is better or significantly better than C-MOEA/DD. For 5-objective DAS-CMaOP2 with the remainder of the difficulty triplets, C-MOEA/DD is better or significantly better than C-NSGA-III except for the cases with difficulty triplets $(0.5, 0.0, 0.0)$ and $(0.0, 0.75, 0.0)$. For 8-objective DAS-CMaOP2, C-NSGA-III is significantly better than C-MOEA/DD in most cases. For 10-objective DAS-CMaOP2 with feasibility-hardness, C-NSGA-III is significantly better than C-MOEA/DD. For 5-objective DAS-CMaOP3, C-NSGA-III is significantly better than C-MOEA/DD in most cases. For 8- and 10-objective DAS-CMaOP3 with diversity-hardness, C-NSGA-III is also significantly better than C-MOEA/DD.

For 5-, 8- and 10-objective DAS-CMaOP4 and DAS-CMaOP5, C-NSGA-III is significantly better than C-MOEA/DD in most cases in $HV$ values, and C-MOEA/DD is significantly better than C-NSGA-III in the case with the difficulty triplet $(0.75, 0.75, 0.75)$. For 5-objective DAS-CMaOP6 with feasibility-hardness, C-NSGA-III is significantly better than C-MOEA/DD, and for 5-objective DAS-CMaOP6 with the remaining difficulty triplets, C-MOEA/DD is better or significantly better than C-NSGA-III. For 8- and 10-objective DAS-CMaOP6 with difficulty triplets $(0.0, 0.0, 0.75)$ and $(0.75, 0.75, 0.75)$, C-MOEA/DD is significantly better than C-NSGA-III. For 8- and 10-objective DAS-CMaOP6 with the remaining difficulty triplets, C-NSGA-III is significantly better than C-MOEA/DD.

Table \ref{tab:dascmaop7-9} presents the statistics on $HV$ values for C-MOEA/DD and C-NSGA-III on 5-, 8- and 10-objective DAS-CMaOP6-9 with the first twelve difficulty triplets in Table \ref{tab:dif_triplets}. We can see that for 5-, 8- and 10-objective DAS-CMaOP7 with feasibility-hardness, C-NSGA-III is significantly better than C-MOEA/DD. For 5-objective DAS-CMaOP7 with simultaneous diversity-, feasibility- and convergence-hardness---i.e., the difficulty triplets are $(0.25,0.25,0.25)$, $(0.5,0.5,0.5)$, $(0.75,0.75,0.75)$---C-MOEA/DD is significantly better than C-NSGA-III. For 8- and 10-objective DAS-CMaOP7 with difficulty triplets $(0.0, 0.0, 0.75)$ and $(0.75, 0.75, 0.75)$, C-MOEA/DD is significantly better than C-NSGA-III.

For 5-objective DAS-CMaOP8, C-MOEA/DD is significantly better than C-NSGA-III in most cases in $HV$ values. For 8- and 10-objective DAS-CMaOP8, C-NSGA-III is significantly better than C-MOEA/DD in most cases. For 8- and 10-objective DAS-CMaOP8 with difficulty triplets $(0.0, 0.0, 0.75)$ and $(0.75, 0.75, 0.75)$, C-MOEA/DD is significantly better than C-NSGA-III.

For 5-objective DAS-CMaOP9 with diversity-hardness, C-NSGA-III is significantly better than C-MOEA/DD, and for 5-objective DAS-CMaOP9 with convergence-hardness, C-MOEA/DD is better or significantly better than C-NSGA-III. For 8- and 10-objective DAS-CMaOP9, C-NSGA-III is significantly better than C-MOEA/DD in most cases. For 8- and 10-objective DAS-CMaOP9 with difficulty triplets $(0.0, 0.0, 0.75)$ and $(0.75, 0.75, 0.75)$, C-MOEA/DD is significantly better than C-NSGA-III.

\begin{figure*}
\begin{tabular}{cc}
\hspace{-0.5cm}
\begin{minipage}[t]{0.25\linewidth}
\includegraphics[width= 4.2cm]{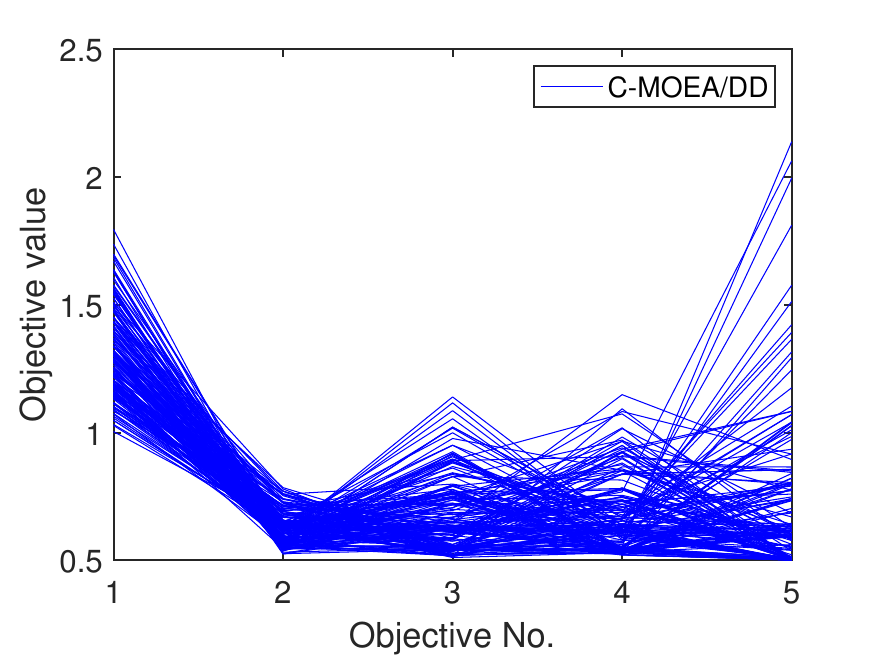}
\centerline{\scriptsize{(a) DAS-CMaOP1(6)}}
\end{minipage}
\begin{minipage}[t]{0.25\linewidth}
\includegraphics[width= 4.2cm]{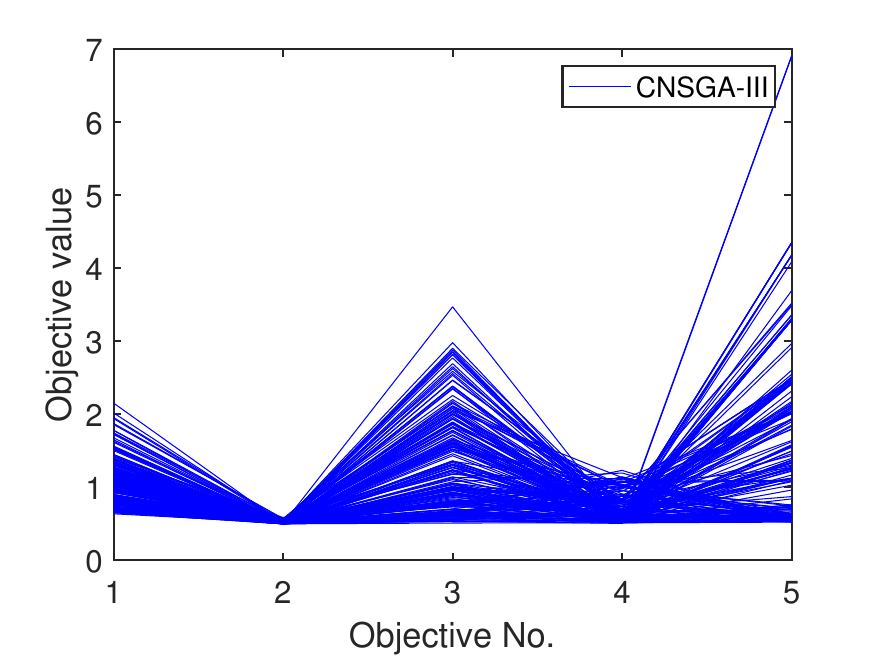}
\centerline{\scriptsize{(b) DAS-CMaOP1(6)}}
\end{minipage}
\begin{minipage}[t]{0.25\linewidth}
\includegraphics[width= 4.2cm]{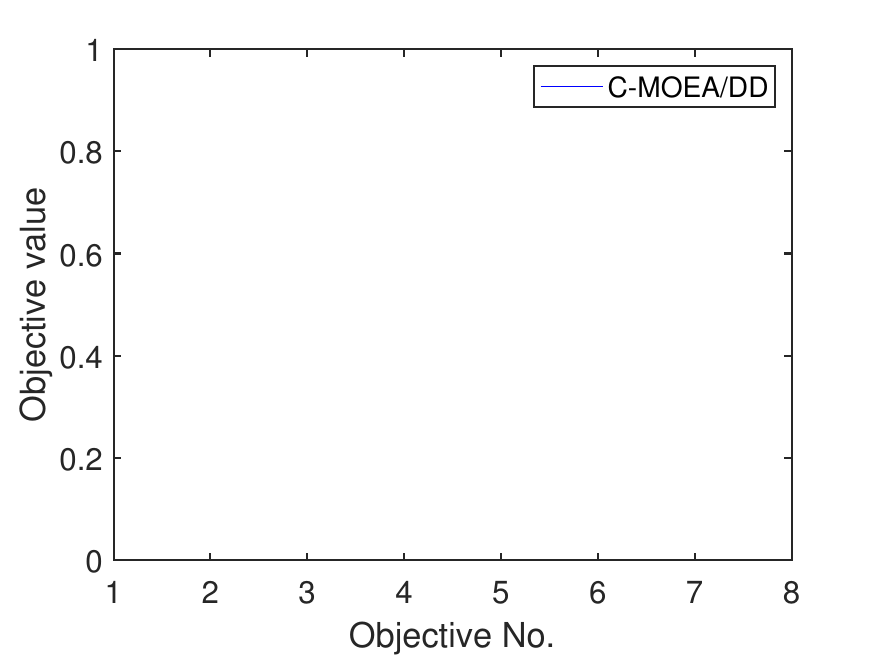}
\centerline{\scriptsize{(c) DAS-CMaOP1(10)}}
\end{minipage}
\begin{minipage}[t]{0.25\linewidth}
\includegraphics[width= 4.2cm]{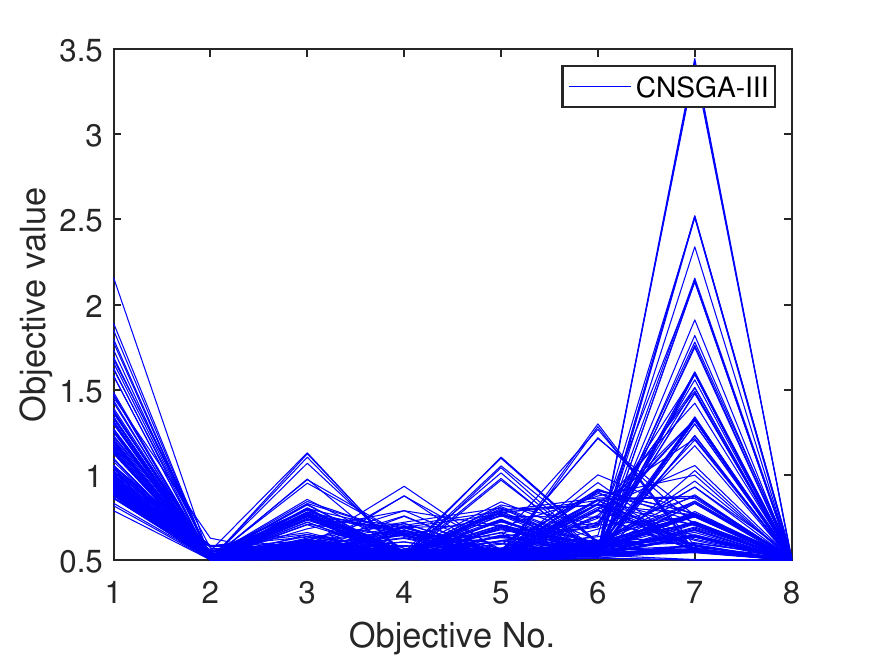}
\centerline{\scriptsize{(d) DAS-CMaOP1(10)}}
\end{minipage}
\end{tabular}

\begin{tabular}{cc}
\hspace{-0.5cm}
\begin{minipage}[t]{0.25\linewidth}
\includegraphics[width= 4.2cm]{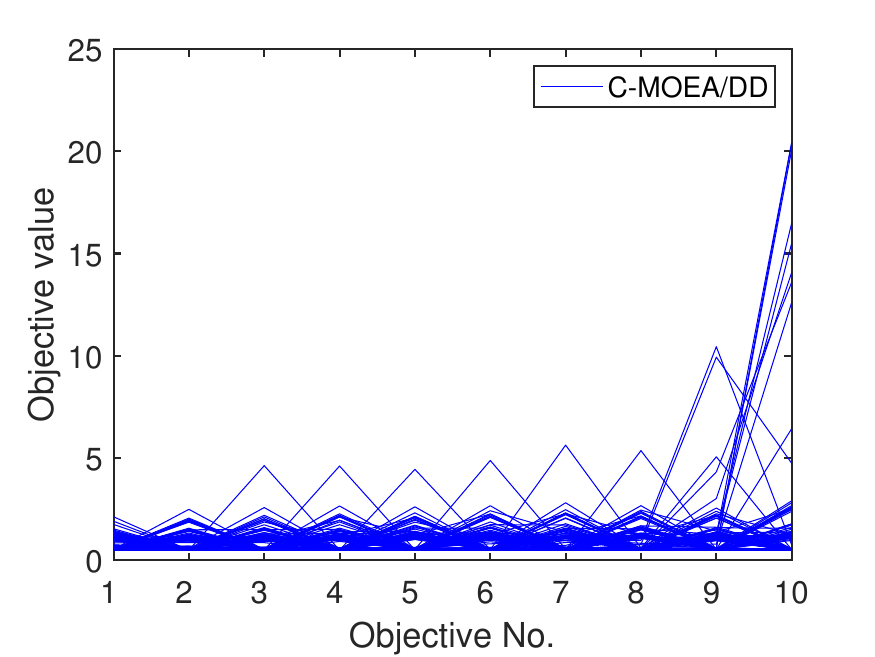}
\centerline{\scriptsize{(e) DAS-CMaOP2(2)}}
\end{minipage}
\begin{minipage}[t]{0.25\linewidth}
\includegraphics[width= 4.2cm]{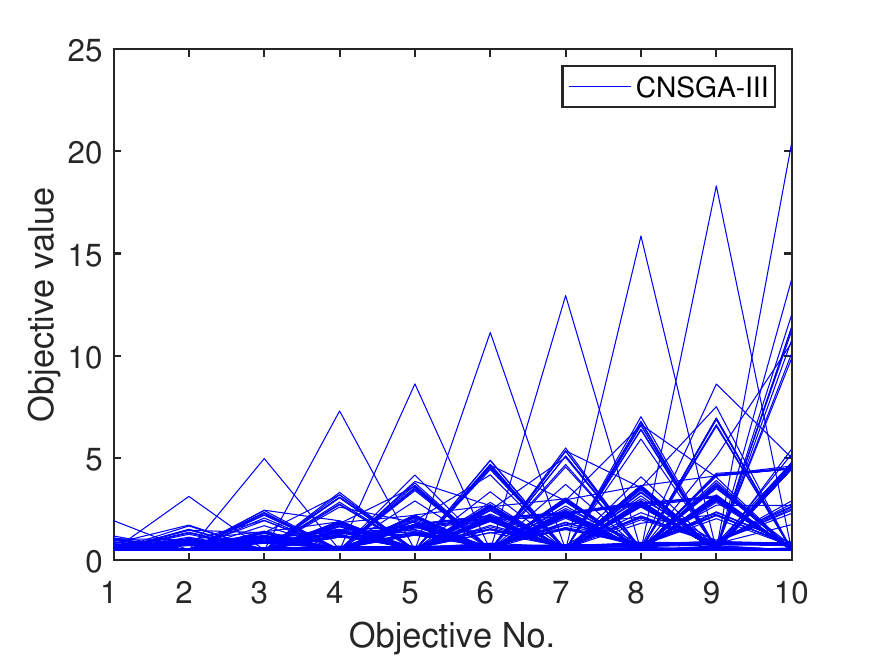}
\centerline{\scriptsize{(f) DAS-CMaOP2(2)}}
\end{minipage}
\begin{minipage}[t]{0.25\linewidth}
\includegraphics[width= 4.2cm]{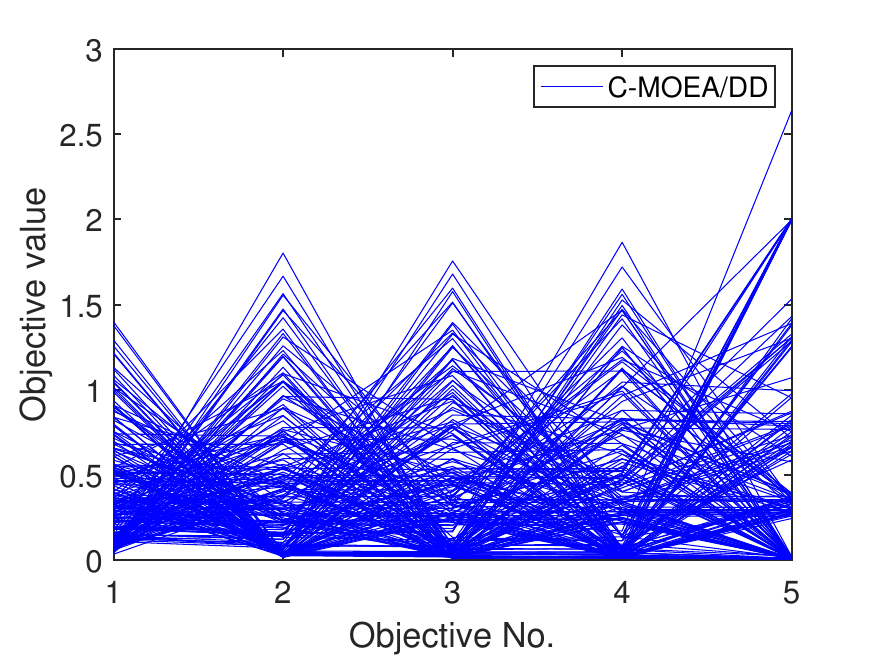}
\centerline{\scriptsize{(g) DAS-CMaOP2(7)}}
\end{minipage}
\begin{minipage}[t]{0.25\linewidth}
\includegraphics[width= 4.2cm]{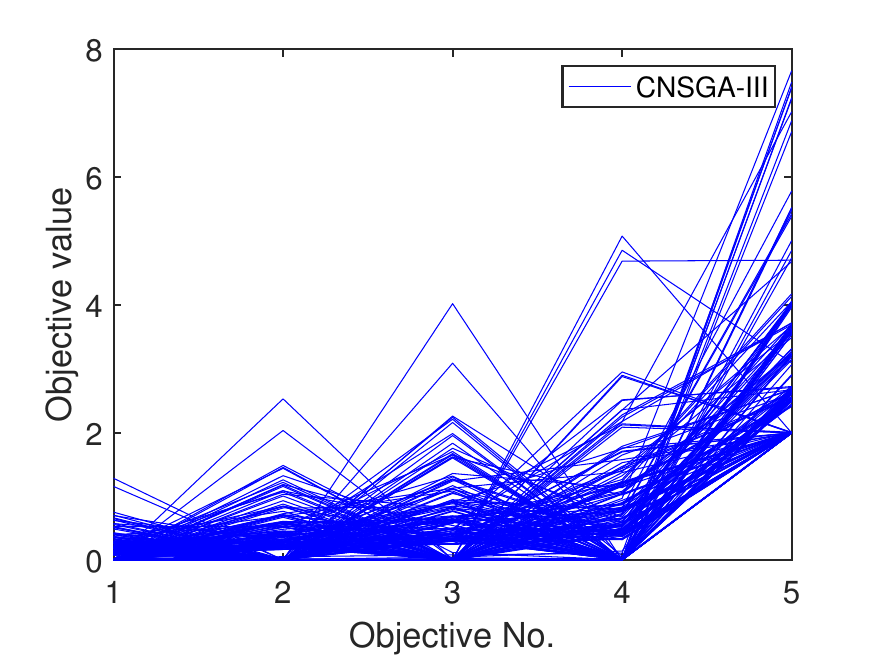}
\centerline{\scriptsize{(h) DAS-CMaOP2(7)}}
\end{minipage}
\end{tabular}

\begin{tabular}{cc}
\hspace{-0.5cm}
\begin{minipage}[t]{0.25\linewidth}
\includegraphics[width= 4.2cm]{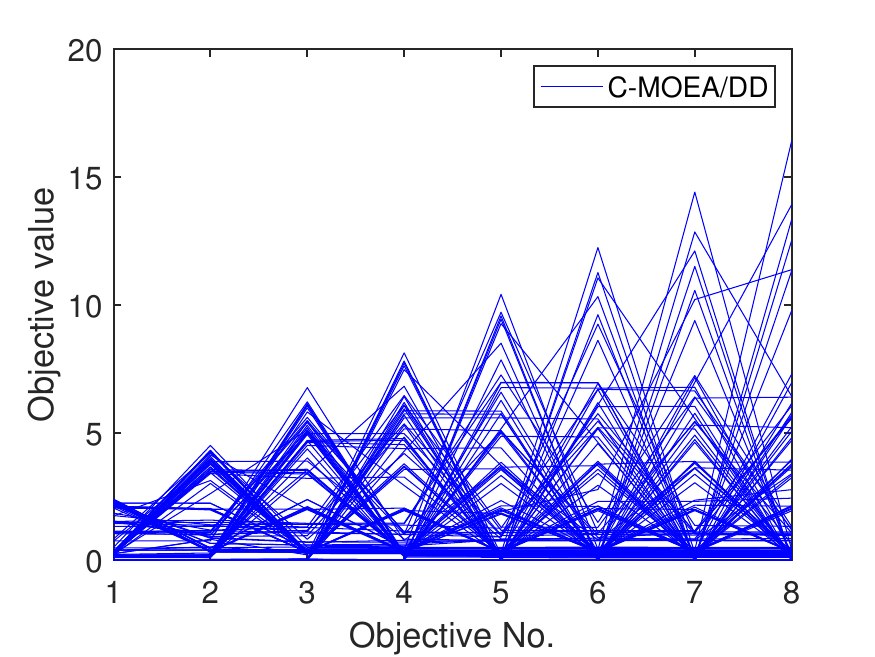}
\centerline{\scriptsize{(i) DAS-CMaOP8(7)}}
\end{minipage}
\begin{minipage}[t]{0.25\linewidth}
\includegraphics[width= 4.2cm]{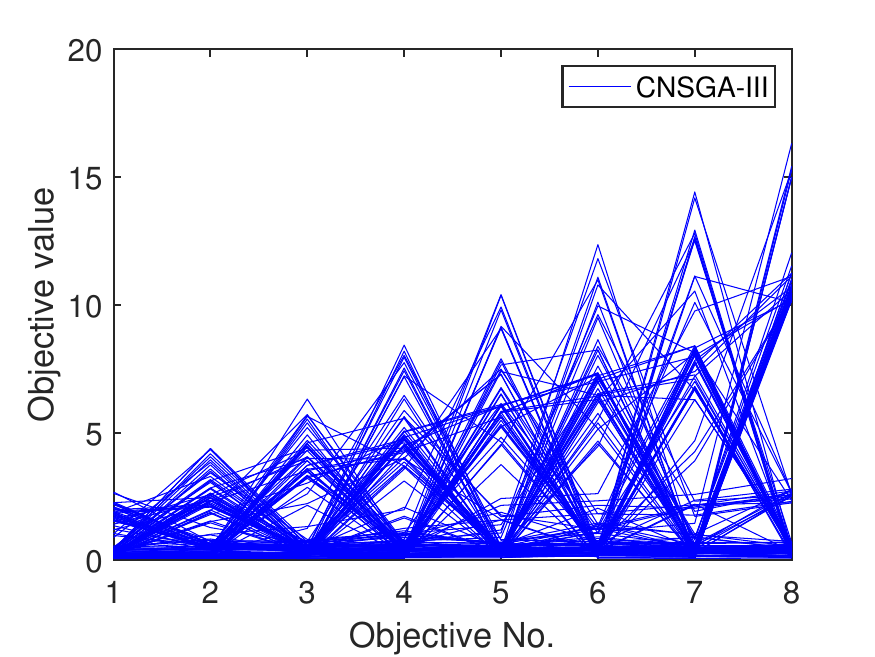}
\centerline{\scriptsize{(j) DAS-CMaOP8(7)}}
\end{minipage}
\begin{minipage}[t]{0.25\linewidth}
\includegraphics[width= 4.2cm]{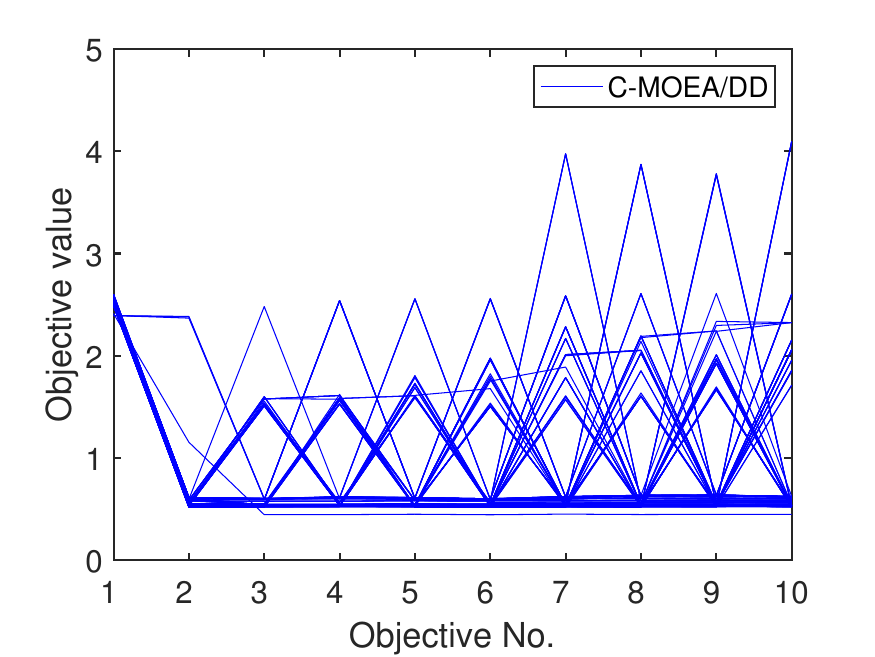}
\centerline{\scriptsize{(k) DAS-CMaOP5(11)}}
\end{minipage}
\begin{minipage}[t]{0.25\linewidth}
\includegraphics[width= 4.2cm]{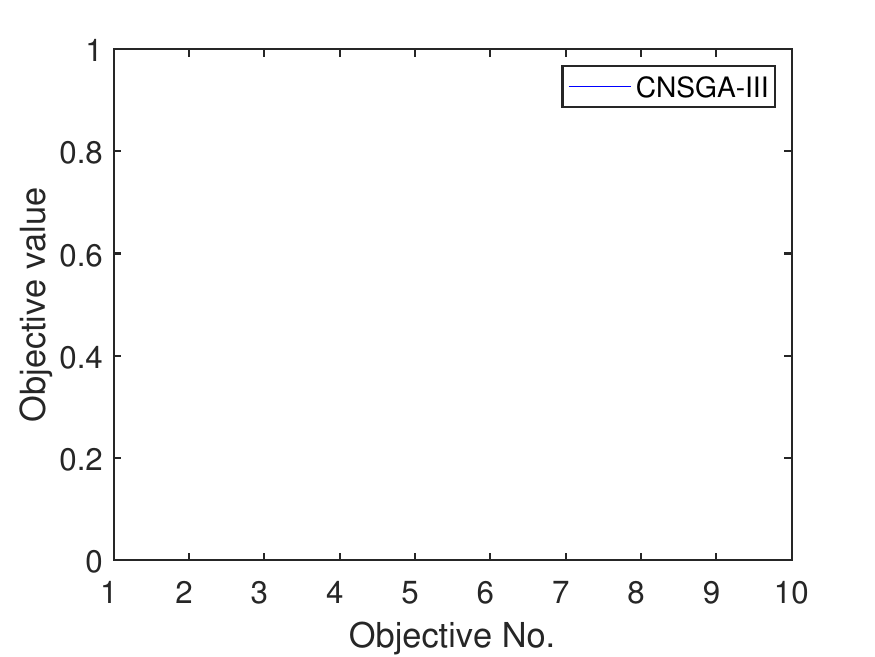}
\centerline{\scriptsize{(l) DAS-CMaOP5(11)}}
\end{minipage}
\end{tabular}

\caption{Parallel coordinate plots of the feasible and non-dominated solutions with the median $HV$ values in 20 independent runs using C-MOEA/DD and C-NSGA-III on DAS-CMaOP1-3, DAS-CMaOP5 and DAS-CMaOP8 with different difficulty triplets. (a)-(f) show the populations achieved by C-MOEA/DD and C-NSGA-III on DAS-CMaOPs with feasibility-hardness. (g)-(l) show the populations achieved by C-MOEA/DD and C-NSGA-III on DAS-CMOPs with convergence-hardness. It is worth noting that no feasible solutions have been found in (c) and (l).} \label{Fig:DAS-CMaOPs}
\end{figure*}

Figure \ref{Fig:DAS-CMaOPs} shows the median $HV$ values of the feasible and non-dominated solutions in 20 independent runs by using C-MOEA/DD and C-NSGA-III on DAS-CMaOP1-3, DAS-CMaOP5 and DAS-CMaOP8 with different difficulty triplets. We can see that according to the $HV$ measure, C-NSGA-III is more suitable for solving feasibility-hard DAS-CMaOPs, while C-MOEA/DD is more suitable for solving convergence-hard DAS-CMaOPs.

\subsection{Analysis of Experimental Results}

From the above performance comparisons on DAS-CMOP1-9 and DAS-CMaOP1-9 with a set of difficulty triplets, it is clear that each type of constraint function generated corresponding difficulties for MOEA/D-CDP, NSGA-II-CDP, C-MOEA/DD and C-NSGA-III. With the increase of each value in the difficulty triplet, the problems became more and more difficult for MOEA/D-CDP, NSGA-II-CDP, C-MOEA/DD and C-NSGA-III to solve.

Furthermore, it can be concluded that NSGA-II-CDP performed better than MOEA/D-CDP on most of the DAS-CMOPs with simultaneous diversity-, feasibility- and convergence-hardness. MOEA/D-CDP performed better than NSGA-II-CDP on most of the convergence-hard DAS-CMOPs. C-NSGA-III performed better than C-MOEA/DD on most DAS-CMaOPs with feasibility-hardness. C-MOEA/DD was significantly better than C-NSGA-III on most of the convergence-hard DAS-CMaOPs. Therefore, through comprehensive and systematic experiments, it is revealed that mechanisms of MOEA/D-CDP may be more effective in solving convergence-hard CMOPs, while mechanisms of NSGA-II-CDP may be more effective in solving CMOPs with simultaneous diversity-, feasibility- and convergence-hardness. It is also revealed that mechanisms in C-NSGA-III may be more effective in solving feasibility-hard CMaOPs, while mechanisms of C-MOEA/DD may be more effective in solving CMaOPs with convergence-hardness.

\begin{table}[htbp]
  \centering
  \caption{Means and standard deviations of $IGD$, $C$-metric and Spacing values obtained by MOEA/D-CDP and NSGA-II-CDP on DAS-CMOP1-5. Wilcoxon's rank sum test at 0.05 significance level is performed between MOEA/D-CDP and NSGA-II-CDP. $\dag$ and $\ddag$ denote that the performance of NSGA-II-CDP is significantly worse than or better than that of MOEA/D-CDP, respectively. DAS-CMOP1(i) means that DAS-CMOP1 has the $i$-th difficulty triplet in Table \ref{tab:dif_triplets}.}
  \scalebox{0.62}[0.62]{
  \setlength{\tabcolsep}{2mm}
    \begin{tabular}{l|cc|cc|cc}
    \toprule
    \multirow{2}[0]{*}{Test Problem} & \multicolumn{2}{c|}{$IGD$}       & \multicolumn{2}{c|}{$C$-metric}   & \multicolumn{2}{c}{Spacing} \\
    \cline{2-7}
    & \multicolumn{1}{c}{MOEA/D-CDP} & \multicolumn{1}{c|}{NSGA-II-CDP} & \multicolumn{1}{c}{MOEA/D-CDP} & \multicolumn{1}{c|}{NSGA-II-CDP} & \multicolumn{1}{c}{MOEA/D-CDP} & \multicolumn{1}{c}{NSGA-II-CDP} \\
    \hline
    DAS-CMOP1(1)& \bb{1.29E-03(1.51E-05)} & 3.70E-01(1.46E-02)$\dag$ & \bb{2.88E-01(9.42E-02)} & 1.11E-01(1.06E-01)$\dag$ & 2.97E-03(7.20E-05) & \bb{1.82E-04(1.33E-04)$\ddag$} \\
    DAS-CMOP1(2)& \bb{1.29E-02(4.35E-02)} & 2.90E-01(1.96E-02)$\dag$ & 1.25E-02(1.78E-02) & \bb{1.61E-01(2.69E-02)$\ddag$} & 5.80E-03(1.51E-03) & \bb{1.09E-03(3.38E-04)$\ddag$} \\
    DAS-CMOP1(3)& \bb{1.30E-03(7.62E-06)} & 3.77E-01(1.21E-02)$\dag$ & \bb{4.13E-01(1.07E-01)} & 1.70E-01(1.72E-01)$\dag$ & 1.37E-03(5.54E-06) & \bb{2.25E-04(1.39E-04)$\ddag$} \\
    DAS-CMOP1(4)& 4.27E-01(8.01E-02) & \bb{3.77E-01(2.06E-02)$\ddag$} & 4.50E-02(6.48E-02) & \bb{1.50E-01(8.24E-02)$\ddag$} & 1.35E-03(4.35E-04) & \bb{5.73E-04(5.36E-04)$\ddag$} \\
    DAS-CMOP1(5)& \bb{9.51E-02(1.57E-01)} & 3.64E-01(1.74E-02)$\dag$ & \bb{1.47E-01(1.34E-01)} & 1.24E-01(1.50E-01) & 2.34E-03(1.32E-03) & \bb{1.91E-04(1.52E-04)$\ddag$} \\
    DAS-CMOP1(6)& \bb{4.37E-03(2.85E-03)} & 2.84E-01(1.96E-02)$\dag$ & \bb{2.90E-01(8.34E-02)} & 1.48E-01(3.01E-02)$\dag$ & 5.61E-03(4.58E-04) & \bb{1.10E-03(4.30E-04)$\ddag$} \\
    DAS-CMOP1(7)& \bb{1.23E-03(4.32E-06)} & 3.10E-01(6.86E-03)$\dag$ & \bb{1.64E-01(5.62E-02)} & 5.94E-02(1.03E-01)$\dag$ & 1.92E-03(3.12E-04) & \bb{2.23E-04(1.25E-04)$\ddag$} \\
    DAS-CMOP1(8)& 7.28E-01(9.83E-02) & \bb{7.09E-01(2.01E-02)$\ddag$} & 1.73E-01(2.96E-01) & \bb{1.81E-01(3.42E-02)$\ddag$} & \bb{1.83E-03(6.96E-04)} & 2.02E-03(2.19E-03)$\dag$ \\
    DAS-CMOP1(9)& \bb{3.57E-01(3.60E-06)} & 3.64E-01(2.08E-02)$\dag$ & \bb{4.05E-01(8.81E-02)} & 2.14E-01(1.06E-01)$\dag$ & \bb{1.19E-04(2.76E-05)} & 2.09E-04(4.81E-04) \\
    DAS-CMOP1(10) & \bb{5.79E-03(5.00E-03)} & 2.77E-01(1.88E-02)$\dag$ & \bb{3.26E-01(1.20E-01)} & 2.06E-01(2.80E-02)$\dag$ & 5.76E-03(1.22E-03) & \bb{1.38E-03(7.94E-04)$\ddag$} \\
    DAS-CMOP1(11) & \bb{9.82E-02(3.24E-04)} & 2.42E-01(5.90E-03)$\dag$ & \bb{2.14E-01(7.51E-02)} & 1.32E-01(1.40E-01)$\dag$ & 1.63E-03(2.69E-04) & \bb{2.46E-04(1.44E-04)$\ddag$} \\
    DAS-CMOP1(12) & 8.96E-01(4.96E-02) & \bb{8.67E-01(2.54E-02)$\ddag$} & \bb{3.33E-02(1.83E-01)} & 0.00E+00(0.00E+00) & \bb{0.00E+00(0.00E+00)} & 9.12E-08(5.00E-07)$\dag$ \\
    DAS-CMOP1(13) & 3.58E-01(2.26E-02) & \bb{3.46E-01(2.05E-02)$\ddag$} & 7.75E-04(2.01E-03) & \bb{8.89E-03(1.95E-02)$\ddag$} & \bb{7.79E-04(5.38E-04)} & 9.03E-04(8.73E-04) \\
    DAS-CMOP1(14) & 3.61E-01(1.63E-02) & \bb{3.43E-01(2.14E-02)} & 9.70E-03(3.47E-02) & \bb{3.02E-02(2.64E-02)$\ddag$} & \bb{4.95E-04(3.00E-04)} & 6.18E-04(1.17E-03) \\
    DAS-CMOP1(15) & 6.79E-01(9.58E-02) & \bb{6.28E-01(1.26E-02)} & \bb{3.33E-02(1.83E-01)} & 0.00E+00(0.00E+00) & \bb{0.00E+00(0.00E+00)} & 2.56E-05(1.16E-04)$\dag$ \\
    DAS-CMOP1(16) & 6.88E-01(1.20E-01) & \bb{5.70E-01(1.04E-02)} & \bb{4.67E-01(5.07E-01)} & 0.00E+00(0.00E+00)$\ddag$ & \bb{0.00E+00(0.00E+00)} & 1.29E-06(6.08E-06)$\dag$ \\
    \hline
    \hline
    DAS-CMOP2(1)& \bb{1.35E-03(9.02E-06)} & 3.35E-01(1.31E-02)$\dag$ & \bb{2.78E-01(5.03E-02)} & 1.24E-01(1.29E-01)$\dag$ & 4.60E-03(7.99E-05) & \bb{2.94E-04(6.34E-04)$\ddag$} \\
    DAS-CMOP2(2)& \bb{4.93E-03(6.35E-03)} & 2.43E-01(1.93E-02)$\dag$ & \bb{5.83E-01(1.76E-01)} & 1.57E-01(3.07E-02)$\dag$ & 5.55E-03(6.72E-04) & \bb{1.40E-03(6.96E-04)$\ddag$} \\
    DAS-CMOP2(3)& \bb{1.32E-03(7.80E-06)} & 3.36E-01(1.17E-02)$\dag$ & \bb{3.96E-01(9.90E-02)} & 1.32E-01(1.35E-01)$\dag$ & 3.46E-03(1.17E-04) & \bb{2.54E-04(1.56E-04)$\ddag$} \\
    DAS-CMOP2(4)& 2.59E-01(4.07E-02) & \bb{2.39E-01(1.47E-02)$\ddag$} & 1.26E-02(1.40E-02) & \bb{1.38E-01(3.07E-02)$\ddag$} & 1.49E-03(4.07E-04) & \bb{1.13E-03(9.84E-04)$\ddag$} \\
    DAS-CMOP2(5)& \bb{9.86E-02(1.50E-01)} & 3.34E-01(1.59E-02)$\dag$ & \bb{1.88E-01(2.02E-01)} & 1.69E-01(1.30E-01) & 3.59E-03(2.28E-03) & \bb{2.05E-04(1.13E-04)$\ddag$} \\
    DAS-CMOP2(6)& \bb{4.60E-03(6.92E-03)} & 2.34E-01(1.71E-02)$\dag$ & \bb{3.17E-01(1.32E-01)} & 1.72E-01(2.57E-02)$\dag$ & 5.53E-03(5.99E-04) & \bb{1.21E-03(4.00E-04)$\ddag$} \\
    DAS-CMOP2(7)& \bb{1.32E-03(7.01E-06)} & 3.55E-01(1.95E-02)$\dag$ & \bb{3.10E-01(8.77E-02)} & 7.02E-02(1.16E-01)$\dag$ & 3.44E-03(1.02E-04) & \bb{2.50E-04(1.51E-04)$\ddag$} \\
    DAS-CMOP2(8)& 3.39E-01(2.01E-01) & \bb{2.16E-01(2.09E-02)$\ddag$} & \bb{8.51E-01(3.48E-01)} & 1.77E-01(2.48E-02)$\dag$ & 1.22E-03(4.67E-04) & \bb{8.93E-04(1.33E-03)$\ddag$} \\
    DAS-CMOP2(9)& \bb{3.19E-01(6.00E-02)} & 3.31E-01(1.37E-02)$\dag$ & 1.51E-02(1.40E-02) & \bb{2.65E-01(1.07E-01)$\ddag$} & 2.24E-04(6.22E-04) & \bb{1.60E-04(9.21E-05)} \\
    DAS-CMOP2(10) & \bb{6.21E-03(7.87E-03)} & 2.35E-01(1.79E-02)$\dag$ & \bb{4.08E-01(1.03E-01)} & 1.49E-01(2.64E-02)$\dag$ & 5.66E-03(1.36E-03) & \bb{1.27E-03(7.24E-04)$\ddag$} \\
    DAS-CMOP2(11) & \bb{1.32E-03(6.25E-06)} & 5.40E-01(2.08E-01)$\dag$ & \bb{2.56E-01(8.50E-02)} & 6.98E-02(2.44E-01)$\dag$ & 3.46E-03(1.17E-04) & \bb{6.89E-04(7.83E-04)$\ddag$} \\
    DAS-CMOP2(12) & 4.54E-01(2.59E-01) & \bb{2.19E-01(2.43E-02)$\ddag$} & 8.06E-02(1.79E-01) & \bb{1.32E-01(3.81E-02)$\ddag$} & \bb{5.93E-04(5.39E-04)} & 1.37E-03(2.57E-03) \\
    DAS-CMOP2(13) & 3.10E-01(2.33E-02) & \bb{3.01E-01(1.82E-02)} & 3.97E-03(8.47E-03) & \bb{1.18E-02(1.79E-02)$\ddag$} & \bb{6.22E-04(3.23E-04)} & 1.44E-03(1.41E-03)$\ddag$ \\
    DAS-CMOP2(14) & 3.31E-01(1.70E-02) & \bb{3.25E-01(1.86E-02)} & 1.11E-02(1.86E-02) & \bb{3.30E-02(4.03E-02)$\ddag$} & 5.18E-04(3.00E-04) & \bb{4.02E-04(3.26E-04)} \\
    DAS-CMOP2(15) & 3.46E-01(1.16E-02) & \bb{3.43E-01(9.25E-03)} & \bb{3.27E-03(8.68E-03)} & 1.83E-02(1.82E-02)$\dag$ & 6.69E-04(4.55E-04 & \bb{6.24E-04(7.77E-04)} \\
    DAS-CMOP2(16) & 3.12E-01(1.36E-02) & \bb{3.09E-01(5.59E-03)} & 1.13E-02(2.08E-02) & \bb{2.88E-02(3.40E-02)$\ddag$} & 4.47E-04(3.20E-04) & \bb{3.25E-04(2.39E-04)} \\
    \hline
    \hline
    DAS-CMOP3(1)& \bb{3.10E-02(5.25E-02)} & 3.37E-01(5.00E-02)$\dag$ & \bb{2.08E-01(7.85E-02)} & 4.54E-02(1.83E-01)$\dag$ & 3.35E-03(2.82E-04) & \bb{1.28E-03(1.31E-03)$\ddag$} \\
    DAS-CMOP3(2)& \bb{1.57E-01(3.10E-02)} & 2.31E-01(1.47E-02)$\dag$ & \bb{2.55E-01(2.26E-01)} & 1.00E-01(1.51E-01)$\dag$ & 2.99E-03(6.78E-04) & \bb{2.89E-03(2.42E-03)$\ddag$} \\
    DAS-CMOP3(3)& \bb{2.23E-02(5.23E-02)} & 2.84E-01(1.50E-02)$\dag$ & 1.35E-01(8.22E-02) & \bb{2.48E-01(6.75E-02)$\ddag$} & 3.72E-03(4.87E-04) & \bb{6.41E-04(4.40E-04)$\ddag$} \\
    DAS-CMOP3(4)& \bb{2.92E-01(8.95E-04)} & 2.94E-01(4.35E-03)$\dag$ & \bb{7.83E-02(4.14E-02)} & 4.84E-02(4.46E-02)$\dag$ & 2.25E-03(4.79E-04) & \bb{1.71E-03(4.11E-03)$\ddag$} \\
    DAS-CMOP3(5)& \bb{9.97E-02(5.29E-02)} & 4.25E-01(1.06E-01)$\dag$ & \bb{2.51E-01(2.11E-01)} & 2.00E-01(4.07E-01)$\dag$ & 4.27E-03(8.49E-04) & NaN(NaN) \\
    DAS-CMOP3(6)& \bb{1.67E-01(2.58E-02)} & 2.30E-01(1.17E-02)$\dag$ & 3.11E-01(1.67E-01) & \bb{4.31E-01(2.30E-01)$\ddag$} & \bb{2.86E-03(5.53E-04)} & 2.93E-03(2.86E-03)$\dag$ \\
    DAS-CMOP3(7)& \bb{3.24E-02(6.16E-02)} & 2.67E-01(1.83E-02)$\dag$ & 1.30E-01(6.73E-02) & \bb{4.31E-01(8.87E-02)$\ddag$} & 3.60E-03(5.69E-04) & \bb{9.46E-04(8.74E-04)$\ddag$} \\
    DAS-CMOP3(8)& 3.94E-01(1.39E-01) & \bb{3.42E-01(2.84E-02)} & 2.79E-03(4.02E-03) & \bb{1.03E-01(9.18E-02)$\ddag$} & \bb{7.94E-04(7.01E-04)} & 1.05E-03(5.11E-03)$\dag$ \\
    DAS-CMOP3(9)& \bb{1.67E-01(8.28E-02)} & 3.08E-01(4.50E-02)$\dag$ & 2.29E-01(3.51E-01) & \bb{3.00E-01(4.66E-01)} & 1.35E-03(1.68E-03) & NaN(NaN) \\
    DAS-CMOP3(10) & \bb{1.70E-01(3.54E-02)} & 2.32E-01(2.10E-02)$\dag$ & 1.85E-01(7.23E-02) & \bb{3.35E-01(2.46E-01)} & \bb{2.81E-03(4.57E-04)} & 5.11E-03(3.37E-03) \\
    DAS-CMOP3(11) & \bb{6.83E-02(9.70E-02)} & 2.42E-01(2.56E-02)$\dag$ & 2.46E-02(5.21E-02) & \bb{2.85E-01(1.15E-01)$\ddag$} & 3.54E-03(8.60E-04) & \bb{9.97E-04(9.11E-04)$\ddag$} \\
    DAS-CMOP3(12) & 8.28E-01(2.38E-01) & \bb{7.95E-01(2.83E-01)} & \bb{1.33E-01(3.46E-01)} & 4.70E-02(1.85E-01) & 7.14E-05(2.29E-04) & \bb{2.47E-05(5.65E-05)} \\
    DAS-CMOP3(13) & 3.45E-01(7.69E-02) & \bb{3.39E-01(9.36E-02)} & 4.55E-02(8.47E-02) & \bb{4.71E-02(7.48E-02)} & 1.62E-03(1.19E-03) & NaN(NaN) \\
    DAS-CMOP3(14) & 4.48E-01(3.25E-02) & \bb{4.33E-01(4.47E-02)} & 3.64E-01(3.40E-01) & \bb{4.36E-02(2.89E-02)} & 1.41E-03(7.86E-04) & NaN(NaN) \\
    DAS-CMOP3(15) & 5.37E-01(2.72E-01) & \bb{3.03E-01(3.75E-02)$\ddag$} & \bb{5.74E-02(1.83E-01)} & 3.43E-02(3.44E-02)$\dag$ & \bb{6.58E-04(7.88E-04)} & 1.97E-03(2.25E-03)$\dag$ \\
    DAS-CMOP3(16) & 6.54E-01(1.69E-01) & \bb{4.73E-01(6.34E-02)$\ddag$} & \bb{5.33E-01(5.07E-01)} & 1.04E-01(3.04E-01)$\dag$ & 3.08E-05(7.80E-05) & NaN(NaN) \\
   \hline
   \hline
    DAS-CMOP4(1)& 5.00E-02(7.13E-02) & \bb{1.04E-03(3.40E-05)$\ddag$} & \bb{6.06E-01(4.92E-01)} & 9.22E-03(3.26E-02)$\dag$ & 3.26E-03(3.91E-04) & \bb{1.56E-03(1.05E-04)$\ddag$} \\
    DAS-CMOP4(2)& 1.81E-03(6.90E-04) & \bb{1.60E-03(9.90E-05)} & \bb{1.72E-01(4.20E-02)} & 1.06E-01(1.55E-02)$\ddag$ & 2.32E-03(7.85E-04) & \bb{1.73E-03(1.45E-04)$\ddag$} \\
    DAS-CMOP4(3)& \bb{8.94E-02(1.78E-01)} & 1.57E-01(9.91E-02)$\dag$ & 6.77E-02(2.16E-01) & \bb{1.36E-01(3.45E-01)} & \bb{2.37E-03(1.51E-03)} & 1.00E-02(1.03E-02)$\dag$ \\
    DAS-CMOP4(4)& 6.90E-02(6.98E-02) & \bb{6.94E-03(2.92E-02)$\ddag$} & \bb{2.78E-01(7.71E-02)} & 9.08E-02(2.40E-02)$\dag$ & 3.37E-03(5.60E-04) & \bb{1.28E-03(2.22E-04)$\ddag$} \\
    DAS-CMOP4(5)& 9.66E-02(9.33E-02) & \bb{9.91E-04(1.78E-05)$\ddag$} & \bb{8.43E-01(3.22E-01)} & 7.67E-03(1.89E-02)$\dag$ & 4.05E-03(3.51E-03) & \bb{1.14E-03(6.30E-05)$\ddag$} \\
    DAS-CMOP4(6)& 1.69E-03(4.57E-04) & \bb{1.58E-03(6.20E-05)} & 6.28E-02(2.89E-02) & \bb{1.71E-01(3.13E-02)$\ddag$} & 2.35E-03(5.04E-04) & \bb{1.76E-03(1.64E-04)$\ddag$} \\
    DAS-CMOP4(7)& \bb{1.13E-01(1.40E-01)} & 4.90E-01(2.37E-01)$\dag$ & 7.07E-01(3.29E-01) & \bb{8.56E-01(3.29E-01)$\ddag$} & \bb{3.28E-03(1.80E-03)} & 7.73E-03(4.91E-03)$\dag$ \\
    DAS-CMOP4(8)& 2.04E-01(2.70E-05) & \bb{1.11E-01(1.57E-01)$\ddag$} & \bb{1.12E-01(4.60E-02)} & 7.98E-02(6.18E-02) & \bb{2.95E-03(7.52E-04)} & 4.01E-03(6.26E-03)$\dag$ \\
    DAS-CMOP4(9)& 1.22E-01(1.54E-01) & \bb{5.40E-04(3.18E-05)$\ddag$} & \bb{2.15E-01(3.70E-01)} & 6.11E-03(2.69E-02)$\dag$ & 3.92E-03(3.17E-03) & \bb{7.60E-04(4.76E-05)$\ddag$} \\
    DAS-CMOP4(10) & \bb{1.48E-03(1.13E-04)} & 1.57E-03(6.01E-05)$\dag$ & 8.73E-02(2.59E-02) & \bb{9.89E-02(1.66E-02)$\ddag$} & 2.29E-03(2.43E-04) & \bb{1.72E-03(1.64E-04)$\ddag$} \\
    DAS-CMOP4(11) & \bb{2.47E-01(2.30E-01)} & 8.53E-01(3.27E-01)$\dag$ & \bb{7.46E-01(2.50E-01)} & 9.51E-02(1.94E-01)$\dag$ & \bb{4.04E-03(2.50E-03)} & 2.29E-02(9.07E-02)$\dag$ \\
    DAS-CMOP4(12) & 2.33E-01(6.95E-05) & 2.33E-01(1.48E-04) & 9.59E-03(1.11E-02) & \bb{1.55E-01(4.84E-02)$\ddag$} & 1.98E-03(1.40E-04) & \bb{2.25E-04(3.22E-05)$\ddag$} \\
    DAS-CMOP4(13) & \bb{1.44E-03(6.30E-05)} & 3.79E-03(7.67E-03)$\dag$ & \bb{7.25E-02(3.87E-02)} & 6.67E-03(5.32E-03)$\dag$ & 2.69E-03(3.65E-04) & \bb{2.52E-03(1.03E-03)$\ddag$} \\
    DAS-CMOP4(14) & \bb{1.59E-03(9.75E-05)} & 2.22E-02(6.76E-02)$\dag$ & \bb{1.72E-01(8.13E-02)} & 6.67E-03(6.06E-03)$\dag$ & 3.56E-03(2.01E-04) & \bb{1.63E-03(2.42E-03)$\ddag$} \\
    DAS-CMOP4(15) & \bb{2.24E-01(4.07E-05)} & 2.25E-01(1.23E-03)$\dag$ & \bb{1.12E-01(1.08E-01)} & 1.11E-03(3.20E-03)$\dag$ & 2.07E-03(3.64E-04) & \bb{1.41E-03(2.07E-03)$\ddag$} \\
    DAS-CMOP4(16) & \bb{2.72E-01(7.92E-05)} & 3.02E-01(9.15E-02)$\dag$ & \bb{4.40E-02(3.17E-02)} & 4.00E-03(2.07E-02)$\dag$ & 2.50E-03(2.13E-04) & \bb{6.99E-04(1.27E-03)$\ddag$} \\
   \hline
   \hline
    DAS-CMOP5(1)& 1.30E-03(8.22E-07) & \bb{1.19E-03(7.00E-05)$\ddag$} & \bb{1.66E-01(3.59E-02)} & 3.67E-03(3.08E-03)$\dag$ & 4.59E-03(3.39E-05) & \bb{2.19E-03(2.11E-04)$\ddag$} \\
    DAS-CMOP5(2)& 1.95E-03(1.91E-04) & \bb{1.48E-03(2.12E-05)$\ddag$} & \bb{3.44E-01(9.58E-02)} & 9.67E-03(3.54E-03)$\dag$ & 5.19E-03(7.01E-04) & \bb{2.28E-03(9.71E-05)$\ddag$} \\
    DAS-CMOP5(3)& \bb{1.29E-03(8.64E-07)} & 1.54E-03(3.93E-05)$\dag$ & \bb{1.70E-01(2.15E-02)} & 4.67E-03(2.85E-03)$\dag$ & 3.41E-03(6.50E-05) & \bb{2.48E-03(2.13E-04)$\ddag$} \\
    DAS-CMOP5(4)& 1.93E-03(2.19E-04) & \bb{1.08E-03(2.45E-05)$\ddag$} & \bb{3.29E-01(5.88E-02)} & 1.00E-02(5.81E-03)$\dag$ & 5.35E-03(5.06E-04) & \bb{1.89E-03(1.21E-04)$\ddag$} \\
    DAS-CMOP5(5)& 1.44E-03(1.42E-06) & \bb{1.05E-03(6.39E-05)$\ddag$} & \bb{8.65E-02(5.02E-02)} & 5.00E-03(3.25E-03)$\dag$ & 4.92E-03(5.17E-05) & \bb{1.73E-03(1.83E-04)$\ddag$} \\
    DAS-CMOP5(6)& 1.94E-03(3.12E-04) & \bb{1.48E-03(2.84E-05)$\ddag$} & \bb{1.70E-01(5.16E-02)} & 3.00E-03(1.60E-03)$\dag$ & 4.96E-03(6.67E-04) & \bb{2.32E-03(1.28E-04)$\ddag$} \\
    DAS-CMOP5(7)& \bb{1.29E-03(1.25E-06)} & 1.52E-03(4.73E-05)$\dag$ & \bb{1.50E-01(2.67E-02)} & 5.11E-03(4.77E-03)$\dag$ & 3.40E-03(7.56E-05) & \bb{2.40E-03(1.27E-04)$\ddag$} \\
    DAS-CMOP5(8)& 2.03E-03(1.48E-04) & \bb{1.02E-03(2.15E-05)$\ddag$} & \bb{3.64E-01(7.98E-02)} & 5.86E-02(1.20E-02)$\dag$ & 5.18E-03(3.79E-04) & \bb{1.50E-03(8.35E-05)$\ddag$} \\
    DAS-CMOP5(9)& 1.22E-03(1.10E-06) & \bb{5.94E-04(4.78E-05)$\ddag$} & \bb{3.21E-01(9.32E-02)} & 2.89E-03(1.15E-03)$\dag$ & 3.50E-03(3.51E-05) & \bb{1.10E-03(1.28E-04)$\ddag$} \\
    DAS-CMOP5(10) & 1.92E-03(3.83E-04) & \bb{1.48E-03(3.41E-05)$\ddag$} & \bb{2.16E-01(3.17E-02)} & 0.00E+00(0.00E+00)$\dag$ & 4.75E-03(6.92E-04) & \bb{2.26E-03(1.37E-04)$\ddag$} \\
    DAS-CMOP5(11) & \bb{1.29E-03(8.23E-07)} & 1.51E-03(4.05E-05)$\dag$ & \bb{1.23E-01(2.95E-02)} & 8.89E-03(5.20E-03)$\dag$ & 3.41E-03(6.89E-05) & \bb{2.41E-03(1.42E-04)$\ddag$} \\
    DAS-CMOP5(12) & \bb{1.17E-01(3.15E-01)} & 1.39E-02(4.24E-04) & \bb{1.52E-01(7.85E-02)} & 7.84E-02(1.26E-02)$\dag$ & 4.03E-03(1.40E-03) & \bb{1.05E-03(4.64E-05)$\ddag$} \\
    DAS-CMOP5(13) & 1.76E-03(1.91E-04) & \bb{1.50E-03(2.98E-05)$\ddag$} & \bb{1.36E-01(6.73E-02)} & 1.67E-03(5.52E-03)$\dag$ & 4.54E-03(7.31E-04) & \bb{2.37E-03(1.23E-04)$\ddag$} \\
    DAS-CMOP5(14) & 1.87E-03(2.08E-04) & \bb{9.73E-04(2.67E-05)$\ddag$} & \bb{2.91E-01(1.42E-01)} & 3.78E-03(3.36E-03)$\dag$ & 4.94E-03(3.61E-04) & \bb{1.45E-03(9.60E-05)$\ddag$} \\
    DAS-CMOP5(15) & 3.02E-01(9.47E-02) & \bb{6.26E-02(9.70E-02)$\ddag$} & \bb{1.22E-01(1.16E-01)} & 2.22E-03(2.53E-03)$\dag$ & 3.54E-03(8.95E-04) & \bb{1.50E-03(2.63E-04)$\ddag$} \\
    DAS-CMOP5(16) & 3.32E-01(1.04E-01) & \bb{6.70E-02(8.60E-02)$\ddag$} & \bb{3.24E-02(7.80E-02)} & 3.78E-03(3.00E-03)$\dag$ & 2.89E-03(1.01E-03) & \bb{8.97E-04(1.00E-04)$\ddag$} \\
    \bottomrule
    \end{tabular}%
    }
  \label{tab:dascmop1-5}%
\end{table}%

\begin{table}[htbp]
  \centering
  \caption{Means and standard deviations of $IGD$, $C$-metric and Spacing values obtained by MOEA/D-CDP and NSGA-II-CDP on DAS-CMOP6-9 with sixteen different difficulty triplets. Wilcoxon's rank sum test at 0.05 significance level is performed between MOEA/D-CDP and NSGA-II-CDP. $\dag$ and $\ddag$ denote that the performance of NSGA-II-CDP is significantly worse than or better than that of MOEA/D-CDP, respectively. DAS-CMOP6(i) means that DAS-CMOP6 has the $i$-th difficulty triplet in Table \ref{tab:dif_triplets}.}
  \scalebox{0.65}[0.65]{
  \begin{tabular}{l|cc|cc|cc}
    \toprule
    \multirow{2}[0]{*}{Test Problem} & \multicolumn{2}{c|}{$IGD$}       & \multicolumn{2}{c|}{$C$-metric}   & \multicolumn{2}{c}{Spacing} \\
    \cline{2-7}
    & \multicolumn{1}{c}{MOEA/D-CDP} & \multicolumn{1}{c|}{NSGA-II-CDP} & \multicolumn{1}{c}{MOEA/D-CDP} & \multicolumn{1}{c|}{NSGA-II-CDP} & \multicolumn{1}{c}{MOEA/D-CDP} & \multicolumn{1}{c}{NSGA-II-CDP} \\
    \hline
    DAS-CMOP6(1)& 8.21E-02(1.29E-01) & \bb{2.66E-02(3.30E-02)$\ddag$} & \bb{2.14E-01(3.38E-01)} & 5.27E-02(7.57E-02) & 3.23E-03(9.82E-04) & \bb{5.18E-04(1.28E-04)$\ddag$} \\
    DAS-CMOP6(2)& 6.16E-02(7.09E-02) & \bb{1.54E-02(2.12E-02)$\ddag$} & 9.03E-02(5.92E-02) & \bb{1.40E-01(4.94E-02)$\ddag$} & 3.64E-03(1.29E-03) & \bb{1.80E-03(2.28E-03)$\ddag$} \\
    DAS-CMOP6(3)& \bb{7.76E-02(1.47E-01)} & 1.28E-01(1.23E-01) & 4.41E-01(3.60E-01) & \bb{7.69E-01(3.45E-01)$\ddag$} & \bb{3.95E-03(9.34E-04)} & 1.51E-02(3.69E-02) \\
    DAS-CMOP6(4)& 1.72E-01(2.08E-01) & \bb{3.85E-02(4.75E-02)$\ddag$} & 9.74E-02(1.12E-01) & \bb{1.90E-01(8.09E-02)$\ddag$} & 3.97E-03(9.16E-03) & \bb{1.72E-03(3.03E-03)$\ddag$} \\
    DAS-CMOP6(5)& 1.35E-01(8.55E-02) & \bb{5.86E-02(4.18E-02)$\ddag$} & \bb{3.01E-01(3.16E-01)} & 1.85E-01(3.04E-01) & \bb{4.01E-03(3.87E-03)} & 3.77E-02(4.53E-03)$\ddag$ \\
    DAS-CMOP6(6)& 3.47E-02(4.22E-02) & \bb{1.37E-02(1.86E-02)$\ddag$} & \bb{2.00E-01(6.61E-02)} & 1.84E-01(3.84E-02) & 3.37E-03(4.50E-04) & \bb{1.43E-03(5.40E-04)$\ddag$} \\
    DAS-CMOP6(7)& \bb{2.23E-01(3.76E-01)} & 5.31E-01(2.43E-01)$\dag$ & \bb{2.26E-01(3.42E-01)} & 1.64E-01(1.86E-01) & \bb{3.88E-03(1.67E-03)} & 1.07E-02(7.54E-03)$\ddag$ \\
    DAS-CMOP6(8)& 6.50E-01(6.44E-01) & \bb{1.90E-01(2.90E-01)$\ddag$} & \bb{4.81E-01(4.73E-01)} & 6.08E-02(5.83E-02)$\dag$ & \bb{1.06E-03(1.10E-03)} & 3.10E-02(1.30E-02)$\dag$ \\
    DAS-CMOP6(9)& 9.57E-02(1.50E-01) & \bb{7.74E-03(1.58E-02)} & 2.28E-01(3.23E-01) & \bb{3.60E-01(3.58E-01)} & \bb{6.89E-03(3.74E-02)} & 1.39E-01(5.51E-02)$\dag$ \\
    DAS-CMOP6(10) & \bb{3.25E-02(4.72E-02)} & 3.45E-02(3.47E-02) & \bb{1.75E-01(5.19E-02)} & 8.73E-02(2.78E-02)$\dag$ & 3.57E-03(6.32E-04) & \bb{1.96E-03(3.01E-03)$\ddag$} \\
    DAS-CMOP6(11) & \bb{5.43E-01(6.27E-01)} & 7.82E-01(3.34E-01) & \bb{5.67E-01(5.04E-01)} & 1.02E-01(1.14E-01)$\dag$ & \bb{3.61E-03(9.31E-04)} & 1.28E-02(1.38E-02)$\dag$ \\
    DAS-CMOP6(12) & 7.81E-01(5.56E-01) & \bb{4.92E-01(5.50E-01)} & \bb{1.45E-01(2.99E-01)} & 1.04E-01(1.02E-01) & 4.83E-04(8.77E-04) & NaN(NaN) \\
    DAS-CMOP6(13) & \bb{7.07E-03(2.77E-02)} & 3.86E-02(4.60E-02)$\dag$ & \bb{1.28E-01(8.68E-02)} & 7.89E-03(1.54E-02)$\dag$ & 4.11E-03(2.77E-04) & \bb{1.09E-03(1.70E-04)$\ddag$} \\
    DAS-CMOP6(14) & \bb{3.65E-02(1.85E-02)} & 9.77E-02(7.51E-02)$\dag$ & \bb{7.30E-01(1.79E-01)} & 1.64E-01(4.96E-02)$\dag$ & \bb{8.29E-04(4.75E-04)} & 3.12E-02(7.03E-03)$\dag$ \\
    DAS-CMOP6(15) & 4.63E-01(6.07E-01) & \bb{3.22E-02(4.47E-02)} & \bb{6.33E-01(4.90E-01)} & 7.67E-03(6.07E-03)$\dag$ & 2.49E-03(1.99E-03) & \bb{2.00E-03(2.08E-03)} \\
    DAS-CMOP6(16) & 5.21E-01(6.09E-01) & \bb{2.34E-01(2.46E-01)} & \bb{5.59E-02(5.61E-02)} & 2.36E-03(3.01E-03)$\dag$ & \bb{6.09E-04(6.75E-04)} & 2.43E-02(1.31E-02)$\dag$ \\
    \hline
    \hline
    DAS-CMOP7(1)& 3.27E-01(5.19E-01) & \bb{2.90E-02(5.96E-03)$\ddag$} & \bb{4.69E-02(1.71E-01)} & 1.73E-02(5.07E-02)$\dag$ & 3.19E-02(4.81E-03) & \bb{2.56E-02(1.31E-03)$\ddag$} \\
    DAS-CMOP7(2)& 5.10E-02(5.35E-02) & \bb{3.00E-02(6.90E-04)$\ddag$} & 3.30E-02(2.70E-02) & \bb{8.22E-02(1.20E-02)$\ddag$} & 3.23E-02(2.70E-03) & \bb{2.43E-02(1.64E-03)$\ddag$} \\
    DAS-CMOP7(3)& 1.02E-01(1.60E-01) & \bb{3.63E-02(6.98E-03)} & 4.46E-02(8.24E-02) & \bb{1.92E-01(1.42E-01)$\ddag$} & 2.81E-02(2.91E-03) & \bb{2.46E-02(1.25E-03)$\ddag$} \\
    DAS-CMOP7(4)& 4.24E-01(2.09E-01) & \bb{2.63E-02(6.39E-04)$\ddag$} & \bb{7.86E-01(2.90E-01)} & 8.00E-02(1.36E-02)$\dag$ & 3.66E-02(4.41E-03) & \bb{2.52E-02(9.84E-04)$\ddag$} \\
    DAS-CMOP7(5)& 1.02E+00(7.46E-01) & \bb{2.56E-02(5.98E-03)$\ddag$} & \bb{7.20E-01(4.24E-01)} & 2.37E-02(1.62E-02)$\dag$ & 2.84E-02(1.12E-02) & \bb{2.42E-02(1.44E-03)$\ddag$} \\
    DAS-CMOP7(6)& 2.98E-02(7.19E-04) & \bb{2.97E-02(8.50E-04)} & 4.46E-02(1.91E-02) & \bb{7.47E-02(1.82E-02)$\ddag$} & 3.09E-02(1.88E-03) & \bb{2.44E-02(1.32E-03)$\ddag$} \\
    DAS-CMOP7(7)& 1.48E-01(2.59E-01) & \bb{4.36E-02(1.23E-02)} & \bb{6.79E-01(4.26E-01)} & 5.28E-02(9.86E-02)$\dag$ & 2.99E-02(4.08E-03) & \bb{2.34E-02(1.89E-03)$\ddag$} \\
    DAS-CMOP7(8)& 1.02E-01(7.00E-02) & \bb{2.25E-02(7.88E-04)$\ddag$} & \bb{5.51E-01(3.50E-01)} & 1.01E-01(1.76E-02)$\dag$ & 3.18E-02(4.07E-03) & \bb{2.33E-02(1.43E-03)$\ddag$} \\
    DAS-CMOP7(9)& 7.47E-01(1.03E+00) & \bb{1.94E-02(3.37E-03)$\ddag$} & \bb{2.83E-01(4.14E-01)} & 4.44E-03(6.91E-03) & 3.00E-02(1.39E-02) & \bb{2.15E-02(2.07E-03)$\ddag$} \\
    DAS-CMOP7(10) & \bb{3.01E-02(1.12E-03)} & 3.02E-02(7.46E-04) & 2.40E-02(2.44E-02) & \bb{6.78E-02(1.39E-02)$\ddag$} & 3.14E-02(2.32E-03) & \bb{2.46E-02(1.20E-03)$\ddag$} \\
    DAS-CMOP7(11) & 1.19E-01(1.66E-01) & \bb{4.49E-02(1.65E-02)} & 1.71E-01(2.74E-01) & \bb{4.52E-01(3.32E-01)$\ddag$} & 2.91E-02(1.66E-03) & \bb{2.17E-02(1.21E-03)$\ddag$} \\
    DAS-CMOP7(12) & 2.61E-02(7.97E-03) & \bb{1.74E-02(1.66E-03)$\ddag$} & 7.77E-02(3.20E-02) & \bb{1.08E-01(1.54E-02)$\ddag$} & 2.73E-02(2.91E-03) & \bb{2.06E-02(1.92E-03)$\ddag$} \\
    DAS-CMOP7(13) & \bb{3.05E-02(2.88E-03)} & 4.09E-02(4.65E-02)$\dag$ & \bb{1.13E-02(1.96E-02)} & 2.22E-04(8.46E-04)$\dag$ & 3.42E-02(4.44E-03) & \bb{2.43E-02(2.64E-03)$\ddag$} \\
    DAS-CMOP7(14) & \bb{2.43E-02(9.99E-04)} & 5.27E-02(7.61E-02) & \bb{1.45E-02(1.65E-02)} & 2.22E-04(8.46E-04)$\dag$ & 3.30E-02(1.69E-03) & \bb{2.18E-02(4.17E-03)$\ddag$} \\
    DAS-CMOP7(15) & \bb{3.04E-02(1.21E-03)} & 3.12E-02(1.24E-02)$\dag$ & \bb{8.90E-02(5.28E-02)} & 0.00E+00(0.00E+00)$\dag$ & 3.47E-02(2.24E-03) & \bb{2.46E-02(1.84E-03)$\ddag$} \\
    DAS-CMOP7(16) & \bb{2.41E-02(1.05E-03)} & 3.82E-02(5.57E-02)$\dag$ & \bb{9.54E-03(5.50E-03)} & 0.00E+00(0.00E+00)$\dag$ & 3.26E-02(1.38E-03) & \bb{2.29E-02(3.12E-03)$\ddag$} \\
    \hline
    \hline
    DAS-CMOP8(1)& 8.95E-02(9.95E-02) & \bb{3.58E-02(3.50E-03)$\ddag$} & \bb{1.15E-01(2.01E-01)} & 9.67E-03(9.28E-03) & 4.43E-02(4.22E-03) & \bb{3.13E-02(1.44E-03)$\ddag$} \\
    DAS-CMOP8(2)& 4.32E-02(5.06E-03) & \bb{3.90E-02(9.67E-04)$\ddag$} & \bb{1.29E-01(4.96E-02)} & 6.86E-02(1.75E-02)$\dag$ & 4.58E-02(4.76E-03) & \bb{3.05E-02(1.42E-03)$\ddag$} \\
    DAS-CMOP8(3)& 9.61E-02(1.13E-01) & \bb{4.44E-02(4.84E-03)} & 9.32E-02(1.19E-01) & \bb{3.31E-01(1.76E-01)$\ddag$} & 4.10E-02(4.20E-03) & \bb{3.23E-02(1.80E-03)$\ddag$} \\
    DAS-CMOP8(4)& 1.36E-01(8.71E-02) & \bb{3.37E-02(1.03E-03)$\ddag$} & \bb{9.41E-02(1.29E-01)} & 7.20E-02(1.43E-02)& 4.79E-02(7.01E-03) & \bb{3.14E-02(1.31E-03)$\ddag$} \\
    DAS-CMOP8(5)& 3.58E-01(2.45E-01) & \bb{3.03E-02(3.77E-03)$\ddag$} & 3.40E-01(4.57E-01) & \bb{3.88E-01(1.59E-01)} & 4.45E-02(6.95E-03) & \bb{2.91E-02(2.03E-03)$\ddag$} \\
    DAS-CMOP8(6)& 3.95E-02(1.53E-03) & \bb{3.94E-02(1.03E-03)} & 5.15E-02(5.52E-02) & \bb{7.42E-02(1.93E-02)$\ddag$} & 4.59E-02(2.95E-03) & \bb{3.06E-02(1.59E-03)$\ddag$} \\
    DAS-CMOP8(7)& 8.61E-02(7.46E-02) & \bb{5.10E-02(6.52E-03)$\ddag$} & 8.19E-02(1.22E-01) & \bb{1.44E-01(1.52E-01)$\ddag$} & 3.84E-02(3.16E-03) & \bb{3.08E-02(1.11E-03)$\ddag$} \\
    DAS-CMOP8(8)& 6.18E-02(3.66E-02) & \bb{2.86E-02(1.15E-03)$\ddag$} & 6.51E-02(6.02E-02) & \bb{8.81E-02(1.36E-02)$\ddag$} & 4.42E-02(5.54E-03) & \bb{2.89E-02(1.55E-03)$\ddag$} \\
    DAS-CMOP8(9)& 6.23E-01(7.87E-01) & \bb{2.29E-02(2.81E-03)$\ddag$} & \bb{3.48E-01(4.42E-01)} & 2.39E-01(9.66E-02) & 2.85E-02(1.76E-02) & \bb{2.69E-02(2.56E-03)} \\
    DAS-CMOP8(10) & 3.95E-02(1.61E-03) & \bb{3.90E-02(1.18E-03)} & 2.96E-02(1.40E-02) & \bb{6.34E-02(1.92E-02)$\ddag$} & 4.61E-02(2.52E-03) & \bb{3.05E-02(1.78E-03)$\ddag$} \\
    DAS-CMOP8(11) & 1.26E-01(1.29E-01) & \bb{5.48E-02(1.07E-02)$\ddag$} & \bb{9.37E-01(2.38E-01)} & 4.32E-01(1.91E-01)$\dag$ & 3.68E-02(5.36E-03) & \bb{2.69E-02(1.68E-03)$\ddag$} \\
    DAS-CMOP8(12) & 3.26E-02(7.91E-03) & \bb{2.05E-02(1.34E-03)$\ddag$} & \bb{1.84E-01(3.79E-02)} & 7.98E-02(2.09E-02)$\dag$ & 4.32E-02(3.03E-03) & \bb{2.59E-02(1.91E-03)$\ddag$} \\
    DAS-CMOP8(13) & \bb{3.90E-02(1.32E-03)} & 5.69E-02(6.05E-02) & \bb{4.25E-02(5.14E-02)} & 2.89E-03(4.08E-03)$\dag$ & 4.93E-02(3.15E-03) & \bb{2.99E-02(4.63E-03)$\ddag$} \\
    DAS-CMOP8(14) & \bb{3.42E-02(2.24E-03)} & 7.04E-02(1.12E-01)$\dag$ & \bb{1.28E-01(6.76E-02)} & 0.00E+00(0.00E+00)$\dag$ & 4.72E-02(2.14E-03) & \bb{2.64E-02(7.02E-03)$\ddag$} \\
    DAS-CMOP8(15) & \bb{3.91E-02(1.99E-03)} & 6.38E-02(9.68E-02) & \bb{3.26E-02(3.10E-02)} & 3.78E-03(3.58E-03)$\dag$ & 4.93E-02(4.92E-03) & \bb{3.10E-02(3.43E-03)$\ddag$} \\
    DAS-CMOP8(16) & \bb{3.34E-02(1.79E-03)} & 8.14E-02(1.24E-01)$\dag$ & \bb{6.53E-02(3.61E-02)} & 0.00E+00(0.00E+00)$\dag$ & 4.72E-02(2.74E-03) & \bb{2.60E-02(6.48E-03)$\ddag$} \\
    \hline
    \hline
    DAS-CMOP9(1)& \bb{1.55E-01(2.68E-01)} & 4.60E-01(8.73E-02)$\dag$ & 6.59E-02(4.45E-02) & \bb{1.39E-01(8.20E-02)$\ddag$} & 3.57E-02(1.61E-02) & \bb{8.16E-03(5.20E-03)$\ddag$} \\
    DAS-CMOP9(2)& \bb{4.92E-02(2.62E-03)} & 1.63E-01(6.57E-02)$\dag$ & \bb{2.67E-01(1.59E-01)} & 3.81E-02(9.37E-03)$\dag$ & 6.28E-02(3.11E-03) & \bb{2.38E-02(5.98E-03)$\ddag$} \\
    DAS-CMOP9(3)& \bb{3.87E-02(2.39E-04)} & 4.38E-01(8.90E-02)$\dag$ & \bb{1.06E-01(2.48E-02)} & 1.01E-01(7.73E-02) & 3.66E-02(5.93E-04) & \bb{8.68E-03(5.48E-03)$\ddag$} \\
    DAS-CMOP9(4)& \bb{4.85E-02(2.70E-03)} & 2.22E-01(9.75E-02)$\dag$ & \bb{2.91E-01(8.59E-02)} & 4.58E-02(1.34E-02)$\dag$ & 6.09E-02(3.30E-03) & \bb{2.03E-02(6.77E-03)$\ddag$} \\
    DAS-CMOP9(5)& \bb{3.31E-02(5.93E-04)} & 4.54E-01(1.09E-01)$\dag$ & \bb{1.23E-01(3.24E-02)} & 1.03E-01(6.75E-02) & 4.28E-02(7.89E-04) & \bb{8.56E-03(6.05E-03)$\ddag$} \\
    DAS-CMOP9(6)& \bb{4.84E-02(2.33E-03)} & 1.91E-01(7.97E-02)$\dag$ & \bb{2.29E-01(1.78E-01)} & 4.48E-02(1.65E-02)$\dag$ & 6.24E-02(2.47E-03) & \bb{2.09E-02(5.74E-03)$\ddag$} \\
    DAS-CMOP9(7)& \bb{4.06E-02(3.18E-04)} & 4.29E-01(8.42E-02)$\dag$ & \bb{9.92E-02(2.21E-02)} & 3.42E-02(4.92E-02)$\dag$ & 3.58E-02(6.55E-04) & \bb{9.52E-03(5.45E-03)$\ddag$} \\
    DAS-CMOP9(8)& \bb{7.67E-02(1.03E-01)} & 2.27E-01(7.77E-02)$\dag$ & 4.36E-02(2.59E-02) & \bb{5.32E-02(1.74E-02)} & 5.77E-02(1.41E-02) & \bb{1.78E-02(5.99E-03)$\ddag$} \\
    DAS-CMOP9(9)& \bb{6.36E-02(1.44E-01)} & 4.29E-01(1.44E-01)$\dag$ & \bb{1.03E-01(5.31E-02)} & 3.40E-02(6.25E-02)$\dag$ & 3.81E-02(9.82E-03) & \bb{8.71E-03(8.06E-03)$\ddag$} \\
    DAS-CMOP9(10) & \bb{5.04E-02(3.16E-03)} & 1.66E-01(6.58E-02)$\dag$ & \bb{3.18E-01(1.09E-01)} & 4.97E-02(1.26E-02)$\dag$ & 6.33E-02(2.44E-03) & \bb{2.34E-02(6.24E-03)$\ddag$} \\
    DAS-CMOP9(11) & \bb{4.18E-02(4.87E-04)} & 4.63E-01(6.67E-02)$\dag$ & 1.19E-01(2.54E-02) & \bb{1.93E-01(1.09E-01)$\dag$} & 3.33E-02(8.43E-04) & \bb{8.24E-03(3.98E-03)$\ddag$} \\
    DAS-CMOP9(12) & \bb{2.66E-01(1.76E-01)} & 2.68E-01(9.44E-02) & 8.16E-03(1.11E-02) & \bb{7.88E-02(2.98E-02)$\ddag$} & 3.08E-02(2.42E-02) & \bb{1.42E-02(7.94E-03)$\ddag$} \\
    DAS-CMOP9(13) & \bb{4.65E-01(6.30E-02)} & 4.67E-01(4.86E-02) & \bb{9.10E-02(1.87E-01)} & 2.11E-03(5.14E-03) & 6.96E-03(3.67E-03) & \bb{6.22E-03(2.95E-03)} \\
    DAS-CMOP9(14) & 5.98E-01(7.62E-02) & \bb{5.34E-01(5.24E-02)$\ddag$} & \bb{1.86E-03(5.74E-03)} & 1.78E-03(4.93E-03) & \bb{2.16E-03(8.30E-04)} & 3.58E-03(2.86E-03)$\dag$ \\
    DAS-CMOP9(15) & 4.77E-01(7.26E-02) & \bb{4.75E-01(4.61E-02)} & \bb{2.32E-01(1.89E-01)} & 2.22E-04(8.46E-04)$\dag$ & \bb{6.13E-03(3.27E-03)} & 6.36E-03(2.93E-03) \\
    DAS-CMOP9(16) & 6.05E-01(8.06E-02) & \bb{5.28E-01(4.95E-02)$\ddag$} & 0.00E+00(0.00E+00) & \bb{3.33E-03(6.00E-03)$\ddag$} & \bb{2.27E-03(1.53E-03)} & 3.79E-03(2.06E-03)$\ddag$ \\
    \bottomrule
    \end{tabular}%
    }
  \label{tab:dascmop6-9}%
\end{table}%

\begin{table}[htbp]
  \centering
  \caption{Means and standard deviations of $HV$ values obtained by C-MOEA/DD and C-NSGA-III on DAS-CMaOP1-6 with 5, 8 and 10 objectives. Wilcoxon's rank sum test at 0.05 significance level is performed between C-MOEA/DD and C-NSGA-III. $\dag$ and $\ddag$ denote that the performance of C-NSGA-III is significantly worse than or better than that of C-MOEA/DD, respectively. DAS-CMaOP1(i) means that DAS-CMaOP1 has the $i$-th difficulty triplet in Table \ref{tab:dif_triplets}.}
  \scalebox{0.65}[0.65]{
    \begin{tabular}{l|cc|cc|cc}
    \toprule
    \multirow{2}[0]{*}{Test Problem} & \multicolumn{2}{c|}{m = 5}       & \multicolumn{2}{c|}{m = 8}   & \multicolumn{2}{c}{m = 10} \\
    \cline{2-7}
    & \multicolumn{1}{c}{C-MOEA/DD} & \multicolumn{1}{c|}{C-NSGA-III} & \multicolumn{1}{c}{C-MOEA/DD} & \multicolumn{1}{c|}{C-NSGA-III} & \multicolumn{1}{c}{C-MOEA/DD} & \multicolumn{1}{c}{C-NSGA-III} \\
    \hline
    DAS-CMaOP1(1)& 6.62E-01(3.34E-02) & \bb{6.71E-01(3.23E-02)} & \bb{7.95E-01(2.54E-02)} & 7.63E-01(7.21E-02)$\dag$         & 7.70E-01(4.77E-02) & \bb{8.27E-01(9.18E-03)$\ddag$} \\
    DAS-CMaOP1(2)& 5.15E-01(7.02E-02) & \bb{5.81E-01(1.29E-02)$\ddag$} & 4.78E-01(6.98E-02) & \bb{5.27E-01(2.19E-02)$\ddag$}         & 4.94E-01(7.05E-03) & \bb{5.00E-01(1.11E-02)$\ddag$} \\
    DAS-CMaOP1(3)& 7.55E-01(2.56E-02) & \bb{7.67E-01(5.78E-02)} & \bb{8.47E-01(1.24E-02)} & 8.45E-01(2.54E-02)         & 8.48E-01(1.72E-02) & \bb{8.63E-01(7.47E-03)$\ddag$} \\
    DAS-CMaOP1(4)& 2.48E-01(1.35E-02) & \bb{4.16E-01(4.22E-02)$\ddag$}  & 2.83E-01(2.81E-02) & \bb{4.60E-01(1.95E-02)$\ddag$}         & 2.95E-01(2.19E-02) & \bb{4.85E-01(4.27E-03)$\ddag$} \\
    DAS-CMaOP1(5)& \bb{6.32E-01(3.20E-02)} & 6.04E-01(3.65E-02)$\dag$ & \bb{7.68E-01(3.94E-02)} & 7.09E-01(1.10E-01)$\dag$ & 7.03E-01(6.09E-02) & \bb{8.22E-01(1.19E-02)$\ddag$} \\
    DAS-CMaOP1(6)& 4.79E-01(5.89E-02) & \bb{5.74E-01(1.55E-02)$\ddag$} & 3.39E-01(4.90E-02) & \bb{5.29E-01(2.15E-02)$\ddag$} & 4.22E-01(5.90E-02) & \bb{4.95E-01(1.43E-02)$\ddag$} \\
    DAS-CMaOP1(7)& \bb{7.60E-01(2.18E-02)} & 7.53E-01(6.32E-02) & \bb{8.44E-01(2.29E-02)} & 8.38E-01(1.14E-02) & 6.91E-01(7.33E-02) & \bb{8.65E-01(8.12E-03)$\ddag$} \\
    DAS-CMaOP1(8)& 2.50E-01(1.52E-02) & \bb{3.36E-01(3.80E-02)$\ddag$} & 2.63E-01(1.31E-02) & \bb{4.32E-01(6.31E-03)$\ddag$} & 2.54E-01(1.39E-02) & \bb{3.97E-01(1.88E-03)$\ddag$} \\
    DAS-CMaOP1(9)& \bb{5.88E-01(3.49E-02)} & 5.41E-01(5.18E-02)$\dag$ & \bb{7.25E-01(7.12E-02)} & 6.59E-01(1.49E-01) & 5.89E-01(8.45E-02) & \bb{8.00E-01(8.06E-02)$\ddag$} \\
    DAS-CMaOP1(10) & 4.70E-01(5.73E-02) & \bb{5.68E-01(2.23E-02)$\ddag$} & 3.52E-01(5.77E-02) & \bb{4.99E-01(2.24E-02)$\ddag$} & 4.19E-01(4.25E-02) & \bb{4.89E-01(6.33E-03)$\ddag$} \\
    DAS-CMaOP1(11) & 1.72E-01(1.81E-03) & \bb{1.84E-01(1.07E-03)$\ddag$} & 1.06E-01(3.01E-03) & \bb{1.22E-01(1.18E-03)$\ddag$} & \bb{9.27E-02(1.19E-03)} & 9.10E-02(7.12E-04)$\dag$ \\
    DAS-CMaOP1(12) & \bb{2.41E-01(1.56E-02)} & 1.66E-01(5.86E-03)$\dag$ & \bb{2.48E-01(1.11E-02)} & 1.36E-01(4.71E-06)$\dag$ & \bb{2.44E-01(1.32E-02)} & 2.00E-01(4.25E-05)$\dag$ \\
    \hline
    \hline
    DAS-CMaOP2(1)& 8.09E-01(3.54E-03) & \bb{8.10E-01(2.93E-03)} & 8.06E-01(5.53E-03) & \bb{8.09E-01(1.12E-03)$\ddag$} & 8.07E-01(1.04E-03) & 8.07E-01(1.28E-03) \\
    DAS-CMaOP2(2)& 6.04E-01(4.25E-02) & \bb{6.20E-01(2.64E-02)$\ddag$} & 5.61E-01(1.67E-03) & \bb{5.66E-01(6.64E-05)$\ddag$} & 5.37E-01(7.37E-04) & \bb{5.38E-01(6.51E-05)$\ddag$} \\
    DAS-CMaOP2(3)& 9.61E-01(5.22E-02) & \bb{9.83E-01(3.93E-02)$\ddag$} & 9.68E-01(3.40E-03) & \bb{9.97E-01(8.16E-04)$\ddag$} & 9.65E-01(3.03E-03) & \bb{9.97E-01(8.92E-04)$\ddag$} \\
    DAS-CMaOP2(4)& 5.06E-01(1.08E-03) & \bb{5.07E-01(6.87E-04)$\ddag$} & 4.56E-01(7.10E-05) & 4.56E-01(6.66E-04)             & 4.33E-01(4.98E-05) & 4.33E-01(4.88E-04) \\
    DAS-CMaOP2(5)& 7.99E-01(4.74E-03) & \bb{8.03E-01(1.67E-03)$\ddag$} & 8.01E-01(2.16E-03) & 8.01E-01(2.38E-03)             & 8.01E-01(9.29E-04) & 8.01E-01(1.43E-03) \\
    DAS-CMaOP2(6)& \bb{6.15E-01(2.60E-02)} & 6.09E-01(4.32E-02)$\dag$ & 5.60E-01(1.85E-03) & \bb{5.66E-01(7.16E-04)$\ddag$} & 5.37E-01(6.81E-04) & \bb{5.38E-01(1.22E-04)$\ddag$} \\
    DAS-CMaOP2(7)& \bb{9.35E-01(7.55E-02)} & 8.94E-01(9.02E-02) & 9.69E-01(3.97E-03) & \bb{9.96E-01(1.01E-03)$\ddag$} & 9.63E-01(3.99E-03) & \bb{9.96E-01(7.62E-04)$\ddag$} \\
    DAS-CMaOP2(8)& \bb{4.98E-01(1.60E-03)} & 4.19E-01(7.53E-02)$\dag$ & \bb{3.95E-01(5.72E-02)} & 2.95E-01(9.37E-02)$\ddag$ & \bb{3.39E-01(4.82E-02)} & 2.75E-01(8.26E-02)$\dag$ \\
    DAS-CMaOP2(9)& \bb{7.82E-01(7.22E-03)} & 7.77E-01(3.78E-02) & 7.83E-01(3.83E-03) & 7.83E-01(3.94E-03)             & \bb{7.84E-01(1.14E-03)} & 7.83E-01(2.35E-03)\\
    DAS-CMaOP2(10) & 5.98E-01(4.66E-02) & \bb{6.03E-01(4.83E-02)$\ddag$} & 5.59E-01(2.40E-03) & \bb{5.66E-01(9.24E-05)$\ddag$} & 5.37E-01(6.80E-04) & \bb{5.38E-01(2.09E-04)$\ddag$} \\
    DAS-CMaOP2(11) & \bb{9.52E-01(9.90E-02)} & 3.00E-01(2.98E-01)$\dag$ & 1.09E-01(5.25E-02) & \bb{1.31E-01(2.24E-02)$\dag$} & \bb{5.48E-02(2.69E-02)} & 4.76E-02(4.84E-04) \\
    DAS-CMaOP2(12) & \bb{7.72E-01(1.75E-01)} & 4.59E-02(1.22E-02)$\dag$ & \bb{8.71E-01(6.59E-03)} & 2.12E-02(6.13E-03)$\dag$ & \bb{8.78E-01(7.53E-03)} & 1.18E-02(2.60E-03)$\dag$ \\
    \hline
    \hline
    DAS-CMaOP3(1)& 6.08E-01(1.96E-02) & \bb{6.46E-01(8.21E-03)$\ddag$} & 4.49E-01(2.83E-02) & \bb{5.32E-01(3.65E-02)$\ddag$} & 4.00E-01(1.14E-02) & \bb{5.39E-01(3.51E-02)$\ddag$} \\
    DAS-CMaOP3(2)& \bb{5.01E-01(1.40E-03)} & 4.98E-01(2.79E-03)$\dag$ & \bb{4.46E-01(2.57E-03)} & 4.24E-01(7.81E-03)$\dag$ & \bb{4.32E-01(1.23E-03)} & 4.21E-01(3.75E-03)$\dag$ \\
    DAS-CMaOP3(3)& 6.26E-01(5.72E-03) & \bb{6.56E-01(9.98E-03)$\ddag$} & 5.52E-01(5.39E-03) & \bb{5.73E-01(3.23E-02)$\ddag$} & 5.26E-01(4.73E-03) & \bb{6.26E-01(1.87E-02)$\ddag$} \\
    DAS-CMaOP3(4)& 4.59E-01(2.31E-03) & \bb{4.67E-01(3.64E-03)$\ddag$} & 3.77E-01(7.59E-03) & \bb{3.87E-01(3.61E-02)$\ddag$} & 3.65E-01(5.18E-03) & \bb{3.92E-01(1.68E-03)$\ddag$} \\
    DAS-CMaOP3(5)& 5.94E-01(2.11E-02) & \bb{6.40E-01(1.22E-02)$\ddag$} & 4.28E-01(3.13E-02) & \bb{5.45E-01(2.09E-02)$\ddag$} & 3.76E-01(2.10E-02) & \bb{5.14E-01(2.97E-02)$\ddag$} \\
    DAS-CMaOP3(6)& 4.93E-01(1.59E-03) & \bb{4.98E-01(2.14E-03)$\ddag$} & \bb{4.35E-01(1.37E-03)} & 4.34E-01(3.24E-03) & \bb{4.20E-01(1.19E-03)} & 4.18E-01(5.70E-03) \\
    DAS-CMaOP3(7)& \bb{6.55E-01(8.81E-03)} & 6.10E-01(3.73E-02)$\dag$ & 4.30E-01(7.21E-02) & \bb{4.71E-01(9.43E-02)} & \bb{4.13E-01(1.05E-01)} & 3.88E-01(4.90E-02) \\
    DAS-CMaOP3(8)& 1.88E-01(5.10E-02) & \bb{2.09E-01(2.89E-03)$\ddag$} & \bb{8.02E-02(5.76E-02)} & 1.57E-01(3.54E-02)$\dag$ & \bb{4.58E-01(7.15E-03)} & 1.30E-01(1.79E-03)$\dag$ \\
    DAS-CMaOP3(9)& 5.94E-01(1.69E-02) & \bb{6.26E-01(1.69E-02)$\ddag$} & 4.07E-01(2.63E-02) & \bb{5.13E-01(4.50E-02)$\ddag$} & 4.55E-01(6.59E-03) & \bb{4.95E-01(4.33E-02)$\ddag$} \\
	DAS-CMaOP3(10) & 4.88E-01(1.94E-03) & \bb{5.00E-01(8.22E-04)$\ddag$} & 4.26E-01(3.48E-03) & \bb{4.33E-01(5.10E-03)$\ddag$} & 4.11E-01(1.48E-03) & \bb{4.19E-01(3.03E-03)$\ddag$} \\
    DAS-CMaOP3(11) & \bb{1.35E-01(3.90E-02)} & 3.32E-02(3.25E-04)$\dag$ & \bb{5.19E-01(1.05E-01)} & 3.61E-02(2.63E-07)$\dag$ & \bb{5.31E-01(2.05E-02)} & 4.75E-02(4.35E-05)$\dag$ \\
    DAS-CMaOP3(12) & \bb{4.74E-01(6.69E-03)} & 3.19E-02(3.67E-04)$\dag$ & \bb{4.53E-01(4.77E-03)} & 2.93E-02(1.02E-04)$\dag$ & \bb{4.57E-01(7.30E-03)} & 2.16E-02(1.03E-04)$\dag$ \\
    \hline
    \hline
    DAS-CMaOP4(1)& 7.44E-01(2.53E-02) & \bb{7.54E-01(2.53E-02)$\ddag$} & 7.61E-01(6.33E-03) & \bb{8.06E-01(5.32E-03)$\ddag$} & 6.92E-01(1.47E-02) & \bb{8.06E-01(3.08E-03)$\ddag$} \\
    DAS-CMaOP4(2)& 5.02E-01(8.74E-04) & \bb{5.12E-01(6.31E-04)$\ddag$} & 4.80E-01(4.21E-03) & \bb{5.16E-01(3.65E-04)$\ddag$} & 4.46E-01(3.66E-03) & \bb{5.16E-01(2.29E-04)$\ddag$} \\
    DAS-CMaOP4(3)& 8.56E-01(2.36E-03) & \bb{8.60E-01(2.45E-03)$\ddag$} & 9.10E-01(4.02E-03) & \bb{9.20E-01(1.79E-02)$\ddag$} & 9.04E-01(5.81E-03) & \bb{9.43E-01(1.88E-02)$\ddag$} \\
    DAS-CMaOP4(4)& 4.27E-01(1.58E-02) & \bb{4.42E-01(1.68E-02)$\ddag$} & 4.09E-01(4.39E-03) & \bb{4.32E-01(8.14E-03)$\ddag$} & 3.93E-01(7.52E-03) & \bb{4.27E-01(2.51E-03)$\ddag$} \\
    DAS-CMaOP4(5)& 7.04E-01(2.54E-02) & \bb{7.19E-01(3.10E-02)$\ddag$} & 7.45E-01(1.01E-02) & \bb{7.85E-01(9.51E-03)$\ddag$} & 6.88E-01(1.43E-02) & \bb{7.85E-01(7.23E-03)$\ddag$} \\
    DAS-CMaOP4(6)& 5.02E-01(5.71E-04) & \bb{5.12E-01(6.58E-04)$\ddag$} & 4.80E-01(3.88E-03) & \bb{5.16E-01(3.30E-04)$\ddag$} & 4.48E-01(3.96E-03) & \bb{5.16E-01(1.89E-04)$\ddag$} \\
    DAS-CMaOP4(7)& \bb{8.48E-01(2.70E-03)} & 8.46E-01(3.70E-03) & \bb{9.07E-01(3.90E-03)} & 8.39E-01(1.46E-02)$\dag$ & \bb{9.05E-01(6.69E-03)} & 8.86E-01(1.27E-02)$\dag$ \\
    DAS-CMaOP4(8)& \bb{4.18E-01(1.09E-02)} & 3.81E-01(4.22E-02)$\dag$ & 3.89E-01(6.98E-03) & \bb{4.19E-01(1.26E-02)$\ddag$} & 3.78E-01(3.42E-02) & \bb{4.13E-01(4.30E-03)$\ddag$} \\
    DAS-CMaOP4(9)& \bb{6.61E-01(2.57E-02)} & 6.38E-01(2.31E-02)$\dag$ & 7.13E-01(1.27E-02) & \bb{7.43E-01(3.24E-02)$\ddag$} & 6.74E-01(1.46E-02) & \bb{7.65E-01(1.01E-02)$\ddag$} \\
	DAS-CMaOP4(10) & 5.02E-01(6.06E-04) & \bb{5.12E-01(5.06E-04)$\ddag$} & 4.80E-01(5.00E-03) & \bb{5.16E-01(2.96E-04)$\ddag$} & 4.47E-01(3.78E-03) & \bb{5.15E-01(2.54E-04)$\ddag$} \\
    DAS-CMaOP4(11) & \bb{6.96E-01(4.19E-02)} & 6.51E-01(1.58E-01) & \bb{2.72E-01(2.04E-02)} & 1.08E-01(9.64E-02)$\dag$ & \bb{1.57E-01(1.62E-01)} & 4.76E-02(4.17E-08)$\dag$ \\
    DAS-CMaOP4(12) & \bb{2.35E-01(6.89E-02)} & 2.80E-02(2.85E-03)$\dag$ & \bb{4.49E-01(1.59E-01)} & 2.44E-02(1.52E-02)$\dag$ & \bb{5.59E-01(1.84E-02)} & 1.44E-02(2.31E-05)$\dag$ \\
    \hline
    \hline
    DAS-CMaOP5(1)& 7.13E-01(1.25E-02) & \bb{7.26E-01(2.33E-02)$\ddag$} & 6.92E-01(1.20E-02) & \bb{7.67E-01(2.45E-03)$\ddag$} & 6.07E-01(1.32E-02) & \bb{7.69E-01(3.30E-03)$\ddag$} \\
    DAS-CMaOP5(2)& 4.89E-01(6.98E-04) & \bb{5.07E-01(8.76E-04)$\ddag$} & 4.71E-01(4.42E-03) & \bb{5.14E-01(7.73E-04)$\ddag$} & 4.40E-01(6.09E-03) & \bb{5.15E-01(2.47E-04)$\ddag$} \\
    DAS-CMaOP5(3)& 8.18E-01(2.61E-03) & \bb{8.34E-01(2.24E-03)$\ddag$} & 8.55E-01(4.33E-03) & \bb{8.97E-01(6.53E-04)$\ddag$} & 8.50E-01(2.85E-03) & \bb{9.19E-01(7.77E-04)$\ddag$} \\
    DAS-CMaOP5(4)& 4.26E-01(8.15E-03) & \bb{4.37E-01(1.64E-02)$\ddag$} & 3.87E-01(6.60E-03) & \bb{4.36E-01(1.92E-03)$\ddag$} & 3.53E-01(8.73E-03) & \bb{4.24E-01(3.13E-03)$\ddag$} \\
    DAS-CMaOP5(5)& 6.85E-01(2.83E-02) & \bb{7.08E-01(3.09E-02)$\ddag$} & 6.78E-01(1.01E-02) & \bb{7.56E-01(5.13E-03)$\ddag$} & 6.08E-01(1.30E-02) & \bb{7.55E-01(4.13E-03)$\ddag$} \\
    DAS-CMaOP5(6)& 4.90E-01(1.39E-03) & \bb{5.06E-01(1.24E-03)$\ddag$} & 4.68E-01(6.31E-03) & \bb{5.14E-01(6.47E-04)$\ddag$} & 4.40E-01(3.46E-03) & \bb{5.15E-01(2.48E-04)$\ddag$} \\
    DAS-CMaOP5(7)& 8.12E-01(2.65E-03) & \bb{8.23E-01(2.54E-03)$\ddag$} & 8.51E-01(4.07E-03) & \bb{8.86E-01(7.82E-03)$\ddag$} & 8.47E-01(4.67E-03) & \bb{9.09E-01(7.52E-03)$\ddag$} \\
    DAS-CMaOP5(8)& 4.12E-01(1.57E-02) & \bb{4.23E-01(1.61E-02)$\ddag$} & 3.78E-01(5.85E-03) & \bb{4.32E-01(1.50E-03)$\ddag$} & 3.46E-01(6.90E-03) & \bb{4.15E-01(2.82E-03)$\ddag$} \\
    DAS-CMaOP5(9)& 6.60E-01(2.90E-02) & \bb{6.61E-01(2.90E-02)} & 6.55E-01(8.74E-03) & \bb{7.45E-01(4.31E-03)$\ddag$} & 6.03E-01(1.53E-02) & \bb{7.38E-01(3.68E-03)$\ddag$} \\
    DAS-CMaOP5(10) & 4.89E-01(1.05E-03) & \bb{5.05E-01(1.25E-03)$\ddag$} & 4.67E-01(4.25E-03) & \bb{5.14E-01(7.68E-04)$\ddag$} & 4.41E-01(4.98E-03) & \bb{5.14E-01(2.61E-04)$\ddag$} \\
    DAS-CMaOP5(11) & 6.68E-01(1.27E-02) & \bb{7.75E-01(7.53E-02)$\ddag$} & \bb{2.71E-01(1.25E-02)} & 1.36E-01(8.65E-02)$\dag$ & \bb{1.28E-01(7.67E-03)} & 4.20E-02(2.68E-04)$\dag$ \\
    DAS-CMaOP5(12) & \bb{2.16E-01(1.91E-02)} & 3.56E-02(1.21E-03)$\dag$ & \bb{8.38E-02(9.43E-05)} & 4.34E-02(2.69E-02)$\dag$ & \bb{4.65E-01(1.76E-02)} & 2.61E-02(3.47E-03)$\dag$ \\
    \hline
    \hline
    DAS-CMaOP6(1)& \bb{6.44E-01(2.04E-02)} & 6.02E-01(3.03E-02)$\dag$ & 7.18E-01(1.40E-02) & \bb{7.72E-01(1.28E-02)$\ddag$} & 6.61E-01(1.22E-02) & \bb{7.77E-01(7.65E-03)$\ddag$} \\
    DAS-CMaOP6(2)& 4.38E-01(9.40E-04) & \bb{4.42E-01(3.84E-03)$\ddag$} & 4.77E-01(7.49E-03) & \bb{5.16E-01(2.22E-04)$\ddag$} & 4.53E-01(5.43E-03) & \bb{5.16E-01(1.18E-04)$\ddag$} \\
    DAS-CMaOP6(3)& \bb{7.19E-01(5.30E-03)} & 6.96E-01(5.48E-03)$\dag$ & 8.56E-01(1.44E-02) & \bb{9.02E-01(5.59E-03)$\ddag$} & 8.47E-01(9.69E-03) & \bb{9.23E-01(5.63E-03)$\ddag$} \\
    DAS-CMaOP6(4)& \bb{3.76E-01(1.54E-02)} & 3.68E-01(2.05E-02) & 4.05E-01(8.41E-03) & \bb{4.40E-01(4.17E-03)$\ddag$} & 3.75E-01(5.99E-03) & \bb{4.29E-01(1.74E-03)$\ddag$} \\
    DAS-CMaOP6(5)& \bb{6.05E-01(1.94E-02)} & 5.73E-01(8.74E-03)$\dag$ & 6.96E-01(1.53E-02) & \bb{7.65E-01(1.04E-02)$\ddag$} & 6.50E-01(1.50E-02) & \bb{7.66E-01(7.38E-03)$\ddag$} \\
    DAS-CMaOP6(6)& 4.39E-01(5.77E-04) & \bb{4.41E-01(4.44E-03)$\ddag$} & 4.71E-01(4.43E-03) & \bb{5.16E-01(2.53E-04)$\ddag$} & 4.45E-01(6.02E-03) & \bb{5.16E-01(1.84E-04)$\ddag$} \\
    DAS-CMaOP6(7)& \bb{7.24E-01(4.87E-03)} & 6.90E-01(4.63E-03)$\dag$ & 8.52E-01(1.01E-02) & \bb{8.93E-01(1.17E-02)$\ddag$} & 8.39E-01(1.21E-02) & \bb{9.04E-01(1.47E-02)$\ddag$} \\
    DAS-CMaOP6(8)& \bb{3.60E-01(1.38E-02)} & 3.49E-01(1.37E-02)$\dag$ & 3.92E-01(8.15E-03) & \bb{4.32E-01(6.08E-03)$\ddag$} & 3.70E-01(8.37E-03) & \bb{4.20E-01(2.22E-03)$\ddag$} \\
    DAS-CMaOP6(9)& \bb{5.71E-01(7.75E-03)} & 5.66E-01(5.09E-03)$\dag$ & 6.79E-01(1.33E-02) & \bb{7.47E-01(9.44E-03)$\ddag$} & 6.46E-01(1.57E-02) & \bb{7.48E-01(8.49E-03)$\ddag$} \\
    DAS-CMaOP6(10) & 4.38E-01(5.96E-04) & \bb{4.41E-01(3.63E-03)$\ddag$} & 4.73E-01(9.24E-03) & \bb{5.16E-01(2.31E-04)$\ddag$} & 4.46E-01(6.64E-03) & \bb{5.16E-01(1.38E-04)$\ddag$} \\
    DAS-CMaOP6(11) & \bb{4.96E-01(7.13E-02)} & 2.53E-01(1.94E-01)$\dag$ & \bb{2.88E-01(4.10E-03)} & 5.59E-02(2.48E-02)$\dag$ & \bb{1.37E-01(9.77E-03)} & 4.38E-02(1.02E-03)$\dag$ \\
    DAS-CMaOP6(12) & \bb{3.87E-01(1.37E-02)} & 3.62E-02(1.59E-02)$\dag$ & \bb{8.39E-02(1.48E-04)} & 2.05E-02(1.65E-03)$\dag$ & \bb{5.39E-01(1.38E-02)} & 1.43E-02(4.88E-04)$\dag$ \\
    \bottomrule
    \end{tabular}%
    }
  \label{tab:dascmaop1-6}%
\end{table}%

\begin{table}[htbp]
  \centering
  \caption{Means and standard deviations of $HV$ values obtained by C-MOEA/DD and C-NSGA-III on DAS-CMaOP7-9 with 5, 8 and 10 objectives. Wilcoxon's rank sum test at 0.05 significance level is performed between C-MOEA/DD and C-NSGA-III. $\dag$ and $\ddag$ denote that the performance of C-NSGA-III is significantly worse than or better than that of C-MOEA/DD, respectively. DAS-CMaOP7(i) means that DAS-CMaOP7 has the $i$-th difficulty triplet in Table \ref{tab:dif_triplets}.}
  \scalebox{0.65}[0.65]{
  \begin{tabular}{l|cc|cc|cc}
    \toprule
    \multirow{2}[0]{*}{Test Problem} & \multicolumn{2}{c|}{m = 5}       & \multicolumn{2}{c|}{m = 8}   & \multicolumn{2}{c}{m = 10} \\
    \cline{2-7}
    & \multicolumn{1}{c}{C-MOEA/DD} & \multicolumn{1}{c|}{C-NSGA-III} & \multicolumn{1}{c}{C-MOEA/DD} & \multicolumn{1}{c|}{C-NSGA-III} & \multicolumn{1}{c}{C-MOEA/DD} & \multicolumn{1}{c}{C-NSGA-III} \\

    \hline
    DAS-CMaOP7(1)& 7.31E-01(2.43E-02) & \bb{7.40E-01(3.47E-02)} & 7.38E-01(1.71E-02) & \bb{7.70E-01(7.75E-02)$\ddag$} & 7.13E-01(2.50E-02) & \bb{7.14E-01(1.27E-01)} \\
    DAS-CMaOP7(2)& 5.01E-01(5.83E-04) & \bb{5.11E-01(4.97E-04)$\ddag$} & 4.76E-01(3.43E-03) & \bb{5.14E-01(2.83E-04)$\ddag$} & 4.48E-01(6.58E-03) & \bb{5.14E-01(3.42E-04)$\ddag$} \\
    DAS-CMaOP7(3)& 8.66E-01(1.69E-03) & \bb{8.76E-01(1.38E-03)$\ddag$} & 9.07E-01(5.59E-03) & \bb{9.39E-01(7.74E-03)$\ddag$} & 9.06E-01(8.64E-03) & \bb{9.66E-01(8.03E-03)$\ddag$} \\
    DAS-CMaOP7(4)& \bb{4.22E-01(1.59E-02)} & 4.18E-01(1.86E-02) & 3.97E-01(1.05E-02) & \bb{3.99E-01(4.88E-02)$\ddag$} & 3.76E-01(1.01E-02) & \bb{3.88E-01(3.92E-02)$\ddag$} \\
    DAS-CMaOP7(5)& \bb{7.09E-01(3.11E-02)} & 6.82E-01(3.46E-02)$\dag$ & \bb{7.17E-01(1.40E-02)} & 6.56E-01(1.54E-01) & \bb{6.92E-01(1.89E-02)} & 5.70E-01(1.74E-01) \\
    DAS-CMaOP7(6)& 5.01E-01(4.07E-04) & \bb{5.11E-01(4.26E-04)$\ddag$} & 4.74E-01(4.96E-03) & \bb{5.14E-01(3.85E-04)$\ddag$} & 4.45E-01(6.42E-03) & \bb{5.14E-01(2.27E-04)$\ddag$} \\
    DAS-CMaOP7(7)& 8.61E-01(1.80E-03) & \bb{8.70E-01(1.68E-03)$\ddag$} & 9.00E-01(7.98E-03) & \bb{9.24E-01(8.93E-03)$\ddag$} & 9.00E-01(9.04E-03) & \bb{9.47E-01(1.22E-02)$\ddag$} \\
    DAS-CMaOP7(8)& \bb{4.13E-01(1.69E-02)} & 3.27E-01(6.29E-02)$\dag$ & \bb{3.80E-01(1.13E-02)} & 2.62E-01(7.02E-02)$\dag$ & \bb{3.60E-01(9.87E-03)} & 3.17E-01(8.34E-02) \\
    DAS-CMaOP7(9)& \bb{6.71E-01(2.74E-02)} & 5.56E-01(1.19E-01)$\dag$ & \bb{6.72E-01(1.77E-02)} & 3.92E-01(1.62E-01)$\dag$ & \bb{6.52E-01(1.97E-02)} & 3.75E-01(1.22E-01)$\dag$ \\
    DAS-CMaOP7(10) & 5.01E-01(5.83E-04) & \bb{5.11E-01(3.90E-04)$\ddag$} & 4.76E-01(5.88E-03) & \bb{5.14E-01(4.22E-04)$\ddag$} & 4.44E-01(6.12E-03) & \bb{5.14E-01(2.64E-04)$\ddag$} \\
    DAS-CMaOP7(11) & \bb{6.24E-01(3.45E-02)} & 5.72E-01(6.82E-02)$\dag$ & \bb{2.76E-01(2.17E-02)} & 3.61E-02(6.84E-11)$\dag$ & \bb{1.26E-01(6.84E-03)} & 4.76E-02(1.72E-07)$\dag$ \\
    DAS-CMaOP7(12) & \bb{1.88E-01(2.93E-02)} & 2.19E-02(3.33E-05)$\dag$ & \bb{1.68E-01(1.73E-01)} & 2.10E-02(5.63E-05)$\dag$ & \bb{5.40E-01(1.78E-02)} & 1.44E-02(2.30E-05)$\dag$ \\
    \hline
    \hline
    DAS-CMaOP8(1)& \bb{6.74E-01(1.79E-02)} & 6.62E-01(2.87E-02) & 7.28E-01(6.03E-03) & \bb{7.37E-01(4.23E-03)$\ddag$} & 7.64E-01(4.02E-03) & \bb{7.77E-01(3.45E-03)$\ddag$} \\
    DAS-CMaOP8(2)& 5.02E-01(1.90E-03) & \bb{5.13E-01(6.55E-04)$\ddag$} & 4.89E-01(5.11E-03) & \bb{5.14E-01(7.99E-03)$\ddag$} & 4.57E-01(4.37E-03) & \bb{5.15E-01(4.12E-03)$\ddag$} \\
    DAS-CMaOP8(3)& 7.68E-01(1.26E-02) & \bb{7.69E-01(3.35E-03)$\ddag$} & 8.36E-01(4.39E-03) & \bb{8.51E-01(2.02E-03)$\ddag$} & 8.89E-01(3.96E-03) & \bb{9.06E-01(3.75E-03)$\ddag$} \\
    DAS-CMaOP8(4)& \bb{4.34E-01(1.14E-02)} & 4.30E-01(2.15E-02) & 4.16E-01(5.20E-03) & \bb{4.40E-01(2.60E-03)$\ddag$} & 4.04E-01(1.13E-02) & \bb{4.29E-01(2.16E-03)$\ddag$} \\
    DAS-CMaOP8(5)& \bb{6.63E-01(1.50E-02)} & 6.32E-01(2.88E-02)$\dag$ & 7.07E-01(3.16E-02) & \bb{7.29E-01(6.89E-03)$\ddag$} & 7.47E-01(1.42E-02) & \bb{7.68E-01(3.73E-03)$\ddag$} \\
    DAS-CMaOP8(6)& 5.03E-01(5.26E-04) & \bb{5.13E-01(7.76E-04)$\ddag$} & 4.93E-01(2.89E-03) & \bb{5.08E-01(8.78E-03)$\ddag$} & 4.66E-01(7.78E-03) & \bb{5.10E-01(5.31E-03)$\ddag$} \\
    DAS-CMaOP8(7)& \bb{7.70E-01(1.24E-02)} & 7.58E-01(6.65E-03)$\dag$ & \bb{8.41E-01(1.18E-02)} & 7.61E-01(1.88E-02)$\dag$ & \bb{8.84E-01(1.24E-02)} & 8.22E-01(4.17E-02)$\dag$ \\
    DAS-CMaOP8(8)& \bb{4.20E-01(1.36E-02)} & 3.91E-01(2.42E-02)$\dag$ & 4.02E-01(1.53E-02) & \bb{4.26E-01(4.82E-03)$\ddag$} & 4.02E-01(3.44E-03) & \bb{4.19E-01(3.00E-03)$\ddag$} \\
    DAS-CMaOP8(9)& \bb{6.36E-01(2.51E-02)} & 5.84E-01(2.80E-02)$\dag$ & 6.74E-01(5.14E-02) & \bb{7.08E-01(1.72E-02)$\ddag$} & 7.19E-01(3.48E-02) & \bb{7.55E-01(4.33E-03)$\ddag$} \\
    DAS-CMaOP8(10) & \bb{5.02E-01(1.46E-03)} & 4.84E-01(2.18E-02)$\dag$ & 4.79E-01(8.86E-03) & \bb{4.98E-01(6.15E-03)$\ddag$} & 4.93E-01(2.35E-03) & \bb{5.04E-01(1.91E-03)$\ddag$} \\
    DAS-CMaOP8(11) & \bb{5.37E-01(3.47E-02)} & 3.88E-01(2.08E-01)$\dag$ & \bb{2.73E-01(1.26E-02)} & 3.61E-02(6.26E-08)$\dag$ & \bb{1.41E-01(3.27E-03)} & 5.25E-02(2.18E-02)$\dag$ \\
    DAS-CMaOP8(12) & \bb{2.21E-01(2.49E-02)} & 2.78E-02(4.23E-03)$\dag$ & \bb{4.83E-01(1.18E-02)} & 2.11E-02(2.14E-06)$\dag$ & \bb{5.24E-01(1.14E-02)} & 1.44E-02(1.12E-06)$\dag$ \\
    \hline
    \hline
    DAS-CMaOP9(1)& 5.64E-01(1.16E-02) & \bb{5.80E-01(1.45E-02)$\ddag$} & 6.39E-01(1.73E-02) & \bb{7.19E-01(2.62E-02)$\ddag$} & 5.87E-01(2.17E-02) & \bb{7.11E-01(2.19E-02)$\ddag$} \\
    DAS-CMaOP9(2)& \bb{4.04E-01(1.93E-03)} & 4.02E-01(4.00E-03)$\dag$ & 4.45E-01(5.55E-03) & \bb{4.99E-01(1.64E-03)$\ddag$} & 4.34E-01(6.10E-03) & \bb{5.01E-01(1.32E-03)$\ddag$} \\
    DAS-CMaOP9(3)& \bb{6.36E-01(1.23E-02)} & 6.52E-01(1.35E-02)$\dag$ & 8.01E-01(9.72E-03) & \bb{8.80E-01(1.75E-02)$\ddag$} & 7.91E-01(1.59E-02) & \bb{9.08E-01(9.63E-03)$\ddag$} \\
    DAS-CMaOP9(4)& 3.45E-01(3.42E-03) & 3.45E-01(5.41E-03) & 3.68E-01(7.64E-03) & \bb{4.04E-01(6.69E-03)$\ddag$}             & 3.57E-01(1.18E-02) & \bb{3.92E-01(4.99E-03)$\ddag$} \\
    DAS-CMaOP9(5)& 5.48E-01(1.22E-02) & \bb{5.68E-01(1.00E-02)$\ddag$} & 6.10E-01(2.65E-02) & \bb{6.89E-01(1.94E-02)$\ddag$} & 4.27E-01(6.90E-02) & \bb{6.98E-01(3.41E-02)$\ddag$} \\
    DAS-CMaOP9(6)& \bb{4.06E-01(1.50E-03)} & 4.01E-01(2.70E-03)$\dag$ & 4.46E-01(7.64E-03) & \bb{4.99E-01(1.67E-03)$\ddag$} & 4.24E-01(5.74E-03) & \bb{5.02E-01(7.63E-04)$\ddag$} \\
    DAS-CMaOP9(7)& \bb{6.45E-01(1.45E-02)} & 6.35E-01(9.99E-03)$\dag$ & 7.85E-01(1.91E-02) & \bb{8.22E-01(1.64E-02)$\ddag$} & 7.71E-01(2.12E-02) & \bb{8.64E-01(1.42E-02)$\ddag$} \\
    DAS-CMaOP9(8)& \bb{3.29E-01(6.23E-03)} & 3.26E-01(9.66E-03) & 3.40E-01(1.35E-02) & \bb{3.92E-01(1.47E-02)$\ddag$} & \bb{4.12E-01(2.58E-02)} & 3.53E-01(3.36E-02)$\dag$ \\
    DAS-CMaOP9(9)& 5.08E-01(2.98E-02) & \bb{5.26E-01(3.26E-02)} & 4.87E-01(6.41E-02) & \bb{6.32E-01(5.59E-02)$\ddag$} & 4.19E-01(3.72E-02) & \bb{6.58E-01(2.48E-02)$\ddag$} \\
    DAS-CMaOP9(10) & 4.05E-01(1.77E-03) & \bb{4.06E-01(2.90E-03)} & 4.45E-01(1.21E-02) & \bb{4.99E-01(1.73E-03)$\ddag$} & 4.25E-01(8.70E-03) & \bb{5.02E-01(1.06E-03)$\ddag$} \\
    DAS-CMaOP9(11) & \bb{2.74E-01(3.89E-02)} & 2.08E-01(1.01E-01) & \bb{2.39E-01(9.42E-03)} & 5.72E-02(4.19E-02)$\dag$ & \bb{1.19E-01(4.03E-03)} & 3.80E-02(1.34E-06)$\dag$ \\
    DAS-CMaOP9(12) & \bb{3.29E-01(1.66E-02)} & 2.34E-02(1.99E-03)$\dag$ & \bb{4.14E-01(3.55E-02)} & 2.10E-02(1.50E-03)$\dag$ & \bb{4.14E-01(2.54E-02)} & 1.69E-02(1.35E-03)$\dag$ \\
    \bottomrule
    \end{tabular}%
    }
  \label{tab:dascmaop7-9}%
\end{table}%

\section{Conclusion} \label{section:conclusion}
In this work, we proposed a construction toolkit to build difficulty-adjustable and scalable CMOPs. The method used to design the construction toolkit is based on three primary constraint functions identified to correspond to three primary difficulty types of DAS-CMOPs. The method is also scalable, because both the number of objectives and the number of constraints can be conveniently extended. As an example, a set of DAS-CMOPs (DAS-CMOP1-9) and a set of DAS-CMaOPs (DAS-CMaOP1-9) were generated using this construction toolkit. To verify the usefulness of the suggested test instances, comprehensive experiments were conducted to test the performance of two popular CMOEAs (MOEA/D-CDP and NSGA-II-CDP) on DAS-CMOPs and two CMaOEAs (C-MOEA/DD and C-NSGA-III) on DAS-CMaOPs with a variety of difficulty triplets. Through analyzing the performance of the four test algorithms, it was found that the three primary types of difficulties did exist in the corresponding test problems, and that the algorithms under test showed different behaviors in their approaches to the $PFs$. The observation demonstrates that the proposed method of constructing the CMOPs and CMaOPs is very effective and that the resulting functions are useful in evaluating the performance of CMOEAs and CMaOEAs. In the future, we will construct problems with both continuous and discrete decision variables to study their impact on the definition of difficulty types and levels of the optimization problems. In other future work, we will investigate the change in difficulty levels with an increasing number of variables, and test the performance of push and pull search algorithm \citep{FAN2019665} on CMOPs with a large number of variables.


\begin{acknowledgments}\\
This research work was supported by the Key Lab of Digital Signal and Image Processing of Guangdong Province, by the National Natural Science Foundation of China under Grant (61175073, 61300159, 61332002, 51375287), by the Natural Science Foundation of Jiangsu Province of China under grant SBK2018022017, by the China Postdoctoral Science Foundation under grant 2015M571751, by the Project of International, as well as Hongkong, Macao\&Taiwan Science and Technology Cooperation Innovation Platform in Universities in Guangdong Province (2015KGJH2014) and by the State Key Lab of Digital Manufacturing Equipment \& Technology under grant DMETKF2019020.
\end{acknowledgments}

\bibliographystyle{apalike}
\bibliography{das-cmops}

\end{document}